\documentclass{article}

\usepackage{arxiv}								   
	 
\usepackage[utf8]{inputenc}			 
\usepackage{url} 								   
\usepackage{hyperref} 					   
\usepackage{amsfonts}       		       
\usepackage{nicefrac}      	 				 
\usepackage{microtype}      			 
\usepackage{lipsum}
\usepackage{amsthm}
\usepackage{xfrac}  
\usepackage{lmodern}

\title{Fast and interpretable Support Vector Classification based on the truncated ANOVA decomposition}
\date{August 26, 2024}
\author{ Kseniya Akhalaya \\
	Faculty of Mathematics\\
	Chemnitz University of Technology\\
	09107 Chemnitz \\
	\texttt{kseniya.akhalaya@math.tu-chemnitz.de} \\ 
	\And
	Franziska Nestler\\
	Faculty of Mathematics\\
	Chemnitz University of Technology\\
	09107 Chemnitz \\
	\texttt{franziska.nestler@math.tu-chemnitz.de} \\
	\And
	Daniel Potts \\
	Faculty of Mathematics\\
	Chemnitz University of Technology\\
	09107 Chemnitz \\
	\texttt{daniel.potts@math.tu-chemnitz.de} \\
}

\usepackage{amsmath}
\usepackage{verbatim}
\usepackage[svgnames, table, hyperref, dvipsnames]{xcolor} 
\usepackage{amsfonts, mathtools, amssymb, amsopn, bm, mathtools}
\usepackage{mathrsfs, cleveref}
\usepackage{algorithm, algpseudocode}
\usepackage{booktabs, siunitx}
\usepackage{multirow}
\usepackage{pgfplots}
\usetikzlibrary{pgfplots.groupplots}
\usepackage{dsfont}
\usepackage{subcaption}
\usepackage{tikz}
\usepackage{pgfplotstable}
\usepackage{graphicx}
\usepackage{adjustbox}
\usepackage{upgreek}
\usepackage{diagbox}

\usepackage{xfrac}
\usepackage{comment}
\pgfplotsset{compat=1.8}
\usepgfplotslibrary{statistics}

\renewcommand{\shorttitle}{Fast Support Vector Classification based on the truncated ANOVA decomposition}

\hypersetup{
	pdftitle={Fast and interpretable Support Vector Classification based on the truncated ANOVA decomposition},
	pdfsubject={scientific computing},
	pdfauthor={Kseniya Akhalya, Franziska Nestler, Daniel Potts},
	pdfkeywords={Support Vector Machines, primal SVM, {\texorpdfstring{$\ell_1$}{l1}-norm regularization}, interpretability, ANOVA, FISTA, gradient descent, trigonometric feature maps, wavelets, grouped transformations},
}

\renewcommand{\b}{\bm}

\newcommand{\C}{\mathbb C}
\newcommand{\N}{\mathbb N}
\newcommand{\R}{\mathbb R}
\newcommand{\Z}{\mathbb Z}
\newcommand{\T}{\mathbb T}
\newcommand{\D}{\mathbb D}
\newcommand{\X}{\mathcal X}
\newcommand{\I}{\mathcal{I}}

\newcommand{\x}{\b x}
\newcommand{\y}{\b y}
\renewcommand{\k}{\b k}
\renewcommand{\j}{\b j}

\newcommand{\f}{\b f}

\newcommand{\w}{{\b w}} 
\newcommand{\bfk}{{\boldsymbol k}}

\newcommand{\bfN}{{\boldsymbol N}}
\newcommand{\bfP}{{\boldsymbol \Phi}}
\newcommand{\chui}{{\text{chui}}}

\renewcommand{\u}{{\b u}}
\renewcommand{\v}{\bm v}
\newcommand{\uc}{\b{u}^{\mathrm c}}

\newcommand{\bfh}{\hat{\f}}

\newcommand{\fh}{\hat{f}}
\newcommand{\sign}{\text{sign}}
\newcommand{\abs}[1]{\left|#1\right|}
\newcommand{\norm}[2]{\left\| #1 \right\|_{#2}}

\DeclareMathOperator*{\supp}{supp}
\DeclareMathOperator*{\argmin}{arg\,min}

\DeclareMathOperator*{\prox}{prox}

\definecolor{blue35}{RGB}{158,191,243}
\definecolor{blue50}{RGB}{122,168,243}
\definecolor{blue65}{RGB}{85,146,243}
\definecolor{blue85}{RGB}{36,116,243}
\definecolor{blue100}{RGB}{0,101,243}

\definecolor{orange35}{RGB}{255,211,166}
\definecolor{orange50}{RGB}{255,192,128}
\definecolor{orange65}{RGB}{255,173,90}
\definecolor{orange85}{RGB}{255,147,38}
\definecolor{orange100}{RGB}{255,158,0}

\numberwithin{equation}{section}
\numberwithin{table}{section}
\numberwithin{figure}{section}

\begin{document}
	
	\maketitle
	
	\begin{abstract}
		
	Support Vector Machines (SVMs) are an important tool for performing classification on scattered data, where one usually has to deal with many data points in high-dimensional spaces. We propose solving SVMs in primal form using feature maps based on trigonometric functions or wavelets. In small dimensional settings the Fast Fourier Transform (FFT) and related methods are a powerful tool in order to deal with the considered basis functions. For growing dimensions the classical FFT-based methods become inefficient due to the curse of dimensionality. Therefore, we restrict ourselves to multivariate basis functions, each of which only depends on a small number of dimensions. This is motivated by the well-known sparsity of effects and recent results regarding the reconstruction of functions from scattered data in terms of truncated  analysis of variance (ANOVA) decompositions, which makes the resulting model even interpretable in terms of importance of the features as well as their couplings. The usage of small superposition dimensions has the consequence that the computational effort no longer grows exponentially but only polynomially with respect to the dimension. In order to enforce sparsity regarding the basis coefficients, we use the frequently applied $\ell_2$-norm and, in addition, $\ell_1$-norm regularization. The found classifying function, which is the linear combination of basis functions, and its variance can then be analyzed in terms of the classical ANOVA decomposition of functions. Based on numerical examples we show that we are able to recover the signum of a function that perfectly fits our model assumptions. Furthermore, we perform classification on different artificial and real-world data sets. We obtain better results with $\ell_1$-norm regularization, both in terms of accuracy and clarity of interpretability.
		
	\end{abstract}
	
	\noindent\textit {\textbf{Keywords:}} Support Vector Machines, primal SVM, {\texorpdfstring{$\ell_1$}{l1}-norm regularization}, interpretability, ANOVA, FISTA, gradient descent, trigonometric feature maps, wavelets, grouped transformations

	\section{Introduction}\label{sec1}
	
In the following, we shortly review some basic work on Support Vector Machines (SVMs) for classification problems using the general references \cite{SteinwartChristmann, Steidl}, which we also recommend for all further details and interesting theoretical results.
Assume, we are given a finite training data set $(\x_1,y_1),(\x_2,y_2),\dots,(\x_M,y_M)\in\X\times\{-1, +1\}$. 
Here, $\X$ is a nonempty finite set consisting of feature vectors $\x_j\in\R^d$, which we also call inputs or rather instances. 
The $y_j$ are called labels or outputs. 
The classical SVM approach aims to construct a classifier $F:\mathcal{X}\to\{-1,+1\}$, also called decision function or  predictor, of the form 
\begin{equation}\label{classifier}
	F(\x)=\sign\left(\langle \w, \x\rangle +b\right),\quad \w \in \R^d, b\in\R,
\end{equation}
where $\langle\cdot,\cdot\rangle:\mathbb{R}^d\times\mathbb{R}^d\to\mathbb{R}$ is an inner product on $\mathbb{R}^d$ with the agreement that $\sign(0):=1$. The main goal is to find a weight vector $\w \in \R^d$ and bias $b\in\R$ such that the prediction given by $F(\x_j)$ is correct for many or ideally all training data points $(\x_j,y_j)$, $j=1,2,\dots,M$. 
Let us assume that the hyperplane defined by 
\begin{equation*}
	H(\w,b):= \left\{\x\in\R^d:\langle \w, \x\rangle+b=0\right\}
\end{equation*}
with normal vector $\w/\norm{\w}{2}$ perfectly separates the training data set according to the given labels $y_j$,
meaning that feature vectors $\x_j$ with negative labels $y_j$ are on the negative side of the hyperplane $H(\w,b)$, i.e., $\langle \w, \x_j\rangle+b<0$ for $j\in\{j\in{1,2,\dots,M}: y_j=-1\}$, and feature vectors $\x_j$ with positive labels $y_j$ are on the positive side, i.e., $\langle \w, \x_j\rangle+b \geq0$ for $j\in \{j\in{1,2,\dots,M}: y_j=+1\}$.
These two conditions are often presented in a single equation $y_j\left(\langle \w, \x_j\rangle+b\right)>0$ and the data set is called linearly separable if it can be fulfilled for all training data points. For a linearly separable data set there may exist various separating hyperplanes. 
To find the best one, we first compute the distance between a point, described by input vector $\x_j$, and a separating hyperplane $H(\w,b)$ by
\begin{equation*}
	y_j\left(\left\langle \frac{\w}{\norm{\w}{2}}, \x_j\right\rangle+\frac{b}{\norm{\w}{2}}\right),\quad j=1,2,\dots, M
\end{equation*}
to derive the margin, which is defined as smallest distance,
\begin{equation*}
	\gamma
	:= \min_{j=1,2,\dots, M}y_j\left(\left\langle \frac{\w}{\norm{\w}{2}}, \x_j\right\rangle+\frac{b}{\norm{\w}{2}}\right).
\end{equation*}
Further, we choose the separating hyperplane that maximizes the margin between these two classes of training data. The corresponding classifier is called maximal margin classifier or hard margin classifier, which was first introduced in~\cite{VapnikLerner}, and is defined as
$$
\gamma\norm{\w}{2}=\min_{j=1,2,\dots,M}y_j\left(\langle\w, \x_j\rangle+b\right).
$$
Finally, by using an additional scaling assumption $y_j\left(\langle \w, \x_j\rangle+b\right)\geq1$,
we see that $\gamma$ becomes maximal if and only if $\norm{\w}{2}$ becomes minimal. In summary, the hard margin classifier aims to find parameters $\w\in\mathbb{R}^d$ and $b\in\mathbb{R}$ solving the following quadratic optimization problem with linear constraints
\begin{equation}\label{hard_margin}
	\frac{1}{2}\norm{\w}{2}^2\rightarrow\min_{\w, b}\quad\text{subject to}\quad y_j\left(\langle\w, \x_j\rangle+b\right)\geq 1,\quad j=1,2,\dots,M
\end{equation}
and unique solution, see~\cite[Chapter 5]{vapnik95}. As we can see, the formulation does not allow any violations of the margin condition, hence the expression "hard".  There are many different methods for solving such optimization problems efficiently, see for example~\cite[Chapter 16]{NocedalWright}. 

To resolve the issue that a given data set is not linearly separable, we use a nonlinear SVM, first proposed in~\cite{BoserGuyonVapnik}, which maps the feature vectors  $\x_j\in\X$ into a possibly infinite-dimensional space $H$, the so-called feature space, by a typically non-linear map $\Phi: \D^d\to H,$ defined on a bounded domain $\mathbb D^d\supset\X$, called feature map. 
In the presence of noise, it can happen that we need to misclassify some feature vectors $\x_j$ in order to avoid overfitting, especially if $d\geq M$. 
Recall that in~\eqref{hard_margin} the constraints force the hyperplanes to cause no error on the training data set. In contrast, the soft margin approach, suggested in~\cite{CortesVapnik}, is obtained by introducing additional slack variables $\xi_j$,  $j=1,2,\dots,M$ and a regularization parameter $C>0$. 
In this case, the resulting primal optimization problem is given by
\begin{equation}\label{soft_margin}
	\setlength{\jot}{-5pt}
	\begin{split}
		\frac{1}{2}\norm{\w}{2}^2+C\sum_{j=1}^M\xi_j\rightarrow \min_{\w, b, \boldsymbol{\xi}}\quad 	\text{subject to}\quad& y_j\left(\left\langle\w,\x_j)+b\right\rangle\right)\geq 1-\xi_j,\\
		&\xi_j\geq 0, \quad j=1, 2,\dots, M,
	\end{split}
\end{equation}
where $\b\xi=(\xi_j)_{j=1}^M\in\R^M$ is vector of slack variables.
As mentioned above, the nonlinear case is now treated based on feature maps.
In this paper we will focus on feature maps $\Phi$ given by
\begin{equation}\label{feature_map}
	\Phi: \mathbb D^d \to \R^{|\I|}, \quad \x\mapsto\Phi(\x):=\left(\varphi_{\bfk}(\x)\right)_{\bfk\in\I},
\end{equation}
where $\varphi_{\bfk}(\x): \mathbb D^d\to\R$ are basis functions with certain multi indices $\bfk\in\I\subset\Z^d$.
The choice of basis functions does not depend directly on the data set. However, it is advisable to use suitable functions to achieve better classification results.
At this point, the question arises as to how this index set $\I$ should be chosen, which is not trivial, especially in high-dimensional spaces, referring to the case of a large number of features $d$.
The choice of the index set $\I$ is very important here and may have a major influence on the performance of the considered algorithms. For details concerning the choice of an index set $\I$ we refer to Section~\ref{theory}.
Using a function $f:\X \to\R$,  we rewrite~\eqref{classifier} as
\begin{equation*}
	F(\x):=\sign\left(f(\x)\right)=\sign\left(\sum_{\bfk\in\I}\fh_{\bfk}\varphi_{\bfk}(\x)\right), \quad \x\in\mathcal{X}, \I\subset\Z^d,
\end{equation*}
i.e., the affine linear function $\langle \w, \x\rangle +b$ is replaced by a linear combination of the basis functions $\varphi_{\bfk}$ with coefficients $\fh_{\bfk}\in\R$ to be determined. Note that we omit the bias $b\in\R$ and include a constant function $\varphi_{\mathbf 0}(\x)\equiv1$ instead.
In the following, we will call $f$ the classifying function, which is again affine linear in the enlarged feature space $\{\left(\varphi_{\b k}(\x)\right)_{\b k \in \I}, \x\in \D^d\}\subset\R^{|\mathcal I|}$, but non-linear in the original  $\mathbb D^d$. Combining this idea  with the slack variables $\xi_j$ we introduced before, we obtain the constrained optimization problem 
\begin{equation}\label{qp_constrained}
	\setlength{\jot}{-5pt}
	\begin{split}
		\frac{1}{2}R(\bfh)+C\sum_{j=1}^M\xi_j\rightarrow \min_{\bfh, \boldsymbol{\xi}}\quad 	\text{subject to}\quad& y_j\left(\left\langle\bfh, \Phi(\x_j)\right\rangle\right)\geq 1-\xi_j,\\
		&\xi_j\geq 0, \quad j=1, 2,\dots, M
	\end{split}
\end{equation}
for the vector of basis coefficients $\bfh=(\hat f_{\b k})_{\b k\in\I}\in \R^{|\I|}$ and regularization term $R(\bfh)$. In this paper, we will restrict our considerations to the cases $R(\bfh)=\|\bfh\|_2^2$ ($\ell_2$-norm regularization) and $R(\bfh)=\|\bfh\|_1$ ($\ell_1$-norm regularization).
Since the bias $b$ is now included in our feature map, also the constant part of the function $f$ is penalized. This can make a small difference at the end, but is simpler in terms of implementation and is not uncommon when using non-linear regression, for example.

Further, we multiply the objective function in \eqref{qp_constrained} by $2\lambda:= \frac{1}{MC}$ and use a loss function $L:\{-1, +1\}\times\R\to[0,\infty)$  to reformulate the constrained problem as an unconstrained one 
\begin{equation}\label{qp_unconstrained}
	\lambda R(\bfh)+\frac{1}{M}\sum_{j=1}^ML\left(y_j, \left\langle\bfh, \Phi(\x_j)\right\rangle\right)\rightarrow \min_{\bfh}, \quad \bfh\in \R^{|\I|}
\end{equation}
with regularization parameter $\lambda>0$. The loss function $L$ is intended to penalize the presence of large slack variables $\xi_j$, i.e., it is ideally zero for all training data points $(\x_j,y_j)$, $j=1,2,\dots,M$. 

The well-known hinge-loss is typically used in the context of SVMs. We will focus our attention on the squared hinge loss or rather truncated least squares loss function defined by
\begin{equation*}
	L\left(y_j, \langle\bfh, \Phi(\x_j)\rangle\right):= \left(\max\left\{0, 1-y_j\left\langle\bfh, \Phi(\x_j)\right\rangle\right\}\right)^2,
\end{equation*}
since it is smooth. But also other loss functions, as for example least squares loss, are possible, see~\cite[Chapter 2]{SteinwartChristmann}.
At this point we would like to mention that well-known libraries for solving such SVMs are LIBSVM~\cite{libsvm} and LIBLINEAR~\cite{liblinear}.
These are two open source machine learning libraries, with existing interfaces and extensions for many programming languages, e.g., Python or Julia.
While LIBSVM incorporates algorithms for kernelized SVMs, LIBLINEAR implements linear SVMs. The mentioned references give an overview about theoretical convergence results, some implementation issues and performance comparisons to other state-of-the-art solvers.

As we can see, the standard form of SVM uses $\ell_2$-norm regularization, i.e., $R(\bfh)=\|\bfh\|_2^2$. By shrinking the magnitude of the coefficients, the $\ell_2$-norm penalty reduces the variance of the estimated coefficients, and thus can achieve better prediction accuracy.
However, the $\ell_2$-norm penalty does in general not produce highly sparse coefficients and hence cannot automatically perform variable selection very well.
This is the major limitation for applying support vector classification for high-dimensional data, where variable selection is essential for providing reasonable interpretations.
Therefore, we use the expanded traditional SVM formulation, see~\cite{TibshiraniZhu}, to enforce more sparsity and better variable selection, using a different regularization approach, e.g., $\ell_1$-norm regularization or rather a lasso penalty term $R(\bfh)=\|\bfh\|_1$ in~\eqref{qp_unconstrained}, see~\cite{Tibshirani}. 

It should also be noted that usually SVMs are solved in the dual form including a kernel matrix of size $M\times M$, cf.~\cite{SchSmo2018} and references therein. If the regularization $R(\bfh)=\|\bfh\|_2^2$ is used in \eqref{qp_constrained}, for instance, the dual optimization problem reads as
\begin{equation}\label{dual}
	\setlength{\jot}{-5pt}
	\begin{split}
		\frac{1}{2}\b\alpha^\top Q\b\alpha  -\b1\b\alpha\rightarrow \min_{\b\alpha}\quad 	\text{subject to}\quad& \b y^\top \b\alpha=0,\\
		&0\geq \alpha_j\geq C, \quad j=1, 2,\dots, M,
	\end{split}
\end{equation}
where $Q\in\R^{M\times M}$ is the so-called kernel matrix, which is typically positive (semi-)definite, and $Q_{ij}=y_iy_jK(\x_i,\x_j)$ for a given kernel function $K(\cdot,\cdot)$.
A frequently used kernel function is the radial basis function (RBF) kernel or rather the Gaussian kernel, for instance. Kernel matrices are typically non-sparse and badly conditioned, which makes it hard to work with them in case of large data sets.
Thus, it has become a common approach in the field of kernel methods to replace the kernel matrix $Q$ by a low-rank approximation $Q\approx\b\Psi^\top\b\Psi$ in order to compute matrix-vector products more efficiently. This is the main idea of the well-known random Fourier features approach~\cite{RaRe2007}, for instance.
A similar approach is applied in \cite{NeStWa2023,WaPeSt2023}, where the authors use an expansion of the kernel function $K$ in terms of trigonometric basis functions and make use of nonuniform FFTs in order to compute the matrix-vector products in an efficient manner.
In both settings we go back from the kernel setting to the feature-map formulation, where one can jump between primal and dual formulations.

This is the main motivation for the present work. Instead of taking a detour through the kernel formulation and solving the dual optimization problem, we directly work with the original or rather primal formulation combined with a finite dimensional feature map~\eqref{feature_map}.
Our feature map is built from basis functions, with which we can work efficiently for huge amount of data. If the optimization \eqref{qp_unconstrained} is solved iteratively, one has to compute matrix-vector products of the form
\begin{equation*}
	\b\Phi\bfh=\left(\left\langle\bfh, \Phi(\x_j)\right\rangle\right)_{j=1}^M, \quad 
	\b\Phi:=\left(\varphi_{\b k}(\x_j)\right)_{j=1,2,\dots,M,\b k\in\I}\in\R^{M\times|\I|}
\end{equation*}
in each iteration. 
In order to obtain an efficient algorithm, it is essential to replace the classic matrix-vector multiplication with a more favorable alternative. In case of trigonometric bases, for instance, we can efficiently approximate the required matrix-vector products using FFT-based methods for nonequispaced data~\cite{duro93,bey95,st97, PlPoStTa18}.
When making use of wavelets, the resulting matrices are sparse and, thus, the matrix-vector products can be computed efficiently, see~\cite[Chapter 3]{Daubechies} or~\cite[Chapter 6]{Walnut}. 

In order to overcome the curse of dimensionality, we assume that high-dimensional interactions between features are negligible and restrict our feature map $\Phi$ to multivariate basis functions that only depend on a small number of dimensions each, which is equivalent to a corresponding restriction of the index set $\I$. The underlying rather simple structure of orthonormal systems makes it even possible to understand the importance of single features or couplings between them, based on the analysis of variance (ANOVA) decomposition. This work thus continues the previous works~\cite{PoSc19a,PoSc19b,lippert}, which dealt exclusively with regression problems. Combining this idea with a customized SVM approach for classification problems, we are able to solve classification tasks in an interpretable fashion.
In summary, our main contribution is the transfer of the ANOVA-based regression idea to classification scenarios by using the corresponding feature maps within SVMs.

This paper is organized as follows. In Section~\ref{sec:fourier_anova} we introduce the applied basis functions as well as the classical ANOVA decomposition of functions. Furthermore, we summarize certain properties of this decomposition and discuss its interpretability in terms of Sobol indices,~\cite{So90, So01}. Finally, the efficient computation based on grouped transformations~\cite{BaPoSc} is explained.
After introducing all the basic concepts in Section 2, we continue with the main part of the paper in Section 3, where we apply this to classification problems for the first time.
We continue with the discussion of the implemented optimization algorithms in Section~\ref{sec:algorithms}, where we distinguish between $\ell_2$-norm and $\ell_1$-norm regularization. Both penalty terms are considered in combination with the squared hinge loss, which is smooth. Thus, the $\ell_2$-norm regularized optimization problem is solved with a gradient descent method, whereas the $\ell_1$-norm approach is treated based on FISTA~\cite{Beck1}.
The discussed algorithms have been implemented as a part of the publicly available Julia software package \textit{ANOVAapprox.jl}~\cite{gitANOVAjl}.
Finally, we present first numerical experiments in Section~\ref{sec:examples}, which are also included as examples in the Julia package mentioned above, and conclude with a short summary in Section~\ref{sec:summary}.

\section{ANOVA Decomposition for Interpretability}

In this section we define the classical analysis of variance (ANOVA) decomposition of functions, cf.~\cite{CaMoOw97, LiOw06, KuSlWaWo09, Gu2013}, and summarize the main theoretical results presented in~\cite{PoSc19a, PoSc19b, lippert}, where the ANOVA decomposition was studied in the context of series expansions involving trigonometric functions or wavelets, see Section~\ref{theory}.
Furthermore, the interpretability of the ANOVA decomposition based on the Sobol indices or rather global sensitivity indices, cf.~\cite{So90, So01}, is discussed in Section~\ref{sec:fourier_anova}.
Finally, we take a closer look into the results presented in \cite{BaPoSc}, namely the concept of grouped index sets and grouped transformations, which provide us an improvement in the complexity for the evaluation of the arising matrix-vector products in~\eqref{qp_unconstrained}, see Section~\ref{grouptransform}.

\subsection{Function spaces, systems and frequency sets}\label{theory}

In this section we lay the foundation  and present basic definitions, that are indispensable for the rest of this paper.
In our setting, we consider square-integrable functions over a domain $\D \in \{\T, [0,\sfrac{1}{2}]\}$ with the torus $\T:=\R/\Z\simeq[-\sfrac{1}{2},\sfrac{1}{2})$, i.e.,
\begin{equation*}
	f\in L_2(\D^d):=\left\{f:\D^d\rightarrow\C:\norm{f}{L_2(\D^d)}:=\sqrt{\int_{\D^d}\abs{f(\x)}^2\mathrm{d}\x}<\infty\right\}
\end{equation*}
with dimension $d\in\N$. 
Note that $L_2(\D^d)$ is a separable Hilbert space with inner product
\begin{equation*}
	\langle f,g\rangle_{L_2(\D^d)}=\frac1{\tau^d}\int_{\D^d}f(\x)\overline{g(\x)}\,\mathrm{d}\x,\quad f,g\in L_2(\D^d),
\end{equation*}
where we denote by $\tau\in\{1, \sfrac 12\}$ the corresponding interval length.
Using the property of separability, we know that there exists a countable complete orthonormal system or basis in the separable Hilbert space. In the case of tensor product spaces, we are able to construct a basis from the one-dimensional case as follows. 
We construct a basis of $L_2(\D^d)$  for $d>1$ by defining the basis functions
\begin{equation}\label{eq:tensorphi}
	\varphi_{\bfk}(\x):=\prod_{i=1}^{d}\varphi_{k_i}(x_i),
\end{equation}
for $\b k=\left(k_i\right)_{i=1}^d\in\Z^d$.
Now we may write any $f\in L_2(\D^d)$ in terms of a basis expansion 
\begin{equation*}
	f (\x)= \sum_{\bfk\in\Z^d} c_{\bfk}(f)\varphi_{\bfk}(\x)
\end{equation*}
with the basis coefficients $c_{\bfk}(f)=\langle f,\varphi_{\bfk}\rangle$, which are simply obtained as inner products of the single basis functions and the function $f$, due to the orthogonality. Note, that Parseval's identity 
\begin{equation*}
	\norm{f}{L_2(\D^d)}=\sqrt{{\sum_{\bfk\in\Z^d}\abs{c_{\bfk}(f)}^2}}
\end{equation*} 
holds true, see~\cite[Chapter 1]{PlPoStTa18}.

In general, a function $f\in L_2(\D^d)$  has an infinite number of basis coefficients $c_{\bfk}(f)$, $\bfk\in\Z^d$, of course. However, if the coefficients $c_{\bfk}(f)$ become negligibly small for large indices $\bfk$, which is typically the case for sufficiently smooth functions, we may truncate the infinite sum and replace $\Z^d$ by a finite frequency set or index set, which we denote by $\I\subseteq\Z^d$, in order to get a good approximation of $f$ in terms of a partial sum
\begin{equation}\label{partial_sum}
	f(\x)\approx\sum_{\bfk\in\I} c_{\bfk}(f)\varphi_{\bfk}(\x).
\end{equation}
However, in practice the function $f$ is not known and we may fit the basis coefficients in \eqref{partial_sum} such that the resulting partial sum describes our data in an appropriate manner.
In case of binary classification, for instance, we may assume that our data is described by a function $f\in L_2(\mathbb D^d)$ but we only have information about the signum of $f$, i.e., we only know labels $y_j=\sign(f(\x_j))$ for the given feature vectors $\x_j$.
Thus, by solving \eqref{qp_unconstrained} we try to find a partial sum with coefficients $\bfh_\bfk$, $\bfk\in\I$, that satisfies
\begin{equation*}
	y_j\overset{!}{=}\sign\left(\sum_{\bfk\in\I} \hat f_\bfk\varphi_\bfk(\x_j)\right).
\end{equation*}
Note, that this does not imply that we are approximating the coefficients of the original function $f$, i.e., in general we will compute coefficients $\hat f_\bfk\not\approx c_\bfk(f)$.
In addition, also the index set $\I$ is in general not known in practice. In other words, we also have to determine a reasonable index set $\I$, which is not straight forward in high dimensions.

In this paper, we make use of the fact that we are able to compute the finite sum in \eqref{partial_sum} for arbitrary feature vectors $\x_j\in\D^d$, $j=1,2,\dots,M$ in an efficient manner in case of small dimensions $d$, see Section \ref{grouptransform}.  But first, let us take a closer look at the possible function systems and corresponding index sets that we will use in the upcoming sections concerning numerical experiments.

\subsubsection{The cosine basis}

The half-period cosine system with basis functions
\begin{equation*}
	\varphi^{\text{cos}}_\bfk(\x):=
	2^{\abs{\supp\,\bfk}}\prod_{i=1}^d \cos(2\pi k_ix_i), \quad  \x=\left(x_i\right)_{i=1}^d,
\end{equation*}
where $\abs{\supp\,\bfk}=|\{j\in\{1,2,\dots,d\}:k_j\neq 0\}|$ denotes the number of non-zero entries in $\bfk\in\N_0^d$, is a well-known basis of  $L_2([0,\sfrac12]^d)$, cf. \cite[Chapter 9]{Sutton2018}. It is obtained by applying the tensor product ansatz~\eqref{eq:tensorphi} to the univariate cosine basis.

To obtain finite sums, we truncate the series expansion using a finite frequency or rather index set. For a given multi degree $\bfN=\left(N_i\right)_{i=1}^d\in (2\N)^d$ we denote by
\begin{equation}\label{eq:IN_cos}
	\I_{\bfN}:=
	\{0,\dots,N_1-1\}\times\cdots\times\{0,\dots,N_d-1\}
\end{equation}
the corresponding frequency index set of cardinality $\prod_{i=1}^{d}N_i$.
Given such an index set $\I_{\bfN}$, we define the classifying function by
\begin{equation}\label{eq:sxi_cos}
	S(\X,\I_{\b N})f(\x):=\sum_{\b k\in \I_{\b N}}\fh_{\bfk}\varphi^{\text{cos}}_{\bfk}(\x),
\end{equation} 
where $\bfh=(\fh_\bfk)_{\bfk\in\I_{\b N}}\text{ solve \eqref{qp_unconstrained}}$. As explained above, the partial sum \eqref{eq:sxi_cos} or rather the included coefficients $\hat f_\bfk$ are computed by solving \eqref{qp_unconstrained} in order to given $S(\X,\I_{\b N})f(\x_j)=\sign(f(\x_j))$ for the given training data set, where it is our model assumption that such a function $f$, which describes the data in this way, exists.

Finally, we create the corresponding feature map via \eqref{feature_map} and define the associated feature matrix by
\begin{equation}\label{Phi_cos}
	\bfP^{\cos}(\X, \I_{\bfN}):=\left(\varphi^{\text{cos}}_{\bfk}(\x)\right)_{\x\in\X, \bfk\in\I_{\bfN}}\in\R^{M\times\abs{\I_{\bfN}}}.
\end{equation}
Note that a straightforward matrix-vector multiplication requires $\mathcal{O}(M\abs{\I_{\bfN}})$ arithmetical operations, which can be improved by applying the fast cosine transform for nonequispaced data (NFCT), see \cite[Chapter 7]{PlPoStTa18} or \cite{KeKuPo09}, yielding a complexity of $\mathcal O(|\I_\bfN|\log|\I_\bfN|+M)$.
With growing dimension $d$ the number of frequencies $|\I_\bfN|$ will grow exponentially in $d$ and computing an NFCT would be far too expensive. However, in case of small dimensions $d\lesssim 4$ such algorithms are still very efficient.

\subsubsection{The Chui-Wang basis}

Next we present a system consisting of  periodized, translated and dilated wavelets following \cite{lippert}, where the authors derived a lot of theory on hyperbolic wavelet regression, including considerations about the ANOVA decomposition of such functions. For more basic theory about wavelets we refer to the classical literature as for instance \cite{Chui, Daubechies, Walnut}. While the above discussed cosine system perfectly fits into the framework described at the beginning of Section~\ref{theory}, the notation becomes somewhat more complicated in case of wavelets.
That is, for example, that we have to deal with two different indices $\j$ and $\k$ now. Moreover, wavelet systems do in general not form an orthonormal basis. Nevertheless, the essential principles remain the same.

The Chui-Wang basis functions are obtained as follows.
First, we introduce the nested function spaces $V_j$ for $j\in\Z$ by $V_j=\text{span}\{\phi(2^j\cdot -k):k\in\Z\}$, where the function $\phi$ is called the scaling function. In the case of Chui-Wang wavelets, the scaling function is a cardinal B-spline $B_m:\R\to\R$ of order $m\in\N$, which are recursively defined via 
\begin{align*}
	B_1(x):=\begin{cases}
		1 ,&-\sfrac{1}{2}<x<\sfrac{1}{2}\\
		0,&\text{otherwise}
	\end{cases}
	\quad\text{and}\quad
	B_m(x):=\int_{x-\sfrac{1}{2}}^{x+\sfrac{1}{2}}B_{m-1}(y)\,\mathrm{d}y.
\end{align*}
As a consequence, we deduce
\begin{equation*}
	\cdots\subset V_{-1}\subset V_0\subset V_1\subset \cdots.
\end{equation*}
Moreover, the following properties, as $\bigcap_{j\in\Z}V_j=\{0\}$ and $\overline{\bigcup_{j\in\Z}V_j}=L_2(\R)$ hold true for the spaces $V_j$. Now, the wavelet space $W_j$ is defined as the orthogonal complement of a space $V_j$ in $V_{j+1}$, i.e., $V_{j+1}=V_j\oplus W_j$ for $j\in\N_0$, and, consequently, we obtain an orthogonal decomposition of $L_2(\R)$ by
\begin{equation*}
	V_{j+1}=V_j\oplus W_j,\quad j\in\N_0,
\end{equation*}
as well as an orthogonal decomposition of $L_2(\R)$ by
\begin{equation*}
	L_2(\R)=V_0\oplus\bigoplus_{j=0}^{\infty}W_j.
\end{equation*}
The wavelet spaces $W_j$, $j\in\N_0$, themselves are generated by a function $\varphi$, called the wavelet, in terms of $W_j=\text{span}\{\varphi_{j,k}:k\in\Z\}$, where
\begin{equation*}
	\varphi_{j,k}(x)=2^{\sfrac{j}{2}}\varphi(2^jx-k),\quad k\in\Z
\end{equation*}
are scaled and shifted versions of the wavelet function $\varphi$. Using a cardinal B-spline of order $m$ the corresponding wavelet function $\varphi$ is the Chui-Wang wavelet, \cite[Chapter 6]{Chui}. Note that in the special case $m=1$ we obtain the well-known Haar wavelets~\cite{Haar1910}. The Haar wavelet is one of the simplest wavelets, but it is not even continuous. However, this apparent weakness can become an advantage in the case one wants to approximate functions containing abrupt changes or jumps.

Now we define the one-periodic versions of the scaling function $\phi$ and the wavelet $\varphi$ by
\begin{equation*}
	\phi^{\text{per}}_{j,k}(x):=\sum_{\ell\in\Z}\phi_{j,k}(x+\ell) \quad\text{and}\quad \varphi^{\text{per}}_{j,k}(x):=\sum_{\ell\in\Z}\varphi_{j,k}(x+\ell).
\end{equation*}
Motivated by the theoretical results presented in~\cite{lippert}, we make use of the periodic Chui-Wang wavelet basis in this paper, although we deal with non-periodic examples later on. Especially, we make use of periodic Haar wavelets in our numerical examples since sharp jumps between classes have to be modeled. Please note that an appropriate extension of the wavelet basis to be used for non-periodic functions on the interval is by no means straightforward.

Note in addition that wavelets do in general not induce an orthonormal basis, but a Riesz-basis.
Thus, in order to compute the wavelet coefficients one has to make use of the so-called dual basis $\varphi^\ast_{j,k}$ fulfilling $\langle\varphi_{j,k}, \varphi^\ast_{i,\ell}\rangle=\delta_{i,j}\delta_{k,\ell}$.
Then, any function $f\in L_2(\T)$ can be decomposed as
\begin{align*}
	f &=\langle f,\varphi_{-1,0}^{\text{per}\ast}\rangle\varphi_{-1,0}^{\text{per}}+\sum_{j\geq 0}\sum_{k\in\Z}\langle f,\varphi^{\text{per}\ast}_{j,k}\rangle \varphi^{\text{per}}_{j,k} \\
	&=: c_{-1,0}(f) \varphi_{-1,0}^{\text{per}}+\sum_{j\geq 0}\sum_{k\in\Z} c_{j,k}(f) \varphi^{\text{per}}_{j,k}.
\end{align*}
In the special case of the Haar wavelet system we have $\varphi_{j,k}^\ast=\varphi_{j,k}$, i.e., we are again in the setting of an orthonormal system.

Applying the tensor-product approach \eqref{eq:tensorphi} to the wavelet setting, we analogously define the multi-variate periodic wavelets via
\begin{equation*}
	\varphi^{\chui, \text{per}}_{\j,\k}(\x)=\prod_{i=1}^d \varphi_{j_i,k_i}^{\chui,\text{per}}(x_i)
\end{equation*}
with multi-indices $\j=(j_i)_{i=1}^d$, $j_i\in\{-1,0,2,\dots\}$, and $\k\in\mathcal{K}_{\j}$, where
\begin{align*}
	\mathcal{K}_{\j}=\bigtimes_{i=1}^d\begin{cases}
		\{0\},&\quad j_i=-1\\
		\{0,1,\dots,2^{j_i}-1\},&\quad j_i\geq 0.
	\end{cases}
\end{align*}
In order to truncate the series appropriately we define the finite index set 
\begin{equation}\label{eq:JN}
	\mathcal{J}_{\b N}=\left\{\j\in\Z^d:\j\geq -\mathbb{1},\sum_{i,j_i\geq 0}\frac{j_i}{N_i}\leq1\right\}
\end{equation}
with multi-degree $\bfN=\left(N_i\right)_{i=1}^d\in\N^d$ and introduce the partial sum
\begin{equation}\label{eq:sxi_chui}
	S(\X, \mathcal{J}_{\b N})f(\x):=\sum_{\j\in\mathcal{J}_{\b N}}\sum_{\bfk\in\mathcal{K}_{\j}}\hat f_{\j,\k} \varphi^{\chui, \text{per}}_{\j,\k}(\x),
\end{equation}
where, again, the unknown coefficients $\bfh=(\fh_{\j,\bfk})_{\substack{\j\in\mathcal{J}_{\b N}\\ \bfk\in\mathcal{K}_{\j}}}$ have to be learned via \eqref{qp_unconstrained} based on the set of feature vectors $\mathcal X$ in order to give $S(\X,\mathcal{J}_{\b N})f(\x_j)=\sign(f(\x_j))$ for a presumed and unknown function $f$.

\begin{figure}[ht]
	\begin{center}
		\begin{tikzpicture}[ xscale=0.75,yscale=0.75,cross/.style={draw, cross out,
				minimum size=2*(#1-1pt), inner sep=0pt, outer sep=0pt}, x=.5cm, y=.5cm,fill opacity=1, draw opacity = 1]
			\fill[fill={rgb:black,1;white,9}] (2,2) -- (2,8) -- (4,8) -- (4,2);
			\fill[fill={rgb:black,1;white,9}] (0,0) -- (0,16) -- (2,16) -- (2,0);
			\fill[fill={rgb:black,1;white,9}] (2,0) -- (2,2) -- (16,2) -- (16,0);
			\fill[fill={rgb:black,1;white,9}] (4,2) -- (4,4) -- (8,4) -- (8,2);
			\draw [color=gray!50]  [step=5mm] (0,0) grid (16,16);
			\draw[-, thick] (0,0) -- (16,0);
			\draw[-, thick] (1,0) -- (1,16);
			\draw[-, thick] (2,0) -- (2,16);
			\draw[-, thick] (4,0) -- (4,16);
			\draw[-, thick] (8,0) -- (8,16);
			\draw[-, thick] (0,16) -- (16,16);
			\draw[-, thick] (16,0) -- (16,16);
			\draw[-,thick] (0,0) -- (0,16) ;
			\draw[-,thick] (0,1) -- (16,1) ;
			\draw[-,thick] (0,2) -- (16,2) ;
			\draw[-,thick] (0,4) -- (16,4) ;
			\draw[-,thick] (0,8) -- (16,8) ;
			\draw(0,8) node[left] {$j_2\quad$};
			\draw(8,0)node[below]{\rotatebox{90}{\rotatebox{-90}{$j_1$}\quad}};
			\draw(0.5,0)node[below]{\small $-1$};
			\draw(1.5,0)node[below]{\small $0$};
			\draw(3,0)node[below]{\small $1$};
			\draw(6,0)node[below]{\small $2$};
			\draw(12,0)node[below]{\small $3$};
			\draw(0,0.5)node[left]{\small $-1$};
			\draw(0,1.5)node[left]{\small $0$};
			\draw(0,3)node[left]{\small $1$};
			\draw(0,6)node[left]{\small $2$};
			\draw(0,12)node[left]{\small $3$};     
		\end{tikzpicture}
		\caption{Illustration of $2$-dimensional indices $\k=(k_1,k_2)$, represented by the gray squares, where $\j=(j_1,j_2)\in \mathcal{J}_{(3,3)}$ and $\k\in\mathcal{K}_{\j}$ for each $\j$. This graphic is taken from~\cite{lippert}.}  
		\label{fig:hyperbolic set}
	\end{center} 
\end{figure}
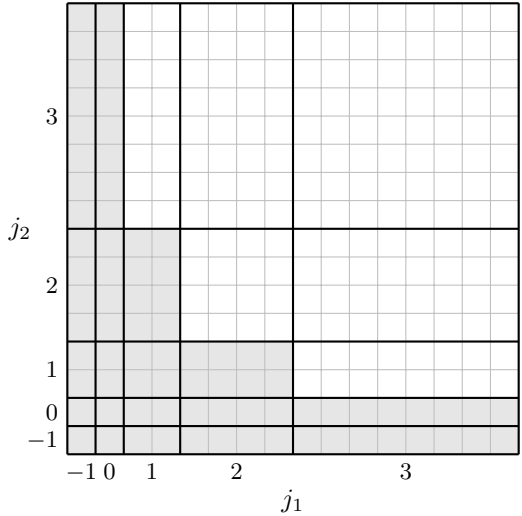

From \eqref{eq:JN} we see that we do not use the full tensor product in the multi-variate setting, but rather a hyperbolic subset of it, see Figure~\ref{fig:hyperbolic set}, motivated by the theoretical results presented in \cite{lippert}. In the special case that the same bandwidth $N$ is applied in all dimensions, i.e., $\bfN=(N,N,\dots,N)^\top\in\N^d$, the last condition in \eqref{eq:JN} becomes
\begin{equation*}
	\sum_{i, j_i\geq0} j_i\leq N.
\end{equation*}

\noindent Building up the resulting feature map via \eqref{feature_map} and defining the resulting feature matrix gives
\begin{equation}\label{Phi_chui}
	\bfP^{\chui}(\X, \mathcal{J}_{\b N}):=\left(\varphi^{\chui, \text{per}}_{\j,\k}(\x)\right)_{\x\in\X,\substack{\j\in\mathcal{J}_{\b N}\\ \k\in\mathcal{K}_{\j}}}\in\R^{M\times\sum_{\j\in\mathcal{J}_{\b N}}\abs{\mathcal{K}_{\j}}}.
\end{equation}
Note that the wavelets are localized functions and in case of the Haar system even compactly supported, which means that the matrix defined above is sparse. Thus, one can compute matrix vector products involving this matrix in an efficient manner, as long as the dimension $d$ is small.

\subsection{ANOVA decomposition and effective dimensions} \label{sec:fourier_anova}

The main idea of the classical ANOVA decomposition of functions is to decompose a $d$-variate function in $2^d$ so-called ANOVA terms $f_\u$, where each term represents a subset of coordinate indices $\u\subseteq[d]:=\{1,2,\dots, d\}$. Each single term depends only on the variables in the corresponding subset and the number of these variables $|\u|$ is called the order of the ANOVA term. At this point we would like to refer to the references, regarding the classical ANOVA decomposition and its use for the function approximation, that were already mentioned at the beginning of this chapter. In order to introduce the ANOVA decomposition of a function $f\in L_2(\D^d)$ we define the projection operators
\begin{equation*}
	\text{P}_{\u}f(\x)=\frac1{\tau^{d-|\u|}}\int_{\D^{d-\abs{\u}}}f(\x)\,\mathrm{d}\x_{\u^c}
\end{equation*}
for a subset of coordinate indices $\u$ and its complementary set $\u^c:=[d] \setminus\u$. Then, the ANOVA terms can be defined recursively via
\begin{equation}\label{ANOVAterm}
	f_{\u}(\x):=\text{P}_{\u}f(\x)-\sum_{\v\subsetneq \u}f_{\v}(\x),
\end{equation}
where $\x_{\u}:=(x_i)_{i\in\u}\in\D^{|\u|}$ denotes the restriction of $\x$ to the variables present in the subset $\u$. In case of the cosine basis we obtain \cite{PoSc19a}
\begin{equation*}
	f_{\u}(\x) = \sum\limits_{\substack{\bfk\in\N_0^d\\ \text{supp}\,\bfk=\u}}c_\bfk(f)\varphi^{\text{cos}}_\bfk(\x),
\end{equation*}
where $\supp \bfk:=\{s\in[d]:k_s\neq 0\}$ is the set of non-zero dimensions in a multi frequency $\bfk$. Thus, the ANOVA decomposition introduces a disjoint decomposition of the set of basis coefficients or rather frequencies, i.e., the ANOVA terms are orthogonal to each other with respect to the $L_2$ inner product. An ANOVA term $f_\u$ includes exactly those frequencies $\bfk$ that are non-zero in the dimensions present in $\u$ and equal to zero otherwise.

In a very similar fashion, we conclude for the wavelet basis \cite{lippert}
\begin{equation*}
	f_{\u}(\x) = \sum\limits_{\substack{\j\in\Z^d\\ \j_\u>-\mathbb{1}, \j_{\u^\text{c}}=-\mathbb{1}}} \sum_{\bfk\in\mathcal{K}_{\j}} c_{\j,\bfk}(f)\varphi^{\chui,\text{\text{per}}}_{\j,\bfk}(\x).
\end{equation*}
The function $f$ can then be uniquely decomposed as
\begin{equation}\label{anova decompos}
	f(\x)=f_{\emptyset}+\sum_{i=1}^{d}f_{\{i\}}(x_i)+\sum_{i=1}^{d-1}\sum_{j=i+1}^{d}f_{\{i,j\}}(\x_{\{i,j\}})+\cdots+f_{[d]}(\x)=\sum_{\u\subseteq [d]}f_{\u}(\x)
\end{equation}
into $\abs{\mathcal{P}([d])}=2^d$ ANOVA terms.
The relation to the series expansion \eqref{ANOVAterm} have first been proven in \cite{PoSc19a}. But we still have the curse of dimensionality in \eqref{anova decompos}, since the number of all ANOVA terms grows exponentially with the dimension $d$ and we need to reduce this in an appropriate manner. To this end, we make use of the truncated ANOVA decomposition by taking into account only variable interactions involving only a small number of dimensions. This is motivated by the well-known sparsity of effects stating that many phenomena are well described by few low-dimensional interactions, see \cite{CaMoOw97,KuSlWaWo09,DeVore, PoSc21}. 

As an example, for small $d_s\in[d]$, $d_s\leq 3$, we could only make use of subsets $\u$
that contain $d_s$ or less variables, i.e., the restricted subset of ANOVA terms is 
\begin{equation}
	U_{d_s}:=\{\u\subseteq [d]:\abs{\u}\leq d_s\}
\end{equation}
where $d_s$ is called superposition threshold or rather the superposition dimension. The cardinality of the set $U_{d_s}$ is given by $1+\binom{d}{1}+\dots+\binom{d}{d_s}\sim d^{d_s}$ and, thus, only grows polynomially in $d$ for fixed superposition dimension $d_s$.

Moreover, using the ANOVA decomposition we are able to obtain information about the importance of each term $f_{\u}(\x)$ with respect to the function $f$.  In \eqref{ANOVAterm} we can see that the ANOVA terms have a nice interpretation in terms of the basis coefficients. The relative importance with respect to $f$ can be measured in terms of global sensitivity indices (GSIs) or rather Sobol indices \cite{So01,So90}, which measure the contribution of the single subsets $\u$ to the overall variance of the function $f$. Combining \eqref{ANOVAterm} and Parseval's identity gives
\begin{equation}\label{gsi}
	\varrho(\u,f):=
	\begin{cases}
		\displaystyle\frac{\sum_{\mathrm{supp}(\b k)=\u} |c_{\bfk}|^2}{\sum_{\bfk\neq\mathbf 0} |c_{\bfk}|^2}
		&: \text{cosine basis}, \\
		\displaystyle\frac{\sum_{\j_\u>-\mathbb 1,\j_{\uc}=-\mathbb 1} \sum_{\bfk\in\mathcal{K}_{\j}} |c_{\j,\bfk}|^2}{\sum_{\j\neq-\mathbb 1}\sum_{\bfk\in\mathcal{K}_{\j}} |c_{\j,\bfk}|^2} &: \text{wavelet basis},
	\end{cases}
\end{equation}
for which we easily conclude $\varrho(\u,f)\in[0,1]$. If $\varrho(\u,f)\in[0,1]$ is large then the ANOVA term $f_\u$ significantly contributes to the overall variance of the function $f$, i.e., making changes in that dimensions might cause a significant change of the function value.
In contrast, variables or subsets $\u$ might be negligible when $\varrho(\u,f)\approx0$. This makes especially sense if one searches for functions for regression or classification with small variance.

For a computed classifying function of the form \eqref{eq:sxi_cos} or rather \eqref{eq:sxi_chui}, depending on given training data set $\{(\x_1,y_1),(\x_2,y_2),\dots,(\x_M,y_M)\}$,
we use the fitted coefficients $\fh_\bfk$ or rather $\fh_{\j,\bfk}$ to compute the corresponding global sensitivity indices via \eqref{gsi}.
Furthermore, we are able to reduce the number of ANOVA terms such that we select only indices for more valuable ones. 

Based on the computed GSIs, we define an active set of ANOVA terms
\begin{equation}\label{eq:ueps1}
	U^{\varepsilon}:=\emptyset\cup\left\{\u\in U:\varrho(\u,S(\X, \I_{\b N}))>\varepsilon\right\}\subseteq U
\end{equation}
or rather
\begin{equation}\label{eq:ueps2}
	U^{\varepsilon}:=\emptyset\cup\left\{\u\in U:\varrho(\u,S(\X, \mathcal{J}_{\b N}))>\varepsilon\right\}\subseteq U,
\end{equation}
for a given tolerance $\varepsilon>0$, in order to identify ANOVA terms with a small contribution to the overall variance of the computed classifying function.
Having identified a suitable active set $U^\varepsilon$, we may re-build the feature map \eqref{feature_map} and solve \eqref{qp_unconstrained} again by only using the necessary groups of variables, where we probably now have a smaller number of coefficients to be calculated.
Moreover, this will be hopefully a beneficial way to improve our prediction of the unknown classifying function.
The same procedure has already been successfully applied in case of regression problems, see \cite{PoSc21}.
In order to be able to implement these theoretical considerations in an efficient manner, we introduce the concept of Grouped Transformations in the following  section.

\subsection{Grouped transformations}\label{grouptransform}

The main  idea of grouped transformations is to omit high-dimensional interactions and consider the sum of low-dimensional transformations, which are efficient, see \cite{BaPoSc}.
Later in our numerical experiments we make use of the \textit{GroupedTransforms.jl} package~\cite{gitGTjl}, with which the below described grouped transformations can be computed efficiently. In case of the cosine basis, the corresponding matrix-vector products are computed based on the NFCT algorithm~\cite{nfft3,gitNFFT3jl}. In case of wavelets, the considered matrices are sparse, allowing for an efficient computation of the matrix-vector product, cf.~\cite{lippert,gitGTjl}.

For a given subset of ANOVA terms $U\subseteq\mathcal{P}([d])$ we define a grouped index set as a disjoint union 
\begin{equation*}
	\I_{\bfN}(U)=\bigcup_{\u\in U}\I_{\b N^{\u}}^{\u, d}
\end{equation*}
where $\bfN=(\b N^{\u})_{\u\in U}$ is a $\abs{U}$-tuple with entries $\b N^{\u}\in(2\N)^{\abs{\u}}$
and 
$$\I_{\b N^{\u}}^{\u, d}=\{\k\in\N_0^d:\supp \k=\u, (\k_{\u})_i\in\I_{N^{\u}_i}, i=1,2,\dots|\u|\}$$
for $\u\in U$ are the corresponding frequency index sets. As an example, in a $4$-dimensional setting let $\u:=\{2,3\}$ and $N^\u:=(6,24)\in (2\N)^2$.  Then, the index set $\I_{\b N^{\u}}^{\u, d}$ contains indices $\k=(0,k_2,k_3,0)$ with $k_2\in\{1,2,\dots,5\}$ and $k_3\in\{1,2,3\}$, see \eqref{eq:IN_cos}.

For the Chui-Wang basis  some adjustments are necessary. Analogously to $\I_{\bfN}(U)$ we define the index sets
\begin{equation*}
	\mathcal{J}_{\b N}(U)=\bigcup_{\u\in U}	\mathcal{J}_{\b N^{\u}}^{\u,d}
\end{equation*}
with tuple $\bfN=(\b N^{\u})_{\u\in U}$, $\b N^{\u}\in\N^{\abs{\u}}$ and for $\u\in U$ we get
\begin{equation}
	\mathcal{J}_{\b N^{\u}}^{\u,d}=\left\{\j\in\Z^d:\j_\u>-\mathbb{1}, \j_{\uc}=-\mathbb{1}, \sum_{i,(j_{\u})_i\geq 0}\frac{(j_{\u})_i}{N^{\u}_i}\leq1\right\}.
\end{equation}

In order to truncate the ANOVA decomposition of $f$, we assume that the function $f$ has low superposition dimension $d_s\ll d$ and define $U_{d_s}:=\{\u\subseteq[d]: |\u|\leq d_s\}$ in the following.

Making use of the grouped index sets introduced above, the feature map \eqref{Phi_cos} or rather \eqref{Phi_chui} has a block structure, i.e., 
\begin{equation*}
	\bfP^{\cos}(\X, \I_{\bfN}(U))=\left(\bfP^{\cos}(\X, \I_{\b N^{\u_1}}^{\u_1,d})\quad \bfP^{\cos}(\X, \I_{\b N^{\u_2}}^{\u_2,d})\quad\cdots\quad \bfP^{\cos}(\X, \I_{\b N^{\u_{\abs{U}}}}^{\u_{\abs{U}},d})\right)
\end{equation*}
and
\begin{equation*}
	\bfP^{\chui}(\X, \mathcal{J}_{\b N}(U))= \left(\bfP^{\chui}(\X, \mathcal{J}_{\b N^{\u_1}}^{\u_1,d})\quad \bfP^{\chui}(\X, \mathcal{J}_{\b N^{\u_2}}^{\u_2,d})\quad\cdots\quad \bfP^{\chui}(\X, \mathcal{J}_{\b N^{\u_{|U|}}}^{\u_{\abs{U}},d})\right).
\end{equation*}
In case of the cosine system, for instance, we see that a matrix-vector product $\bfP^{\cos}(\X, \I_{\bfN}(U))\bfh$ is simply a sum of low-dimensional transformations
\begin{equation}
	\bfP^{\cos}(\X, \I_{\bfN}(U))\bfh=\sum_{\u\subseteq U}\bfP^{\cos}\left(\X, \I_{\b N^{\u}}^{\u,d}\right)\bfh(\u),
\end{equation}
for which the computational effort no longer grows exponentially for growing $d$ and fixed group structure or superposition dimension $d_s$. By $\bfh(\u)$ we mean the vector of basis coefficients restricted to indices $\bfk$ belonging to the ANOVA term $f_\u$.
The same also applies to the wavelet case, see \cite[Section 4.1]{lippert}.
Note that the number of ANOVA terms still grows polynomially in $d$ with order $d_s$.
So if the number of dimensions or rather features $d$ is extremely large, we will also reach a limit at a certain point. As the matrix multiplications for the different ANOVA terms are obviously independent of each other, the calculations can of course be carried out in parallel, which is what the software is also designed for.
In our numerical examples, however, we will limit ourselves to moderately large dimensions $d\leq28$.
Note, in addition, that fast transformations only pay off if the number of basis functions for each ANOVA term is very large.
So if the number $M$ of data points is also not too large, it is best to carry out the matrix-vector multiplications directly, as denoted above.
If the number of data points becomes too large and the matrices can no longer be stored, one should switch to the discrete transformations (NDCT) or, in the case of many basis functions, to the fast transformations (NFCT).

\begin{figure}[ht!]
	\centering
	\begin{tikzpicture}
		\begin{axis}[
			domain=3:50, 
			samples=100,
			xlabel=$d$, 
			ylabel=$|\mathcal I|$,
			ymode=log,
			width=0.4\linewidth,
			ytick={1,1e6,1e12,1e18,1e24,1e30},
			ymin=1,
			legend entries={$d_s=1$, $d_s=2$, $d_s=3$, $d_s=d$},
			legend style={legend pos=north west}
			]
			\addplot[red,dashed] {1+x*3};
			\addplot[green!50!black,densely dotted] {1+x*3+x*(x-1)/2*3^2};
			\addplot[blue,dashdotted] {1+x*3+x*(x-1)/2*3^2+x*(x-1)*(x-2)/6*3^3};
			\addplot[black,solid] {4^x};
		\end{axis}
	\end{tikzpicture}
	\caption{Cardinality of the index set $\I$ depending on the dimension $d$ for different superposition dimensions $d_s\in\{1,2,3,d\}$.
		We consider the cosine basis approach, for which we restrict the space of possible multi-frequencies $\k\in\N_0^d$ to $\{0,1,2,3\}^d$.}
	\label{fig:cardinality_I_cosine}
\end{figure}
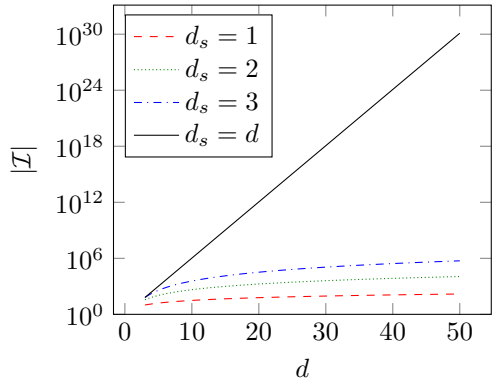

To conclude this section, we want to illustrate the dependence of the number of basis coefficients for increasing dimension $d$ for different superposition dimensions $d_s$.
Note that this number corresponds to the number of columns of the feature matrix, where we consider the cosine basis as a concrete example with frequencies $\k\in\{0,1,2,3\}^d$.
Depending on the number of features $d$ and the applied superposition dimension $d_s$, the cardinality of the index set $\I$ grows as $\sim d^{d_s}$.
In Figure~\ref{fig:cardinality_I_cosine} we plot the resulting number of coefficients in the index set $\I$ for dimensions $d\in\{3,\dots,50\}$.
It can be seen that the full grid $\{0,1,2,3\}^d$ with $4^d$ many frequencies quickly reaches sizes that are no longer manageable.

\section{Optimization Algorithms}\label{sec:algorithms}

In this chapter we focus on the problem of estimating the model parameters $\bfh$ for the given training data set $\{(\x_j,y_j)\}_{j=1,2,\dots,M}$, i.e., solving the optimization problem \eqref{qp_unconstrained}. 
In order to obtain a classifier we need to construct some function $f$, such that $\sign(f(\x_j))=y_j$ for as many inputs $\x_j$, $j=1,2,\dots,M$, as possible.
In this chapter we write $\b\Phi$ for given grouped transform, which are the feature matrices $\bfP^{\cos}(\X, \I_{\bfN}(U))$ or $\bfP^{\chui}(\X, \mathcal{J}_{\b N}(U))$, given in \eqref{Phi_cos} and \eqref{Phi_chui}, where $U$ is a given subset of ANOVA terms and $\X$ is a given set of instances.
The related fitted coefficients $\bfh^{\cos}$ and $\bfh^{\chui}$ we  denote in general by $\bfh$ with index set $\I$.

The common way to solve the SVM problem \eqref{qp_unconstrained} is solving the corresponding dual optimization problem based on kernel functions. However, the kernel matrices become too large and badly conditioned for huge amount of data. Well, if we use the approach, what we built in the previous chapter, we are able to solve our SVMs in primal form using grouped transformations as feature maps. 

In this chapter we present two algorithms, which can be used for solving the optimization problem \eqref{qp_unconstrained} with $\ell_2$-norm regularization, see Section \ref{L2}, and $\ell_1$-norm regularization in Section \ref{L1}.

\subsection{$\ell_2$-norm regularization}\label{L2}

First we discuss an unconstrained optimization problem of the form
\begin{equation}\label{a}
	P(\bfh):=\lambda\norm{\bfh}{2}^2+\frac{1}{M}\sum_{j=1}^M\left(\max\left\{0,1-y_j\left\langle\bfh,\Phi(\x_j)\right\rangle\right\}\right)^2\rightarrow \min_{\bfh}, \quad \bfh\in \R^{|\I|}
\end{equation}
using the regularization term $\|\bfh\|_2^2$, which enforces the small variance of the classifying function. To solve this problem numerically, we apply a gradient descent method, which uses the negative gradient direction $d^k:=-\nabla P(\bfh^{(k)})$ in each iteration and improves the objective in that direction.
In order to compute the gradient of $P(\bfh)$ with respect to the vector of basis coefficients $\bfh$, we first introduce the following function $h:\R\to\R$ and its derivative, given by
\begin{equation*}
	h(t):= \max\{0,t\}^2, \quad h'(t)=\begin{cases}2t&:t\geq0,\\
		0&:t<0.
	\end{cases}
\end{equation*}
Consequently we have
\begin{equation}\label{eq:gradient}
	\nabla P(\bfh)=2\lambda\bfh-\frac{1}{M}\sum_{j=1}^{M}y_j\cdot \Phi(\x_j)\cdot h'(1-y_j\langle\bfh,\Phi(\x_j)\rangle).
\end{equation}
Thinking in terms of matrix-vector multiplications, the approximation of the objective function~\eqref{a} and its gradient~\eqref{eq:gradient} is realized as follows.
The matrix-vector product $ \b \Phi\bfh$ is a vector of length $M$ containing the vector products $\langle\bfh,\Phi(\x_j)\rangle$.
Due to the ANOVA structure of the considered index sets, the matrix-vector product $\b \Phi\bfh$ can be approximated efficiently in terms of a grouped transform, as explained above. In case of the cosine system, for instance, a sum of NFCTs in small dimensions is computed.
Having computed $\b \Phi\bfh$, we just need pointwise operations in order to evaluate $P(\bfh)=\lambda \|\bfh\|_2^2+\frac 1M \texttt{sum}(\texttt{max.}(\mathbf{0},\mathbb{1}-\y\texttt{.*}(\b \Phi\bfh)))$, where a dot indicates a pointwise operation.

In order to compute $\nabla P(\bfh)$ we also need to multiply with the transposed matrix $\b \Phi^\top$ in an efficient way.
This is possible in a very similar fashion. In case of the cosine basis, the resulting algorithm is known as transposed NFCT, see \cite[Chapter 7]{PlPoStTa18}.
Finally, we obtain
$\nabla P(\bfh)=2\lambda\bfh-\frac1M \texttt{sum}( (\b \Phi^\top\y) \texttt{.*} h'\texttt{.}(\mathbb{1}-\y \texttt{.*} (\b \Phi\bfh)))$.

We summarize the gradient descent method in Algorithm~\ref{alg:2}, where the choice of the step size is realized by the well-known Armijo backtracking strategy~\cite{Armijo}, see Algorithm \ref{alg:3}. Using a stopping criterion, which ensures that the gradient $\nabla P(\bfh)$ is sufficiently small, we can also stop earlier.

\begin{algorithm}[ht!]
	\vspace{2mm}
	\begin{tabular}{ l l l }
		\textbf{Input:} & set of instance-label pairs $\{(\x_j,y_j): j=1,2,\dots,M\}\in \X \times\{-1, +1\}$, $\X \subseteq \D^d$, grouped transform $\b\Phi$,\\
		& initial guess $\bfh^{(0)}$, regularization parameter $\lambda>0$, maximal number of iterations $K\in \N$\\
	\end{tabular}
	\begin{algorithmic}[1]
		\For{$k=1,2,\dots,K$}
		\State{$d^k=-\nabla P(\bfh^{(k)})$}
		\If{$\|\nabla P(\bfh^{(k)})\|_2< 10^{-8}$}
		\State{\textbf{break}}
		\EndIf
		\State{compute a step size $s_k$} via Algorithm \ref{alg:3}
		\State{$\bfh^{(k+1)}\longleftarrow \bfh^{(k)}+s_k d^k$}
		\EndFor
	\end{algorithmic}
	\begin{tabular}{ l l l }
		\textbf{Output:} &$  \bfh^{(K)}$& minimizer of \eqref{a}
	\end{tabular}
	\caption{Gradient descent algorithm for the solution of \eqref{a}}
	\label{alg:2}
\end{algorithm}

\begin{algorithm}[ht!]
	\vspace{2mm}
	\begin{tabular}{ l l l }
		\textbf{Input:} &current iterate $\bfh^{(k)}$, search direction $d^k$, initial step size $s_k^{(0)}$, $\sigma\in(0,1), \xi\in(0,1)$,\\
		& precomputed function value $P(\bfh^{(k)})$ and gradient $\nabla P(\bfh^{(k)})$
	\end{tabular}
	\begin{algorithmic}[1]
		\State{set $n:=0$}
		\While{$P(\bfh^{(k)}+s_k^{(n)}d^k)>P(\bfh^{(k)})+\sigma s_k^{(n)}\norm{\nabla P(\bfh^{(k)})}{2}^2$}
		\State{$s_k^{(n+1)}\longleftarrow\xi s_k^{(n)}$}
		\State{$n\longleftarrow n+1$}
		\EndWhile
	\end{algorithmic}
	\begin{tabular}{ l l l }
		\textbf{Output:} &$s^{(n)}_k$&step size
	\end{tabular}
	\caption{Armijo backtracking rule}
	\label{alg:3}
\end{algorithm}

\subsection{$\ell_1$-norm regularization}\label{L1}

In this section we discuss the optimization method for the following unconstrained convex optimization problem 
\begin{equation}\label{b}
	P(\bfh):=\lambda\norm{\bfh}{1}+\frac{1}{M}\sum_{j=1}^M\left(\max\left\{0,1-y_j\left\langle\bfh, \Phi(\x_j)\right\rangle\right\}\right)^2\rightarrow \min_{\bfh}, \quad \bfh\in \R^{|\I|},
\end{equation}
for given $\b\Phi \in \R^{M\times{|\I|}}$, which will either be equal to $\bfP^{\cos}(\X, \I_{\bfN}(U))$, as defined in \eqref{Phi_cos}, or $\bfP^{\chui}(\X, \mathcal{J}_{\b N}(U))$, cf.~\eqref{Phi_chui}.
Solving such a problem is more challenging than the one discussed in the previous section, because $\|\bfh\|_1$ is non-smooth, whereas the second part
\begin{equation*}
	q(\bfh):= \frac{1}{M}\sum_{j=1}^M\left(\max\left\{0,1-y_j\left\langle\bfh, \Phi(\x_j)\right\rangle\right\}\right)^2
\end{equation*}
has a Lipschitz continuous gradient with respect to $\bfh$.
In order to solve \eqref{b} we will make use of the well-known fast iterative shrinkage thresholding algorithm (FISTA), see \cite{Beck1}, which became popular during the past decades.
In addition, we would like to mention that FISTA has also been used in \cite{BaPoSc} for high-dimensional explainable ANOVA approximation, where regression problems have been considered using a group Lasso regularization term.

Let us briefly review the basic approximation model, which we are going to use. For any $L>0$, consider the following quadratic approximation of $P(\bfh)$ at a given point $\hat{\b h}$
\begin{align*}
	Q_L(\bfh,\hat{\b h}):=&
	\;q(\hat{\b h})+\left\langle\bfh-\hat{\b h},
	\nabla q(\hat{\b h})
	\right\rangle
	+\frac{L}{2}\norm{\bfh-\hat{\b h}}{2}^2+\lambda\norm{\bfh}{1}
\end{align*}
which admits a unique minimizer 
\begin{equation}\label{eq:p_L}
	p_L(\hat{\b h})=\frac{1}{L}\argmin_{\bfh}\left\{L\lambda\norm{\bfh}{1}+\frac{1}{2}\norm{\bfh-\underbrace{
			\left(
			\hat{\b h}-\frac{1}{L}
			\nabla
			q(\hat{\b h})
			\right)
		}_{=:\b a}}{2}^2\right\},
\end{equation}
where $L$ plays the role of a step size. Using the well-known definition of the proximal operator and its scaling properties, see \cite[Chapter 6]{Beck2}, \cite{Parikh}, given by
\begin{equation*}
	\prox{_{L\lambda\norm{\cdot}{1}}}(\b a)=\argmin_{\bfh}\left(L\lambda\norm{\bfh}{1}+\frac{1}{2}\norm{\bfh -\b a}{2}^2\right),
\end{equation*}
we get for $i=1,2,\dots,|\bfh|$ the following solution of \eqref{eq:p_L}
\begin{equation}
	p_L(\hat{h}_i)=\frac{1}{L}\prox{_{L\lambda\norm{\cdot}{1}}}(a_i)=\sign(a_i)\max(0,|a_i|-\L\lambda).
\end{equation}
We summarize the FISTA algorithm with backtracking step size rule in Algorithm~\ref{fista}.
\begin{algorithm}[ht!]
	\vspace{2mm}
	\begin{tabular}{ l l l }
		\textbf{Input:} &set of instance-label pairs $\{(\x_j,y_j): j=1,2,\dots,M\}\in \X \times\{-1, +1\}$, $\X \subseteq \D^d$, grouped transform $\b\Phi$,\\ 
		&initial guess $\bfh^{(0)}$, regularization parameter $\lambda>0$, maximal number of iterations $K\in \N$, $L_0>0$ and $\theta>1$
	\end{tabular}
	\begin{algorithmic}[1]
		\State{$\hat{\b h}^{(1)}\longleftarrow\bfh^{(0)}, t_1\longleftarrow1$}
		\For{$k=1,2,\dots,K$}
		\State{$L_k\longleftarrow L_{k-1}$}
		\State{$\bfh^{(k)}\longleftarrow p_{L_k}(\hat{\b h}^{(k)})$} 
		\If{$|P(\bfh^{(k)})-P(\bfh^{(k-1)})|< 10^{-8}$ \textbf{or} $ \|\bfh^{(k-1)}-\bfh^{(k)}\|_2< 10^{-8}$}
		\State{\textbf{break}}
		\EndIf
		\While{$q(\bfh^{(k)}) < q(\hat{\b h}^{(k)})
			+\left\langle \bfh^{(k)}-\hat{\b h}^{(k)}, \nabla q(\hat{\b h}^{(k)})\right\rangle+\frac{L_k}{2}\norm{\bfh^{(k)}-\hat{\b h}^{(k)}}{2}^2$}
		\State{$L_k\longleftarrow\theta L_k$}
		\State{$\bfh^{(k)}\longleftarrow p_{L_k}(\hat{\b h}^{(k)})$}
		\EndWhile
		\State{$t_{k+1}\longleftarrow (1+\sqrt{1+4t_k^2})/2$}
		\State{$ \b {\hat{h}}^{(k+1)}\longleftarrow \bfh^{(k)}+(t_{k+1})(\bfh^{(k)}-\bfh^{(k-1)})$}
		\EndFor
	\end{algorithmic}
	\begin{tabular}{ l l l }
		\textbf{Output:} &$  \bfh^{(K)}$&minimizer of \eqref{b}
	\end{tabular}
	\caption{FISTA with backtracking for the solution of \eqref{b}}
	\label{fista}
\end{algorithm}

\section{Numerical Experiments}\label{sec:examples}

In this chapter we present the results of some numerical experiments. 
We begin with self-generated toy examples, see Section~\ref{sec:toy}, including one-dimensional, six-dimensional and ten-dimensional examples. 
We also test our method on the freely available data sets in Section~\ref{sec:datasets} and compare the results with those from the literature.

In order to measure the performance of our method we define the classification accuracy (CA) and use the following notation.
If for a given feature vector $\x_j$ the corresponding label is positive and it is classified as positive, it will be counted as true positive (TP); if it is classified as negative, it is called as false negative (FN). The similar notation if for the given feature vector $\x_j$ the corresponding label is negative and it is classified as negative, it is counted as true negative (TN) otherwise false negative (FN). In order to calculate the CA, we only need to compare the number of correct predictions among the total number of cases examined
\begin{equation*}
	\text{CA} = \frac{\text{TP}+\text{TN}}{\text{TP}+\text{TN}+\text{FP}+\text{FN}}.
\end{equation*}

Due to the unbalanced nature of the data sets, which is common in most real-world applications, it is more convenient to use different measures than the CA, such as the well-known AUC, which stands for “area under the curve”.
The AUC measures the quality of a binary classifier in terms of the receiver operating characteristics (ROC) curve with a value between $0.5$ and $1.0$, see~\cite{Therrien1989} and~\cite{Fukunada1990} for instance.
The ROC curve plots the relationship between true positive rate (TPR) and false positive rate (FPR)
\begin{equation*}
	\text{TPR} = \frac{\text{TP}}{\text{TP}+\text{FN}}\quad \text{and}\quad\text{FPR} = \frac{\text{FP}}{\text{FP}+\text{TN}}
\end{equation*}
at all classification thresholds. The use of the ROC curve as a performance measure for machine learning algorithms was explicitly discussed in \cite{Bradley97}. 
Otherwise, it is common to avoid problems with the imbalance of data by using so-called resampling techniques~\cite{Marques2013}, as it was done in~\cite{WaPeSt2023}.

We have implemented our algorithms in the programming language Julia as a part of the publicly available Julia package \textit{ANOVAapprox.jl}~\cite{gitANOVAjl}, which itself makes use of the~\textit{GroupedTransforms.jl} Julia package~\cite{gitGTjl}. 
All data sets from Table~\ref{table:1} are freely available from the UCI Machine Learning Repository\footnote{\url{https://archive.ics.uci.edu/}} or can also be easily downloaded by using the~\textit{LIBSVM.jl}~\cite{libsvm} package and its extension~\textit{LIBSVMdata.jl}, which is a summary of many classification, regression, multi-label and string data sets stored in the LIBSVM format.
We make use of the second variant, because that way the data are already free of missing values and categorical variables are converted to real values. In addition, each attribute is scaled to $[-1,1]$ or rather $[0,1]$.

\subsection{Toy examples \label{sec:toy}}

In the following toy examples we generate our data as follows.
The available set of instances $\X\subset\D^d$ consists of $M$ randomly chosen and uniformly distributed feature vectors $\x_j\in[0,\sfrac 12]^d$ or rather $\x_j\in[-\sfrac12,\sfrac12)^d$, $j=1,2,\dots,M$, depending on the basis function system used.
Using a known test function $f$ we obtain the corresponding labels $y_j\in\{-1,+1\}$ by computing $y_j=\sign f(\x_j)$.
Our goal is not to recover the function $f$, but rather to construct a classifying function that fits the given training data points $(\x_j,y_j)$, $j=1,2,\dots,M$.
This means that we determine basis coefficients $\bfh$ on a presumed grouped index set $\I$ with presumed active set $U$, as explained before.

In these numerical examples we construct the test function in such a way that it perfectly fits to our feature map, in order to ensure that a perfectly classifying functions exists and should be found by our algorithm.
In the first two examples, we even know the required bandwidth with which we should be able to find such a classifying function.
This is a fairly simple test scenario, but it is certainly the first to take in order to verify the functionality of our method.

\subsubsection{One-dimensional example \label{sec:one}}

The one-dimensional example is theoretically very simple, but it offers a good possibility of visualization.
We generate our training data set by using a zero-mean function $f$ defined via
\begin{equation*}
	f^{\cos}(x):=\sum_{k=1}^{5} c_k\varphi^{\cos}_{k}(x),\quad c_k=\frac{k+1}{4}
\end{equation*}
or rather
\begin{equation*}
	f^{\chui}(x):=\sum_{j=0}^{2}\sum_{k\in\mathcal{K}_{j}}c_{j,k}\varphi^{\chui, \text{per}}_{j,k}(x),\quad c_{j,k}=\frac{k+1}{4}.
\end{equation*}
As we can see, we construct the test function in such a way that we have $|\I_{(6)}(\{1\})|=5$ non-zero basis coefficients in case of the cosine basis, and $\sum_{j\in\mathcal{J}_{(2)}(\{1\})}\abs{\mathcal{K}_{j}}=7$ non-zero wavelet coefficients, otherwise.
For the entire experiment, we set $M=50$ and use $100$ randomly chosen instances and their corresponding labels for testing.
We determine the coefficients by solving the optimization problems~\eqref{a} or~\eqref{b}, depending on the applied regularization.

As the presumed active set we use $U=\{\emptyset,\{1\}\}$, which is not surprising in a one-dimensional setting, and exactly incorporate those basis functions in the feature map, which have non-zero coefficients in the above defined test functions.
In the following we denote by $\bfh^{\cos}$ and $\bfh^{\chui}$ the fitted basis coefficients of the cosine and the wavelet basis functions, respectively.

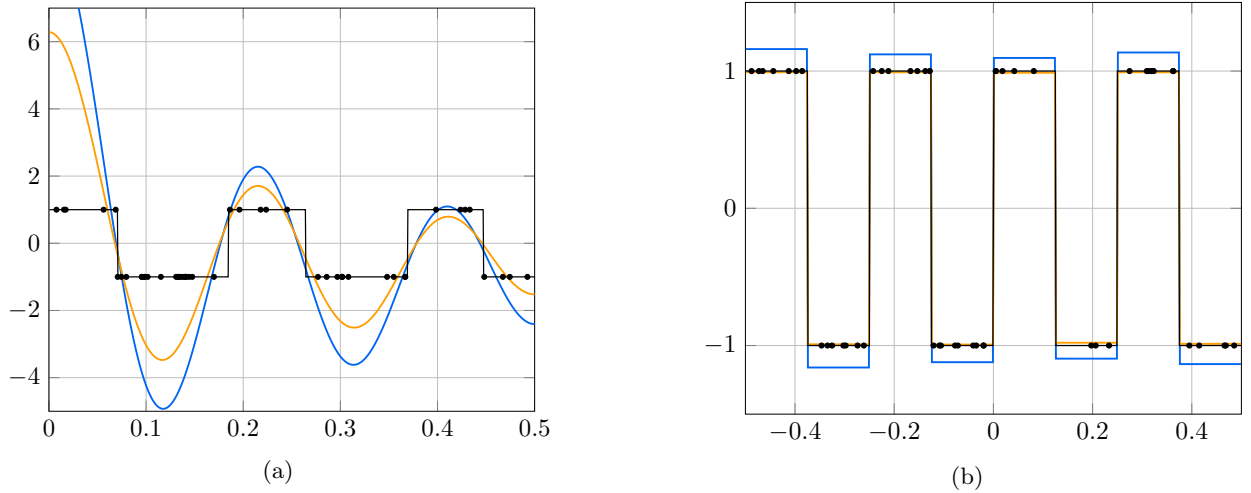
\begin{figure}[ht!]
	\centering
	\begin{subfigure}[c]{0.45\linewidth}
		\begin{adjustbox}{width=\linewidth}
			\begin{tikzpicture}
				\begin{axis}[
					grid=major,
					xmin=0.0,
					xmax=0.5,
					ymin=-5.0,
					ymax=7.0
					]
					\addplot[color = blue100, mark=false, line width=0.75pt] table [x=x_cos, y=l1_cos, col sep=comma] {cos.csv};
					\addplot[color = orange100, mark=false, line width=0.75pt] table [x=x_cos, y=l2_cos, col sep=comma] {cos.csv};
					
					\addplot[mark=false, line width=0.5pt]table [x=x_cos, y=ys_cos, col sep=comma] {cos.csv};
					\addplot [black, only marks, mark size=1.0pt] table [x=x_cos, y=y_cos, col sep=comma] {data.csv};
				\end{axis}
			\end{tikzpicture}
		\end{adjustbox}
		\caption{}
	\end{subfigure}
	\hfill
	\begin{subfigure}[c]{0.44\linewidth}
		\begin{adjustbox}{width=\linewidth}
			\begin{tikzpicture}
				\begin{axis}[
					grid=major,
					xmin=-0.5,
					xmax=0.5,
					ymin=-1.5,
					ymax= 1.5
					]
					\addplot[color = blue100, mark=false, line width=0.75pt] table [x=x_chui, y=l1_chui, col sep=comma] {chui.csv};
					\addplot[color = orange100, mark=false, line width=0.75pt] table [x=x_chui, y=l2_chui, col sep=comma] {chui.csv};
					
					\addplot[mark=false, line width=0.5pt]table [x=x_chui, y=ys_chui, col sep=comma] {chui.csv};
					\addplot [black, only marks, mark size=1.0pt] table [x=x_chui, y=y_chui, col sep=comma] {data.csv};
				\end{axis}
			\end{tikzpicture}
		\end{adjustbox}
		\caption{}
	\end{subfigure}
	\caption{
		The sign of the univariate test functions and the training data points $(x_j,y_j)$, $j=1,2,\dots,M$ are visualized in black. The obtained classifying functions are depicted in orange ($\ell_2$-norm regularization) and in blue ($\ell_1$-norm regularization). The regularization parameter is set to $\lambda=0.01$.
		The classifying function $S(\X,\I_{\b N})f(\x)$ with $\bfh=\bfh^{\cos}$ is visualized in (a) and $S(\X,\mathcal J_{\b N})f(\x)$ with $\bfh=\bfh^{\chui}$ in (b).}
	\label{fig:1}
\end{figure}

As regularization parameter we use, without further considerations, $\lambda= 0.01$. Note that we consider different regularization parameters in the multi-variate examples later on.
In Figure~\ref{fig:1} we plot the obtained classifying functions for the $\ell_2$-norm regularization approach (orange) and when applying $\ell_1$-norm regularization (blue). In addition, we plot the sign of the test functions in black and the randomly generated training data by the little black dots.
In order to obtain a more representative result, we performed $100$ runs in total, resulting in an average test accuracy of $94.93\%$ for the approach with cosine basis with $\ell_1$-norm regularization and $94.87\%$ if we apply $\ell_2$-norm regularization.
In case of using wavelets we achieve an average accuracy of $100\%$ for both regularization approaches.

Note that the zeros of the classifying wavelet function have fixed positions, making it fairly easy to achieve this good result. In addition, our model fits the test function perfectly.
Although the perfect ansatz is also used in case of the cosine system, we face some problems if training data points are missing close to the zero points of the test function. Then, test data points close to the jumps may be misclassified.
So, even in such a constructed test scenario, it is not easy to achieve an accuracy of 100\%.

\subsubsection{Six-dimensional example \label{sec:six}}

In this example we consider test functions $f:\R^6\to\R$.
Similar to the previous section, we construct these functions as linear combinations of cosine or wavelet basis functions, respectively.
As the true active set we use $U^\star:=(\{4\}, \{6\}, \{2,3\})$ and define
\begin{equation*}
	f^{\cos}(\x)=\sum_{\bfk\in\I_{\left((6), (4), (2,4)\right)}(U^\star)} 1\cdot\varphi^{\cos}_{\k}(\x)
\end{equation*}
and
\begin{equation*}
	f^{\chui}(\x)=\sum_{\j\in\mathcal{J}_{ \left((3), (3), (3,3)\right)}(U^\star)}\sum_{\bfk\in\mathcal{K}_{\j}}1\cdot\varphi^{\chui,\text{per}}_{\j,\k}(\x)
\end{equation*}
depending on the considered system of basis functions. With the applied band widths we have $19$ non-zero basis coefficients in case of the cosine system and $71$ non-zero wavelet coefficients.
We generate our training data points $(\x_j, y_j)$, $j=1,2,\dots,M$ using the test functions $f^{\cos}$ or rather $f^{\chui}$.

As the presumed active set we use $U_{d_s}$ with $d_s=2$, i.e., we incorporate all subsets of cardinality $\leq2$ into our feature map.
The feature map or rather the corresponding data matrices are built by using the following band widths $\b N= (\b N^{\u})_{\u\in U}$ with $\b N^\u\in\N^{|\u|}$,
\begin{equation}\label{eq:N_Phi_cos}
	\b \Phi^{\cos}:=\b \Phi^{\cos}(\X, \I_{\b N}(U_2)),\quad \b N^{\u}=\begin{cases}
		(6)&:\quad\abs{\u}=1,\\
		(4,4)&:\quad\abs{\u}=2
	\end{cases}
\end{equation}
and
\begin{equation}\label{eq:N_Phi_chui}
	\b \Phi^{\chui}:= \b \Phi^{\chui}(\X, \mathcal{J}_{\b N}(U_2)),\quad \b N^{\u}=\begin{cases}
		(3)&:\quad\abs{\u}=1,\\
		(3,3)&:\quad\abs{\u}=2.
	\end{cases}
\end{equation}
So, we incorporate all multi-indices that are non-zero in the constructed test functions, but we also make use of a lot of basis functions that do not have any influence.
In case of the cosine system we have $\bfh^{\cos}\in\R^{166}$, whereas only $19$ are present in $f^\text{cos}$.
For the wavelet system we have $\bfh^{\chui}\in\R^{876}$ and only 71 active coefficients in $f^\chui$.

Note that we need much more data points as in the previous example, since we are now in a six-dimensional setting.
We use $M=1000$ in case of the cosine basis and $M=5000$ for the wavelet basis approach.
The same number of randomly generated points are used as the test data set.
Please note that we cannot directly compare both test cases, since the test functions are completely different and have different numbers of non-zero basis coefficients.
Since the number of non-zero coefficients is roughly 5 times larger in the wavelet setting, we choose a 5 times larger training data set for that setting. 

The choice of the regularization parameter is not negligible and, thus, we solve the considered optimization problems with different regularization parameters $\lambda$.
In Figure~\ref{fig:2} we plot the achieved accuracy on the test data set with respect to the applied $\lambda$.
Again, all accuracies have to be understood as average values achieved over 100 runs.
Using the $\ell_1$-norm regularization approach as an example, we see that there is an optimal value for $\lambda$. For larger or smaller values the classification rate becomes worse.

\begin{figure}[ht!]%
	\begin{center}
		\begin{subfigure}[t]{0.45\linewidth}
			\begin{adjustbox}{width=\linewidth}
				\begin{tikzpicture}
					\begin{axis}[  ymin=90, ymax=97, xmode=log, xmin=0.0009765625, xmax=0.25, xtick={0.0009765625, 0.001953125, 0.00390625, 0.0078125, 0.015625, 0.03125, 0.0625, 0.125, 0.25, 0.5},enlargelimits=0.05, grid = major, xtick={0.0009765625, 0.00390625,0.015625,  0.0625,0.25}, xticklabels={$2^{-10}$, $2^{-8}$, $2^{-6}$, $2^{-4}$, $2^{-2}$}, xlabel = $\lambda$, ylabel = CA]
						\addplot+[blue100,  mark=x, thick]  plot coordinates{
							(0.0009765625, 94.52699999999999)
							(0.001953125, 95.679)
							(0.00390625, 96.229)
							(0.0078125, 96.294)
							(0.015625, 96.49)
							(0.03125, 96.28200000000001)
							(0.0625,  95.274)
							(0.125, 93.17400000000002)
							(0.25, 90.64)
						};
						\addplot[orange100, mark=x, thick]  plot coordinates{
							(0.0009765625, 93.271)
							(0.001953125, 93.505)
							(0.00390625, 93.233)
							(0.0078125, 92.95200000000001)
							(0.015625, 92.975)
							(0.03125, 92.97399999999999)
							(0.0625, 92.69900000000001)
							(0.125, 92.44900000000001)
							(0.25, 92.30799999999999)
						};
					\end{axis}
				\end{tikzpicture}
			\end{adjustbox}
			\subcaption[(A)]{}
		\end{subfigure}
		\hspace{5pt}%
		\begin{subfigure}[t]{0.45\linewidth}
			\begin{adjustbox}{width=\linewidth}
				\begin{tikzpicture}
					\begin{axis}[  ymin=90, ymax=97, xmode=log, xmin=0.0009765625, xmax=0.25, xtick={0.0009765625, 0.001953125, 0.00390625, 0.0078125, 0.015625, 0.03125, 0.0625, 0.125, 0.25, 0.5},enlargelimits=0.05, grid = major,xtick={0.0009765625, 0.00390625,0.015625,  0.0625,0.25}, xticklabels={$2^{-10}$, $2^{-8}$, $2^{-6}$, $2^{-4}$, $2^{-2}$},  xlabel = $\lambda$, ylabel = CA]
						\addplot+[blue100,  mark=x, thick]  plot coordinates{
							(0.0009765625, 95.6576)
							(0.001953125,96.0096)
							(0.00390625, 96.28380000000001)
							(0.0078125, 96.45379999999999)
							(0.015625,96.55300000000001)
							(0.03125, 95.76799999999999)
							(0.0625,  93.43799999999999)
							(0.125, 86.45)
							(0.25, 34.0394)
						};
						\addplot[orange100, mark=x, thick]  plot coordinates{
							(0.0009765625,  92.68520000000001)
							(0.001953125, 92.79)
							(0.00390625, 92.9942)
							(0.0078125, 93.0942)
							(0.015625, 93.0352)
							(0.03125,  92.78240000000001)
							(0.0625,  92.47140000000002)
							(0.125, 91.824)
							(0.25, 91.06)
						};
					\end{axis}
				\end{tikzpicture}
			\end{adjustbox}
			\caption{}
		\end{subfigure}
		\caption{Average of mean CA over $100$ runs using $\ell_1$-norm regularization (blue) and $\ell_2$-norm regularization (orange)  for the six-dimensional test functions. In (a) we visualize the results achieved by the classifying functions $S(\X,\I_{\b N}(U_2))f^{\cos}(\x)$ with $\bfh=\bfh^{\cos}$, where the number of generated training and test data points have been set to $1000$. In (b), we see the results achieved by the classifying functions $S(\X,\mathcal J_{\b N}(U_2))f(\x)$ with $\bfh=\bfh^\text{\chui}$ and $5000$ generated training and test data points.} 
		\label{fig:2}
	\end{center}
\end{figure}

Although we used a lot of basis functions for fitting the model that have not been present in the original test function, we already achieve a good classification rate.

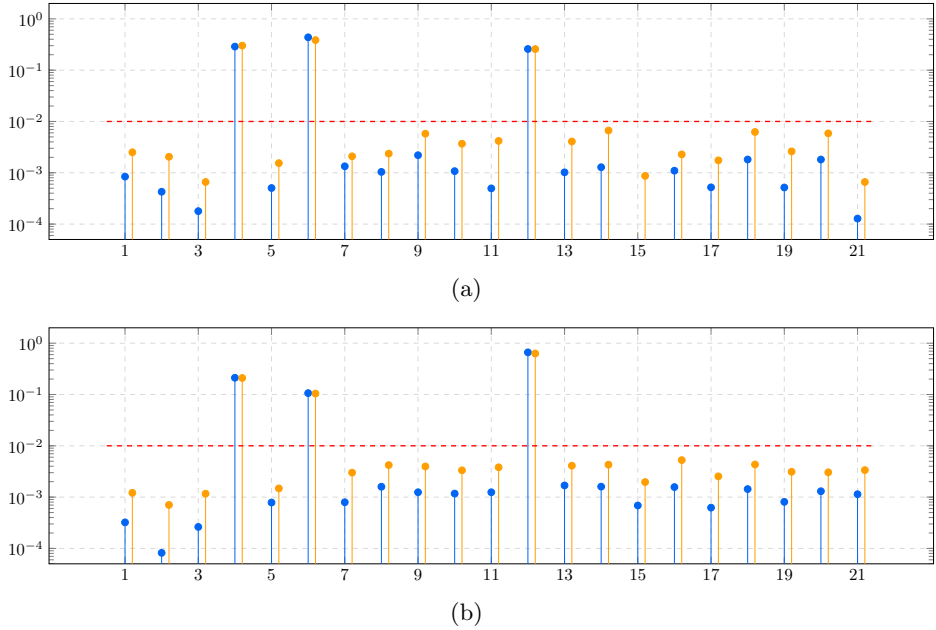
\begin{figure}[ht!]%
	\centering
	\begin{subfigure}[t]{0.75\linewidth}
		\begin{adjustbox}{width=\linewidth}
			\begin{tikzpicture}
				\begin{axis}[ width=200mm, height=6.5cm, ymode=log, log origin=infty,
					ymin=1e-4, ymax=1e0, xmin=0.5, xmax=21.5, enlargelimits=0.075, grid = major,
					grid style={dashed, gray!30}, xtick={1,3,5,7,9,11,13,15,17,19,21}, xticklabels={1,3,5,7,9,11,13,15,17,19,21}]
					\addplot+[ycomb, blue100, mark options={blue100},  mark=*] plot coordinates { (1.0,0.0008413693861324692)
						(2.0,0.00042681056274026396)
						(3.0,0.00017870997100545206)
						(4.0,0.28830068275850224)
						(5.0,0.0005043251306005572)
						(6.0,0.4369700272481633)
						(7.0,0.0013376207059387218)
						(8.0,0.0010402150926086713)
						(9.0,0.0022021255538716296)
						(10.0,0.0010758833417094796)
						(11.0,0.0004958730144732962)
						(12.0,0.25842955090804237)
						(13.0,0.0010202762079872276)
						(14.0,0.0012834498520912074)
						(15.0,3.06423685402885e-6)
						(16.0,0.0010980333953458318)
						(17.0,0.000520157581563887)
						(18.0,0.0018155854253929405)
						(19.0,0.0005169653743982506)
						(20.0,0.0018112740474024625)
						(21.0,0.00012800020517549448)};
					\addplot+[ycomb, orange100, mark options={orange100}, mark=*] plot coordinates {(1.2,0.0025072511114673656)
						(2.2,0.0020502193973364205)
						(3.2,0.0006613233134997316)
						(4.2,0.3007970671541295)
						(5.2,0.0015418493203854199)
						(6.2,0.3856275226660418)
						(7.2,0.002092666827159254)
						(8.2,0.0023634622106316785)
						(9.2,0.005781072477743792)
						(10.2,0.0036792667912170086)
						(11.2,0.004176996163012594)
						(12.2,0.257690241275886)
						(13.2,0.004073003719860557)
						(14.2,0.006669726058492489)
						(15.2,0.0008696901485739435)
						(16.2,0.0022837512918201545)
						(17.2,0.001748574234516392)
						(18.2,0.00624136884354733)
						(19.2,0.002612886302141495)
						(20.2,0.005870542201257109)
						(21.2,0.0006615184912801116)};
					\addplot[red,sharp plot,update limits=false, dashed, thick] 
					coordinates {(0.5,1e-2) (21.5,1e-2)};
				\end{axis}
			\end{tikzpicture}
		\end{adjustbox}
		\caption{}
		\hspace{5pt}%
	\end{subfigure}
	\begin{subfigure}[t]{0.75\linewidth}
		\begin{adjustbox}{width=\linewidth}
			\begin{tikzpicture}
				\begin{axis}[ width=200mm, height=6.5cm, ymode=log,log origin=infty,
					ymin=1e-4, ymax=1e0, xmin=0.5, xmax=21.5, enlargelimits=0.075, grid = major,
					grid style={dashed, gray!30}, xtick={1,3,5,7,9,11,13,15, 17,19,21}, xticklabels={1,3,5,7,9,11,13,15, 17,19,21}]
					\addplot+[ycomb, blue100, mark options={blue100}, mark=*] plot coordinates {(1.0,0.00032198380024324523)
						(2.0,8.203234745359922e-5)
						(3.0,0.0002625820914614492)
						(4.0,0.2117933569361955)
						(5.0,0.0007843056881497932)
						(6.0,0.10633122588181323)
						(7.0,0.0007943024808267674)
						(8.0,0.0015939406793699959)
						(9.0,0.0012435821781668368)
						(10.0,0.00117179314806885)
						(11.0,0.0012429665622121114)
						(12.0,0.6635255168899135)
						(13.0,0.0016885513152971239)
						(14.0,0.0016006161027453637)
						(15.0,0.0006872610420661649)
						(16.0,0.0015692420403358403)
						(17.0,0.0006235411875073014)
						(18.0,0.0014344236923197866)
						(19.0,0.0008078705156425895)
						(20.0,0.0013025946524030014)
						(21.0,0.001138310767807822)};
					\addplot+[ycomb, orange100, mark options={orange100}, mark=*] plot coordinates {(1.2,0.001213197761370055)
						(2.2,0.0007039803691017711)
						(3.2,0.0011661068663833456)
						(4.2,0.20989002984359303)
						(5.2,0.0014804845404446491)
						(6.2,0.10430176292316932)
						(7.2,0.003000600516167711)
						(8.2,0.004216213121007179)
						(9.2,0.00395829107740235)
						(10.2,0.0033318981309668452)
						(11.2,0.003812177401901598)
						(12.2,0.6309318545684394)
						(13.2,0.004095654924708931)
						(14.2,0.004290465786537245)
						(15.2,0.00196531123474549)
						(16.2,0.005261548824643134)
						(17.2,0.0025373543256824917)
						(18.2,0.0043251793931651725)
						(19.2,0.003112316113234913)
						(20.2,0.003045345479947005)
						(21.2,0.0033602267973887007)};
					\addplot[red,sharp plot,update limits=false, dashed, thick] coordinates {(0.5,1e-2) (21.5,1e-2)};
				\end{axis}
			\end{tikzpicture}
		\end{adjustbox}
		\caption{}
	\end{subfigure}
	\caption{The global sensitivity indices $\varrho(\u, S(\X,\I_{\b N}(U_2))f^\text{cos})$  with $\bfh=\bfh^{\cos}$ are visualized in (a) and $\varrho(\u, S(\X,\mathcal{J}_{\b N}(U_2)f^{\chui})$ with $\bfh=\bfh^{\chui}$ in (b), for the six-dimensional test function.
		The results using $\ell_1$-norm regularization are shown in blue and for $\ell_2$-norm regularization in orange for $\u\in U_2$, where $|U_2|=21$.}
	\label{fig:3}
\end{figure}

Now, of course, the question remains as to whether we can reconstruct the true active set with the help of the global sensitivity indices of the obtained classifying functions.
To this end, we compute the global sensitivity indices of the classifying functions, which are plotted in Figure~\ref{fig:3}.
It can be seen that the global sensitivity indices are significantly larger than zero only for subsets $\u\in U^\star$, which is a very nice results.
Almost all the other subsets have global sensitivity indices smaller or equal to $10^{-2}$ in case of $\ell_2$-norm regularization and even $10^{-6}$ when $\ell_1$-norm regularization is applied.
In summary, with both optimization approaches we were able to detect the original active set $U^\star$.
The result is even clearer in the case of $\ell_1$-norm regularization.

In addition, we applied the threshold parameter $\varepsilon=0.01$ in order to compute the corresponding active set $U^{0.01}$, see~\eqref{eq:ueps1} or~\eqref{eq:ueps2}, and fitted the classifying functions again, only using this active set, applying the same bandwidths as defined above.
By that we are able to improve the classification performance further, see Figure~\ref{fig:6dimproved}.

\begin{figure}[ht!]%
	\begin{center}
		\begin{subfigure}[c]{0.45\linewidth}
			\begin{adjustbox}{width=\linewidth}
				\begin{tikzpicture}
					\begin{axis}[  ymin=91, ymax=100, xmode=log, xmin=0.0009765625, xmax=0.3, xtick={0.0009765625, 0.001953125, 0.00390625, 0.0078125, 0.015625, 0.03125, 0.0625, 0.125, 0.25, 0.5},enlargelimits=0.05, grid = major, xtick={0.000244140625, 0.0009765625, 0.00390625,0.015625,  0.0625,0.25}, xticklabels={$2^{-12}$, $2^{-10}$, $2^{-8}$, $2^{-6}$, $2^{-4}$, $2^{-2}$}, xlabel = $\lambda$, ylabel = CA]
						\addplot+[blue100,  mark=x, thick]  plot coordinates{
							(0.0009765625,99.009)
							(0.001953125,98.934)
							(0.00390625,98.71900000000001)
							(0.0078125,98.42699999999999)
							(0.015625,98.04899999999999)
							(0.03125,97.506)
							(0.0625,96.227)
							(0.125,94.325)
							(0.25,90.77699999999999)
						};
						\addplot+[orange100, mark=x, thick]  plot coordinates{
							(0.0009765625,98.891)
							(0.001953125,98.65899999999999)
							(0.00390625,98.355)
							(0.0078125,98.02199999999999)
							(0.015625,97.665)
							(0.03125,97.29899999999999)
							(0.0625,96.71800000000002)
							(0.125,95.929)
							(0.25,95.41799999999999)
						};
					\end{axis}
				\end{tikzpicture}
			\end{adjustbox}
			\caption{}
		\end{subfigure}
		\hspace{5pt}%
		\begin{subfigure}[c]{0.45\linewidth}
			\begin{adjustbox}{width=\linewidth}
				\begin{tikzpicture}
					\begin{axis}[  ymin=95, ymax=100, xmode=log, xmin=0.0009765625, xmax=0.25,enlargelimits=0.05, grid = major,xtick={ 0.0009765625, 0.00390625,0.015625,  0.0625,0.25}, xticklabels={$2^{-10}$, $2^{-8}$, $2^{-6}$, $2^{-4}$, $2^{-2}$}, xlabel = $\lambda$, ylabel = CA]
						\addplot+[blue100,  mark=x, thick]  plot coordinates{
							(0.0009765625,100.0)
							(0.001953125,99.99939999999998)
							(0.00390625,99.96560000000001)
							(0.0078125,99.77579999999998)
							(0.015625,99.392)
							(0.03125,98.23)
							(0.0625,96.27420000000001)
							(0.125,93.01100000000001)
							(0.25,84.69460000000001)
						};
						\addplot[orange100, mark=x, thick]  plot coordinates{
							(0.0009765625,100.0)
							(0.001953125,99.9912)
							(0.00390625,99.96800000000002)
							(0.0078125,99.9268)
							(0.015625,99.79019999999998)
							(0.03125,99.55319999999998)
							(0.0625,99.0608)
							(0.125,98.3382)
							(0.25,97.26140000000004)
						};
					\end{axis}
				\end{tikzpicture}
			\end{adjustbox}
			\caption{}
		\end{subfigure}
		\caption{Average of mean CA over $100$ runs using $\ell_1$-norm regularization (blue) and $\ell_2$-norm regularization (orange)  for the six-dimensional test functions. In (a) we visualize the results achieved by the classifying functions $S(\X,\I_{\b N}(U^{0.01}))f^{\cos}(\x)$ with $\bfh=\bfh^{\cos}$, where the number of generated training and test data points have been set to $1000$. In (b), we see the results obtained with the classifying functions $S(\X,\mathcal J_{\b N}(U^{0.01}))f^{\chui}(\x)$ with $\bfh=\bfh^\text{\chui}$ and $5000$ generated training  and test data points.}
		\label{fig:6dimproved}
	\end{center}
\end{figure}

\newpage
\subsubsection{Ten-dimensional example}\label{sec:ten}

For our third numerical example we introduce one of the so-called Friedman functions~\cite{Friedman}, which were used as benchmark example in~\cite{MeLeHo03} and have since become an often used example in approximation of functions with scattered data, see~\cite{BeGaMo09, BiDaLa11}. Moreover, in \cite{PoSc19a} the authors achieved very good results regarding the reconstruction of Friedman functions. We focus on the Friedman $1$ function
\begin{equation}\label{eq:friedman1}
	F^1: [0,1]^{10}\to \R, F^1(\x)= 10\sin(\pi x_1 x_2)+20\left(x_3-\tfrac12\right)^2+10x_4+5x_5,
\end{equation}
where only five of the ten variables have any influence on the function. Further, the mean value of the function is given by
\begin{equation}\label{eq:MF1}
	M(F^1)=\int_{[0,1]^d} F^1(\x)\mathrm d\x
	=10\int_0^1 \frac{1-\cos(\pi x_1)}{\pi x_1}\mathrm dx_1 +\frac{55}{6} \approx 14.4133,
\end{equation}
where the included integral can be expressed in terms of the well-known incomplete Gamma function and has to be computed numerically.
For our synthetic generated data set $\{(\x_j, y_j): j=1,2,\dots,M\}$ we generate $M$ random uniformly distributed feature vectors $\x_j\in[0,\sfrac 12]^{10}$ or rather $\x_j\in[-\sfrac12,\sfrac12)^{10}$ and obtain the corresponding labels, similar to the previous examples, by $y_j=\sign (f^\text{cos}(\x_j))$ or rather $y_j=\sign (f^\text{chui}(\x_j))$, where we define the zero-mean functions
\begin{equation*}
	f^\text{cos}: \left[0,\tfrac12\right]^{10}\to\R, \quad f^\text{cos}(\x)=F^1(2\x)-M(F^1)
\end{equation*}
and
\begin{equation*}
	f^{\chui}: \left[-\tfrac12,\tfrac12\right)^{10}\to\R,\quad  f^\text{chui}(\x)=F^1(\x+\tfrac 12)-M(F^1),
\end{equation*}
for the two considered bases.

Note that the ANOVA terms of the Friedman $1$ function can be stated explicitly, see Appendix~\ref{sec:gsi_friedman}, for an active set $U^\star=\{\emptyset,\{1\},\{2\},\{3\},\{4\},\{5\},\{1,2\}\}$, where only one two-dimensional term is relevant. Furthermore, we numerically verified that all present ANOVA terms significantly contribute to $\sign(F^{1}(\x)-M(F^1))$, as presented in Appendix~\ref{sec:importance_friedman}.
Our goal will be to determine the true active set by analyzing the global sensitivity indices of the classifying function, similarly to the previous examples.

\begin{figure}[ht!]%
	\begin{center}
		\begin{subfigure}[t]{0.45\linewidth}
			\begin{adjustbox}{width=\linewidth}
				\begin{tikzpicture}
					\begin{axis}[  ymin=75, ymax=95, xmode=log, xmin=0.000244140625, xmax=0.25,enlargelimits=0.05, grid = major,xtick={0.000244140625,0.0009765625, 0.00390625,0.015625,  0.0625,0.25}, xticklabels={$2^{-12}$,$2^{-10}$, $2^{-8}$, $2^{-6}$, $2^{-4}$, $2^{-2}$}, xlabel = $\lambda$, ylabel = CA, legend pos= south west]
						\addplot[blue35, mark=x, thick]  plot coordinates{
							(0.000244140625,91.15)
							(0.00048828125,91.28)
							(0.0009765625,91.47999999999999)
							(0.001953125,91.67999999999999)
							(0.00390625,91.83000000000001)
							(0.0078125,92.01000000000002)
							(0.015625,92.22)
							(0.03125,91.92999999999999)
							(0.0625,91.61)
							(0.125,90.34)
							(0.25,86.68000000000002)
						};
						\addlegendentry{$N_1=6, N_2=2$}
						\addplot[blue100, mark=x, thick]  plot coordinates{
							(0.000244140625,90.08999999999999)
							(0.00048828125,91.33000000000001)
							(0.0009765625,92.09)
							(0.001953125,92.55)
							(0.00390625,92.78)
							(0.0078125,93.52000000000001)
							(0.015625,94.46)
							(0.03125,95.13999999999999)
							(0.0625,95.35000000000001)
							(0.125,94.05)
							(0.25,88.99999999999999)
						};
						\addlegendentry{$N_1=6, N_2=4$}
						\addplot[blue65,  mark=x, thick]  plot coordinates{
							(0.000244140625,86.09)
							(0.00048828125,88.67999999999999)
							(0.0009765625,90.68)
							(0.001953125,91.83999999999999)
							(0.00390625,92.22999999999999)
							(0.0078125,92.53999999999999)
							(0.015625,93.10999999999999)
							(0.03125,93.97999999999999)
							(0.0625,94.86999999999999)
							(0.125,94.14999999999999)
							(0.25,88.99999999999999)
						};
						\addlegendentry{$N_1=6, N_2=6$}
					\end{axis}
				\end{tikzpicture}
			\end{adjustbox}
			\caption{}
		\end{subfigure}
		\hspace{5pt}%
		\begin{subfigure}[t]{0.45\linewidth}
			\begin{adjustbox}{width=\linewidth}
				\begin{tikzpicture}
					\begin{axis}[  ymin=75, ymax=95, xmode=log, xmin=0.000244140625, xmax=0.25,enlargelimits=0.05, grid = major,xtick={0.000244140625, 0.0009765625, 0.00390625,0.015625,  0.0625,0.25}, xticklabels={$2^{-12}$, $2^{-10}$, $2^{-8}$, $2^{-6}$, $2^{-4}$, $2^{-2}$}, xlabel = $\lambda$, ylabel = CA, legend pos=south west]
						\addplot[orange100, mark=x, thick]  plot coordinates{
							(0.000244140625,91.16000000000001)
							(0.00048828125,91.25)
							(0.0009765625,91.28)
							(0.001953125,91.24)
							(0.00390625,91.32000000000001)
							(0.0078125,91.32000000000001)
							(0.015625,91.13000000000001)
							(0.03125,90.78)
							(0.0625,90.45)
							(0.125,90.07)
							(0.25,89.6)
						};
						\addlegendentry{$N_1=6, N_2=2$}
						\addplot[orange65, mark=x, thick]  plot coordinates{
							(0.000244140625,89.54)
							(0.00048828125,90.11)
							(0.0009765625,90.02)
							(0.001953125,88.53999999999999)
							(0.00390625,86.83000000000001)
							(0.0078125,85.05)
							(0.015625,84.15)
							(0.03125,84.55)
							(0.0625,85.11)
							(0.125,85.43)
							(0.25,85.35999999999999)			
						};
						\addlegendentry{$N_1=6, N_2=4$}
						\addplot[orange35,  mark=x, thick]  plot coordinates{
							(0.000244140625,85.69000000000001)
							(0.00048828125,87.67)
							(0.0009765625,88.91)
							(0.001953125,88.09)
							(0.00390625,85.30999999999999)
							(0.0078125,79.97)
							(0.015625,77.13)
							(0.03125,75.31)
							(0.0625,74.71000000000001)
							(0.125,74.81)
							(0.25,75.22)
						};
						\addlegendentry{$N_1=6, N_2=6$}
					\end{axis}
				\end{tikzpicture}
			\end{adjustbox}
			\caption{}
		\end{subfigure}
		\caption{Visualization of mean CA over 10 runs achieved by the ten-dimensional classifying functions $S(\X,\I_{\b N}(U_2))f^\text{cos}(\x)$ with $\bfh=\bfh^{\cos}$, using different regularization parameters $\lambda$ and bandwidths $\b N$, cf. \eqref{eq:N_Phi_cos_f}.
			The results achieved with $\ell_1$-norm regularization are depicted in (a) and the results achieved with $\ell_2$-norm regularization are shown in (b).
			The number of generated training and test data points have been set to $1000$.
			\label{fig:7}}
	\end{center}
\end{figure}
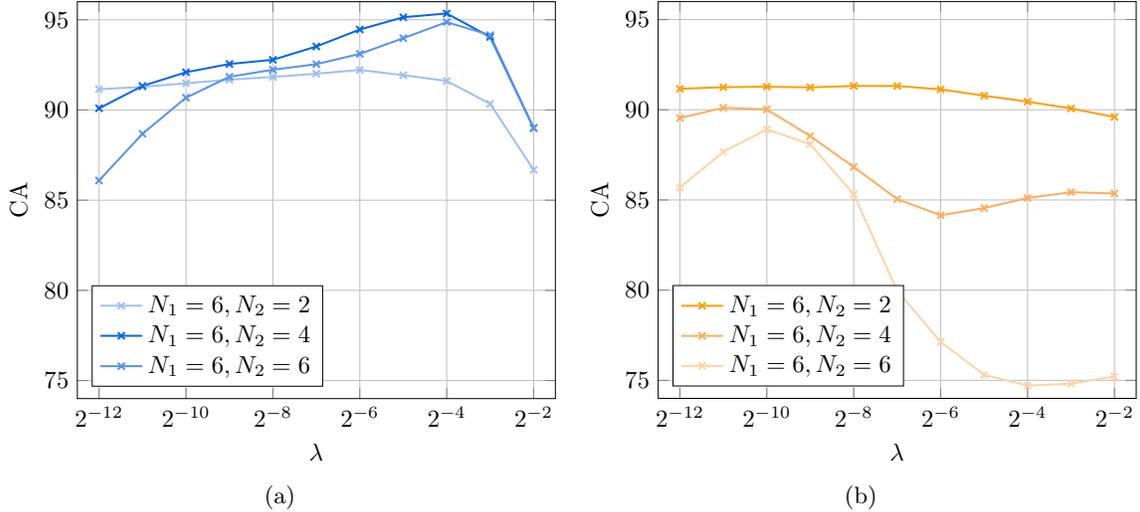

We begin by setting the superposition threshold to $d_s=2$. Since the bandwidth $\b N= (\b N^{\u})_{\u\in U}$ with $\b N^\u\in\N^{|\u|}$ is unknown in this setting, we consider different bandwidths in order to generate the feature map.
To this end, we use the following notation in order to define the corresponding feature map and feature matrices.
\begin{equation}\label{eq:N_Phi_cos_f}
	\b \Phi^{\cos}:=\b \Phi^{\cos}(\X, \I_{\b N}(U_2)),\quad \b N^{\u}=\begin{cases}
		(N_1)&:\quad\abs{\u}=1,\\
		(N_2, N_2)&:\quad\abs{\u}=2
	\end{cases}
\end{equation}
and
\begin{equation}\label{eq:N_Phi_chui_f}
	\b \Phi^{\chui}:= \b \Phi^{\chui}(\X, \mathcal{J}_{\b N}(U_2)),\quad \b N^{\u}=\begin{cases}
		(N_1)&:\quad\abs{\u}=1,\\
		(N_2, N_2)&:\quad\abs{\u}=2.
	\end{cases}
\end{equation}
Let us first analyze the results obtained by using the cosine basis, visualized in Figure~\ref{fig:7}. We generate $M=1000$ training data points $(\x_j, y_j)$, $j=1,2,\dots,M$ and solve the considered optimization problems with different regularization parameters $\lambda$.
Depending on the applied bandwidths $\b N$ from \eqref{eq:N_Phi_cos_f} with $N_1=6$ and $N_2\in\{2,4,6\}$, we have $\bfh^{\cos}\in\R^{96}$ if $N_2=2$, $\bfh^{\cos}\in\R^{456}$ if $N_2=4$ and $\bfh^{\cos}\in\R^{1176}$ for $N_2=6$.
Afterwards we compute the classification accuracy using the corresponding classification functions $S(\X,\I_{\b N}(U_2))f^\text{cos}(\x)$ and the generated test data set, containing $M$ test data points.
All accuracies have to be understood as average values achieved over $10$ runs.

As we can see, we get different CA performance depending on the regularization. We achieve the highest mean CA with $N_1=6$, $N_2=4$ and $\lambda = 2^{-4}$, in case of $\ell_1$-norm regularization. For the best results provided by $\ell_2$-norm regularization we need to choose $N_1=6$, $N_2=2$ and $\lambda = 2^{-7}$.

\begin{figure}[ht!]%
	\centering
	\begin{subfigure}[t]{0.75\linewidth}
		\begin{adjustbox}{width=\linewidth}
			\begin{tikzpicture}
				\begin{axis}[ width=200mm, height=6.5cm, ymode=log,log origin=infty,
					ymin=1e-7, ymax=0.5,  xmin=0.5, xmax=55.5, enlargelimits=0.075, grid = major,
					grid style={dashed, gray!30}, xtick={1,5,10,15,20,25,30,35, 45,50,55}, xticklabels={1,5,10,15,20,25,30,35, 45,50,55}]
					\addplot+[ycomb, blue100, mark options={blue100}, mark=*] plot coordinates {
						(1.0,0.20790754099012215)
						(2.0,0.21287222618060736)
						(3.0,0.07522980206771326)
						(4.0,0.3604003092043927)
						(5.0,0.08466431603147545)
						(6.0,0.0)
						(7.0,0.0)
						(8.0,0.0)
						(9.0,0.0)
						(10.0,0.0)
						(11.0,0.058925306317024397)
						(12.0,0.0)
						(13.0,4.992086646322236e-7)
						(14.0,0.0)
						(15.0,0.0)
						(16.0,0.0)
						(17.0,0.0)
						(18.0,0.0)
						(19.0,0.0)
						(20.0,0.0)
						(21.0,6.206274602642142e-18)
						(22.0,0.0)
						(23.0,0.0)
						(24.0,0.0)
						(25.0,0.0)
						(26.0,0.0)
						(27.0,0.0)
						(28.0,0.0)
						(29.0,0.0)
						(30.0,0.0)
						(31.0,0.0)
						(32.0,0.0)
						(33.0,0.0)
						(34.0,0.0)
						(35.0,0.0)
						(36.0,0.0)
						(37.0,0.0)
						(38.0,0.0)
						(39.0,0.0)
						(40.0,0.0)
						(41.0,0.0)
						(42.0,0.0)
						(43.0,0.0)
						(44.0,0.0)
						(45.0,0.0)
						(46.0,0.0)
						(47.0,0.0)
						(48.0,0.0)
						(49.0,0.0)
						(50.0,0.0)
						(51.0,0.0)
						(52.0,0.0)
						(53.0,0.0)
						(54.0,0.0)
						(55.0,0.0)
					};
					\addplot[red,sharp plot,update limits=false, dashed, very thick] coordinates {(0.5,1e-2) (55.5,1e-2)};
				\end{axis}
			\end{tikzpicture}
		\end{adjustbox}
		\caption{}
		\hspace{5pt}%
	\end{subfigure}
	\begin{subfigure}[t]{0.75\linewidth}
		\begin{adjustbox}{width=\linewidth}
			\begin{tikzpicture}
				\begin{axis}[ width=200mm, height=6.5cm, ymode=log, log origin=infty,
					ymin=1e-7, ymax=1e0, xmin=0.5, xmax=55.5, enlargelimits=0.075, grid = major,
					grid style={dashed, gray!30}, xtick={1,5,10,15,20,25,30,35, 40,45,50,55}, xticklabels={1,5,10,15,20,25,30,35, 40,45,50,55}]
					\addplot+[ycomb, orange, mark options={orange}, mark=*] plot coordinates {
						(1.0,0.25018040472118885)
						(2.0,0.2581053930161319)
						(3.0,0.07372212782929553)
						(4.0,0.2906782729996294)
						(5.0,0.07995647816285126)
						(6.0,0.00018939459259995574)
						(7.0,0.0002507160051477396)
						(8.0,0.0005030740593014548)
						(9.0,0.0001771084811451153)
						(10.0,0.00020669783581272192)
						(11.0,0.021097615822544078)
						(12.0,1.504325262764953e-5)
						(13.0,0.008119745951115195)
						(14.0,0.0017176381580648083)
						(15.0,2.5076434478640808e-5)
						(16.0,9.082539102739204e-6)
						(17.0,8.237482858979012e-5)
						(18.0,0.0002316918764653741)
						(19.0,5.786176954462407e-5)
						(20.0,0.0002546290814710234)
						(21.0,0.008397000457225391)
						(22.0,0.00330069411245901)
						(23.0,0.0004656607892477459)
						(24.0,2.8846359238436075e-5)
						(25.0,5.325509296222334e-6)
						(26.0,6.79879365822452e-5)
						(27.0,0.00035751167376053437)
						(28.0,9.641803141152022e-5)
						(29.0,6.005148055158917e-5)
						(30.0,1.0658434732991519e-5)
						(31.0,1.4326772376796455e-7)
						(32.0,3.269363224053944e-7)
						(33.0,3.131424440691287e-5)
						(34.0,5.577673137018085e-6)
						(35.0,0.00015117794723948995)
						(36.0,8.903389226440895e-5)
						(37.0,2.1779032012697065e-5)
						(38.0,2.5976380634272697e-5)
						(39.0,0.00011502520683109474)
						(40.0,2.4334975119212828e-5)
						(41.0,0.00015568616476922424)
						(42.0,2.8854851806767745e-5)
						(43.0,7.004025601137504e-6)
						(44.0,4.3264970746128715e-5)
						(45.0,7.514825352832682e-5)
						(46.0,5.658144977597373e-6)
						(47.0,9.693100136183071e-6)
						(48.0,0.00018315898846277156)
						(49.0,0.00045119168860653994)
						(50.0,3.327128539814575e-5)
						(51.0,2.79287466595607e-6)
						(52.0,7.007751002641762e-5)
						(53.0,2.138485510487298e-5)
						(54.0,6.838008432327031e-6)
						(55.0,7.070352043276507e-5)
					};
					\addplot[red,sharp plot,update limits=false, dashed, very thick]coordinates {(0.5,1e-2) (55.5,1e-2)};
				\end{axis}
			\end{tikzpicture}
		\end{adjustbox}
		\caption{}
	\end{subfigure}
	\caption{Global sensitivity indices $\varrho(\u, S(\X,\I_{\b N}(U_2))f^\text{cos})$ with $\bfh=\bfh^{\cos}$ using $\ell_1$-norm regularization (a), $\ell_2$-norm regularization (b) for the ten-dimensional function and corresponding best parameter choices, see Figure \ref{fig:7}, for $\u\in U_2$, where $|U_2|=56$. The resulting active set  is $U^{0.01}=\{\{1\}, \{2\}, \{3\}, \{4\}, \{5\}, \{1, 2\}\}$.}
	\label{fig:8}
\end{figure}
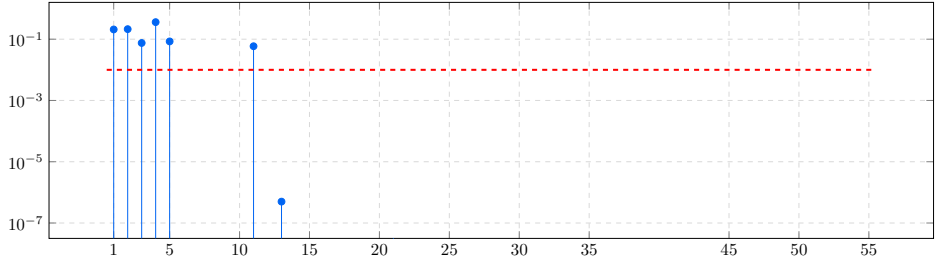
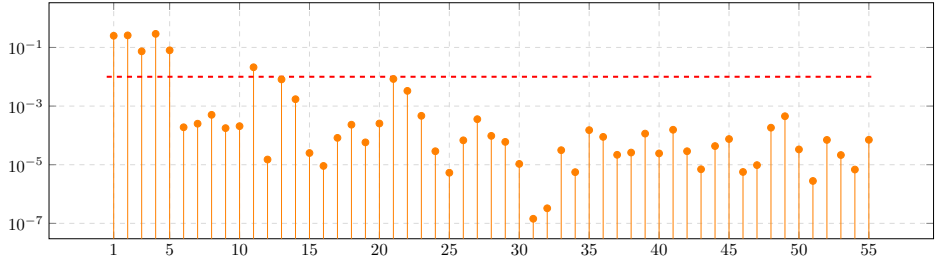

In Figure~\ref{fig:8} we plot the resulting global sensitivity indices, which clearly show that the variables $x_6$ to $x_{10}$ are significantly less important than the others.
By applying the threshold parameter $\varepsilon=0.01$ we are able to recover the actual active set
\begin{equation*}
	U^{0.01}=\{\emptyset,\{1\},\{2\},\{3\},\{4\},\{5\},\{1,2\}\}=U^\star.
\end{equation*}

Next, let us consider the wavelet basis approach and test if we are able to get the same result. 
In this setting, we generate $M=5000$ random feature vectors $\x_j\in \left[-\sfrac12,\sfrac12\right)^{10}$ and compute the corresponding labels by $y_j=\sign(f^{\chui})$ to train our prediction models.
Similarly to the cosine basis approach, we solve the considered optimization problems with different regularization parameters $\lambda$. Depending on different bandwidths $\b N$ from \eqref{eq:N_Phi_chui_f} with $N_1=3$ and $N_2\in\{1,2,3\}$, we have $\bfh^{\chui}\in\R^{376}$ if $N_2=1$, $\bfh^{\chui}\in\R^{916}$ if $N_2=2$ and $\bfh^{\chui}\in\R^{2356}$ for $N_2=3$.
We compute the classification accuracy on the test data set, containing $M$ random test data points.
Again, all accuracies have to be understood as average values achieved over $10$ runs.

\begin{figure}[ht!]%
	\begin{center}
		\begin{subfigure}[t]{0.45\linewidth}
			\begin{adjustbox}{width=\linewidth}
				\begin{tikzpicture}
					\begin{axis}[  ymin=80, ymax=95, xmode=log, xmin=0.000244140625, xmax=0.25,enlargelimits=0.05, grid = major,xtick={0.000244140625,0.0009765625, 0.00390625,0.015625,  0.0625,0.25}, xticklabels={$2^{-12}$,$2^{-10}$, $2^{-8}$, $2^{-6}$, $2^{-4}$, $2^{-2}$}, xlabel = $\lambda$, ylabel = CA, legend pos=south west]
						\addplot[blue35, mark=x, thick]  plot coordinates{
							(0.000244140625,92.554)
							(0.00048828125,92.61800000000001)
							(0.0009765625,92.782)
							(0.001953125,93.05799999999999)
							(0.00390625,93.304)
							(0.0078125,93.50199999999998)
							(0.015625,93.474)
							(0.03125,92.85600000000001)
							(0.0625,91.304)
							(0.125,87.856)
							(0.25,80.22000000000001)
						};
						\addlegendentry{$N_1=3, N_2=1$}
						\addplot[blue65, mark=x, thick]  plot coordinates{
							(0.000244140625,91.914)
							(0.00048828125,92.266)
							(0.0009765625,92.718)
							(0.001953125,93.18199999999999)
							(0.00390625,93.824)
							(0.0078125,94.41199999999999)
							(0.015625,94.58000000000001)
							(0.03125,93.824)
							(0.0625,91.726)
							(0.125,87.96799999999999)
							(0.25,80.22000000000001)
						};
						\addlegendentry{$N_1=3, N_2=2$}
						\addplot[blue100,  mark=x, thick]  plot coordinates{
							(0.000244140625,91.152)
							(0.00048828125,91.88399999999999)
							(0.0009765625,92.28400000000002)
							(0.001953125,92.618)
							(0.00390625,93.25800000000001)
							(0.0078125,93.96199999999999)
							(0.015625,94.434)
							(0.03125,93.97)
							(0.0625,91.754)
							(0.125,87.97200000000001)
							(0.25,80.22000000000001)
						};
						\addlegendentry{$N_1=3, N_2=3$}
					\end{axis}
				\end{tikzpicture}
			\end{adjustbox}
			\caption{}
		\end{subfigure}
		\hspace{5pt}%
		\begin{subfigure}[t]{0.45\linewidth}
			\begin{adjustbox}{width=\linewidth}
				\begin{tikzpicture}
					\begin{axis}[  ymin=80, ymax=95, xmode=log, xmin=0.000244140625, xmax=0.25,enlargelimits=0.05, grid = major,xtick={0.000244140625, 0.0009765625, 0.00390625,0.015625,  0.0625,0.25}, xticklabels={$2^{-12}$, $2^{-10}$, $2^{-8}$, $2^{-6}$, $2^{-4}$, $2^{-2}$}, xlabel = $\lambda$, ylabel = CA, legend pos=south west]
						\addplot[orange100, mark=x, thick]  plot coordinates{
							(0.000244140625,92.5)
							(0.00048828125,92.50399999999999)
							(0.0009765625,92.53)
							(0.001953125,92.56400000000001)
							(0.00390625,92.60199999999999)
							(0.0078125,92.51400000000001)
							(0.015625,92.376)
							(0.03125,92.196)
							(0.0625,91.946)
							(0.125,91.52799999999999)
							(0.25,91.094)
						};
						\addlegendentry{$N_1=3, N_2=1$}
						\addplot[orange65, mark=x, thick]  plot coordinates{
							(0.000244140625,91.19000000000001)
							(0.00048828125,91.122)
							(0.0009765625,90.708)
							(0.001953125,90.362)
							(0.00390625,90.47600000000001)
							(0.0078125,90.696)
							(0.015625,90.764)
							(0.03125,90.676)
							(0.0625,90.50399999999999)
							(0.125,90.18199999999999)
							(0.25,89.75600000000001)
						};
						\addlegendentry{$N_1=3, N_2=2$}
						\addplot[orange35,  mark=x, thick]  plot coordinates{
							(0.000244140625,90.62)
							(0.00048828125,90.838)
							(0.0009765625,90.028)
							(0.001953125,87.34600000000002)
							(0.00390625,85.72999999999999)
							(0.0078125,84.352)
							(0.015625,83.67)
							(0.03125,83.816)
							(0.0625,84.236)
							(0.125,84.57399999999998)
							(0.25,84.66799999999999)
						};
						\addlegendentry{$N_1=3, N_2=3$}
					\end{axis}
				\end{tikzpicture}
			\end{adjustbox}
			\caption{}
		\end{subfigure}
		\caption{Visualization of mean CA over 10 runs achieved by the ten-dimensional classifying functions $S(\X,\mathcal{J}_{\b N}(U_2))f^\chui(\x)$ with $\bfh=\bfh^{\chui}$, using different regularization parameters $\lambda$ and bandwidths parameters $\b N$, cf. \eqref{eq:N_Phi_chui_f}. The number of generated training and test data points have been set to $5000$.
			The results obtained with $\ell_1$-norm regularization are shown in (a) and results obtained  with $\ell_2$-norm regularization are shown in (b).
			\label{fig:9}}
	\end{center}
\end{figure}
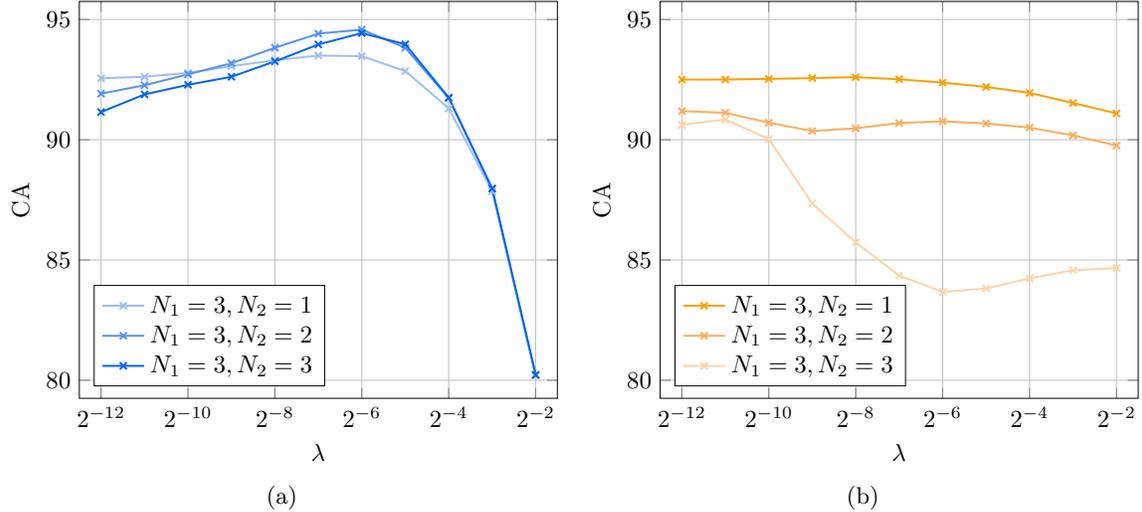

As before, we get different CA performance depending on the regularization. We achieve the highest mean CA with $\N_1=3$, $N_2=2$ and $\lambda = 2^{-6}$ for the $\ell_1$-norm regularization. For the best results provided by $\ell_2$-norm regularization we need to choose $N_1=3$, $N_2=1$ and $\lambda = 2^{-8}$.

\begin{figure}[ht]%
	\centering
	\begin{subfigure}[t]{0.75\linewidth}
		\begin{adjustbox}{width=\linewidth}
			\begin{tikzpicture}
				\begin{axis}[ width=200mm, height=6.5cm, ymode=log, log origin=infty,
					ymin=1e-7, ymax=1e0, xmin=0.5, xmax=55.5, enlargelimits=0.075, grid = major,
					grid style={dashed, gray!30}, xtick={1,5,10,15,20,25,30,35, 40, 45,50,55}, xticklabels={1,5,10,15,20,25,30,35, 40, 45,50,55}]
					\addplot+[ycomb, blue100, mark options={blue100}, mark=*] plot coordinates {
						(1.0,0.2138009446570672)
						(2.0,0.2019708618666535)
						(3.0,0.08796199868257953)
						(4.0,0.3443284875415569)
						(5.0,0.08906446736100129)
						(6.0,4.8700548132154965e-5)
						(7.0,0.00013837617265438794)
						(8.0,2.8883828413965726e-5)
						(9.0,6.393675301609851e-6)
						(10.0,2.3807867708446718e-5)
						(11.0,0.05839661788913703)
						(12.0,2.3217464833528984e-6)
						(13.0,0.0009776139201781453)
						(14.0,4.5440915186080455e-5)
						(15.0,0.00011925888017140096)
						(16.0,9.123183952866506e-5)
						(17.0,3.7299651418544424e-8)
						(18.0,2.0449577532887e-6)
						(19.0,8.27982596101147e-5)
						(20.0,3.4284974040068617e-7)
						(21.0,0.00047925962693246607)
						(22.0,0.0003177423113631978)
						(23.0,2.214196140784882e-5)
						(24.0,3.649691186076929e-6)
						(25.0,2.0202934518010926e-5)
						(26.0,3.918469736084499e-5)
						(27.0,1.860408903169184e-5)
						(28.0,3.8671413681020845e-5)
						(29.0,0.00020101309073439804)
						(30.0,8.024688582160969e-5)
						(31.0,4.492219618362875e-6)
						(32.0,3.092046558248743e-5)
						(33.0,2.3879925301500215e-5)
						(34.0,3.7598908628126294e-7)
						(35.0,0.0004993574833474652)
						(36.0,3.241773164741381e-6)
						(37.0,7.718674322699063e-5)
						(38.0,0.000178728750613208)
						(39.0,1.330584526116315e-5)
						(40.0,4.90798453285127e-6)
						(41.0,0.00019963051706077688)
						(42.0,3.54136125068923e-6)
						(43.0,0.0002332195973941091)
						(44.0,5.670879355039226e-5)
						(45.0,2.2203776259311e-5)
						(46.0,4.146223971501187e-5)
						(47.0,5.8070329184520616e-5)
						(48.0,4.2481028454090215e-5)
						(49.0,0.0)
						(50.0,4.056750889911216e-5)
						(51.0,1.9614204725406785e-5)
						(52.0,2.211685707308373e-5)
						(53.0,2.559693363652068e-5)
						(54.0,5.1174825834548554e-5)
						(55.0,3.586738668110455e-5)
					};
					\addplot[red,sharp plot,update limits=false, dashed, very thick] coordinates {(0.5,1e-2) (55.5,1e-2)};
				\end{axis}
			\end{tikzpicture}
		\end{adjustbox}
		\caption{}
		\hspace{5pt}%
	\end{subfigure}
	\begin{subfigure}[ht!]{0.75\linewidth}
		\begin{adjustbox}{width=\linewidth}
			\begin{tikzpicture}
				\begin{axis}[ width=200mm, height=6.5cm, ymode=log,log origin=infty,
					ymin=1e-7, ymax=1e0, xmin=0.5, xmax=55.5, enlargelimits=0.075, grid = major,
					grid style={dashed, gray!30}, xtick={1,5,10,15,20,25,30,35, 40,45,50,55}, xticklabels={1,5,10,15,20,25,30,35, 40,45,50,55}]
					\addplot+[ycomb, orange100, mark options={orange100}, mark=*] plot coordinates {
						(1.0,0.22317748749878544)
						(2.0,0.20907380976892673)
						(3.0,0.09207499706183796)
						(4.0,0.31843841563907793)
						(5.0,0.08640548300783493)
						(6.0,0.0011163478082069178)
						(7.0,0.0013674978384574036)
						(8.0,0.0014897315054023105)
						(9.0,0.0008808553588755158)
						(10.0,0.0008471265416994906)
						(11.0,0.03550259863476762)
						(12.0,0.0005596366033409821)
						(13.0,0.005843649065499698)
						(14.0,0.00183804805423826)
						(15.0,0.0004279253766617103)
						(16.0,0.0006725726070144442)
						(17.0,0.0002842297418878741)
						(18.0,0.00034572087044074936)
						(19.0,0.00040663876016658297)
						(20.0,0.00027759451918014515)
						(21.0,0.00468360191522359)
						(22.0,0.0024143443664245592)
						(23.0,0.0004395663135099815)
						(24.0,0.00011879966783270507)
						(25.0,0.0006409459228974007)
						(26.0,0.00031020267250691445)
						(27.0,0.00017618637793013136)
						(28.0,0.0004488683287193499)
						(29.0,0.000521769689645175)
						(30.0,0.00012865812705077368)
						(31.0,0.0003729391075267129)
						(32.0,0.00028772333487517645)
						(33.0,0.0003972448377380591)
						(34.0,0.00018458120001447475)
						(35.0,0.0009342682181454872)
						(36.0,0.00028790817402391216)
						(37.0,0.00017038731957580005)
						(38.0,0.0005971632784931723)
						(39.0,0.00015835060231562972)
						(40.0,0.00043598377890478743)
						(41.0,0.00016489159819755493)
						(42.0,0.0002709643432457087)
						(43.0,0.0009091414584748499)
						(44.0,0.00035871506549354744)
						(45.0,0.00010104504212951252)
						(46.0,0.00021402979978558535)
						(47.0,0.000646448871304425)
						(48.0,0.00024235413839968764)
						(49.0,0.0002748126102549035)
						(50.0,7.111663947393526e-5)
						(51.0,0.0007988438002729145)
						(52.0,0.00010635886895083579)
						(53.0,0.00013858739877920377)
						(54.0,0.0005446163215425124)
						(55.0,0.00041821454803824465)};
					\addplot[red,sharp plot,update limits=false, dashed, very thick]coordinates {(0.5,1e-2) (55.5,1e-2)};
				\end{axis}
			\end{tikzpicture}
		\end{adjustbox}
		\caption{}
	\end{subfigure}
	\caption{Global sensitivity indices $\varrho(\u, S(\X,\mathcal{J}_{\b N}(U_2))f^{\chui})$ of the ten-dimensional classifying function with $\bfh=\bfh^{\chui}$ using $\ell_1$-norm regularization (a), $\ell_2$-norm regularization (b) and corresponding best parameter choices, see Figure \ref{fig:9}, for $\u\in U_2$, where $|U_2|=56$. The resulting active set  is $U^{0.01}=\{\{1\}, \{2\}, \{3\}, \{4\}, \{5\}, \{1, 2\}\}$.}
	\label{fig:fr_gsi_chui}
\end{figure}
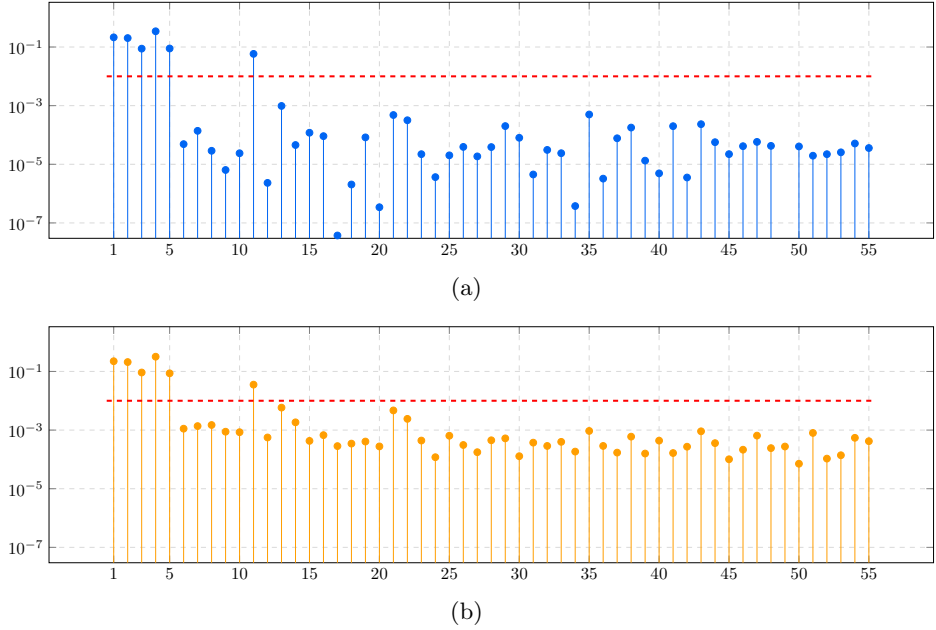

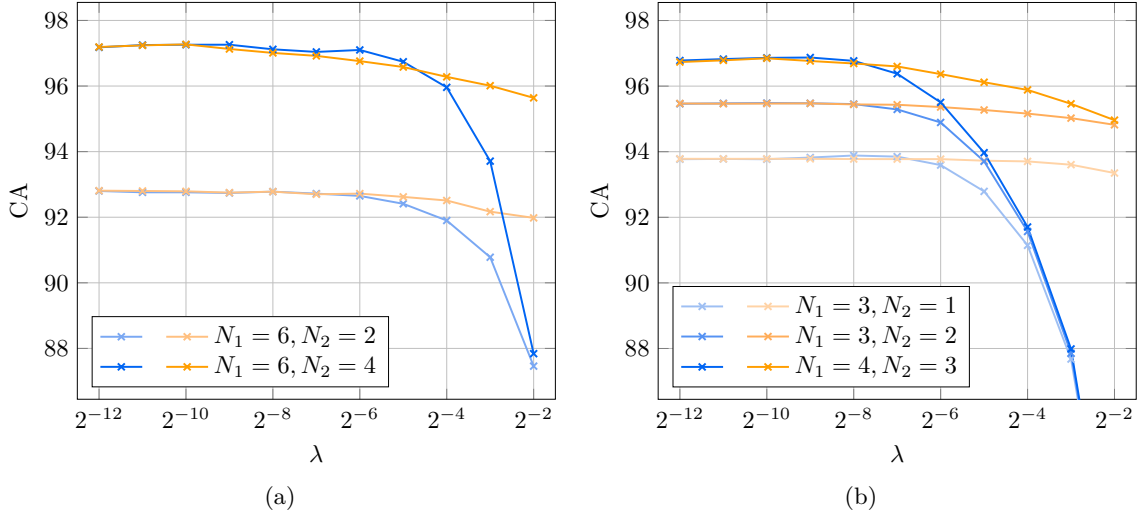
\begin{figure}[ht!]%
	\begin{center}
		\begin{subfigure}[t]{0.45\linewidth}
			\begin{adjustbox}{width=\linewidth}
				\begin{tikzpicture}
					\begin{axis}[  ymin=87, ymax=98, xmode=log, xmin=0.000244140625, xmax=0.25,enlargelimits=0.05, grid = major,xtick={0.000244140625,0.0009765625, 0.00390625,0.015625,  0.0625,0.25}, xticklabels={$2^{-12}$,$2^{-10}$, $2^{-8}$, $2^{-6}$, $2^{-4}$, $2^{-2}$}, xlabel = $\lambda$, ylabel = CA, legend pos=south west, legend columns=2, legend style={
							/tikz/column 2/.style={
								column sep=5pt}
						}]
						\addplot[blue50, mark=x, thick]  plot coordinates{
							(0.000244140625,92.8)
							(0.00048828125,92.76000000000002)
							(0.0009765625,92.76000000000002)
							(0.001953125,92.74000000000001)
							(0.00390625,92.78000000000002)
							(0.0078125,92.72)
							(0.015625,92.65)
							(0.03125,92.41)
							(0.0625,91.9)
							(0.125,90.78)
							(0.25,87.46000000000001)
						};
						\addlegendentry{ }
						\addplot[orange50, mark=x, thick]  plot coordinates{
							(0.000244140625,92.80999999999999)
							(0.00048828125,92.80000000000003)
							(0.0009765625,92.79)
							(0.001953125,92.75000000000003)
							(0.00390625,92.78)
							(0.0078125,92.70000000000002)
							(0.015625,92.72)
							(0.03125,92.62)
							(0.0625,92.51000000000002)
							(0.125,92.16999999999999)
							(0.25,91.98)
						};
						\addlegendentry{$N_1=6, N_2=2$}
						\addplot[blue100, mark=x, thick]  plot coordinates{
							(0.000244140625,97.18)
							(0.00048828125,97.24999999999999)
							(0.0009765625,97.26)
							(0.001953125,97.26)
							(0.00390625,97.11999999999999)
							(0.0078125,97.03999999999999)
							(0.015625,97.10000000000001)
							(0.03125,96.73999999999998)
							(0.0625,95.96000000000001)
							(0.125,93.71000000000001)
							(0.25,87.84)
						};
						\addlegendentry{}
						\addplot[orange100, mark=x, thick]  plot coordinates{
							(0.000244140625,97.19)
							(0.00048828125,97.24000000000001)
							(0.0009765625,97.27000000000001)
							(0.001953125,97.12999999999998)
							(0.00390625,97.00999999999999)
							(0.0078125,96.92)
							(0.015625,96.76)
							(0.03125,96.58)
							(0.0625,96.27999999999999)
							(0.125,96.01)
							(0.25,95.64)
						};
						\addlegendentry{$N_1=6, N_2=4$}
					\end{axis}
				\end{tikzpicture}
			\end{adjustbox}
			\caption{}
		\end{subfigure}
		\hspace{5pt}%
		\begin{subfigure}[t]{0.45\linewidth}
			\begin{adjustbox}{width=\linewidth}
				\begin{tikzpicture}
					\begin{axis}[  ymin=87, ymax=98, xmode=log, xmin=0.000244140625, xmax=0.25,enlargelimits=0.05, grid = major,xtick={0.000244140625,0.0009765625, 0.00390625,0.015625,  0.0625,0.25}, xticklabels={$2^{-12}$,$2^{-10}$, $2^{-8}$, $2^{-6}$, $2^{-4}$, $2^{-2}$}, xlabel = $\lambda$, ylabel = CA, legend pos=south west, legend columns=2, legend style={
							/tikz/column 2/.style={
								column sep=5pt}
						}]
						\addplot[blue35, mark=x, thick]  plot coordinates{
							(0.000244140625,93.772)
							(0.00048828125,93.78)
							(0.0009765625,93.77000000000001)
							(0.001953125,93.82199999999999)
							(0.00390625,93.884)
							(0.0078125,93.85)
							(0.015625,93.59200000000001)
							(0.03125,92.78999999999999)
							(0.0625,91.146)
							(0.125,87.678)
							(0.25,79.85)
						};
						\addlegendentry{}
						\addplot[orange35, mark=x, thick]  plot coordinates{
							(0.000244140625,93.78399999999999)
							(0.00048828125,93.782)
							(0.0009765625,93.78399999999999)
							(0.001953125,93.782)
							(0.00390625,93.77599999999998)
							(0.0078125,93.77399999999999)
							(0.015625,93.77399999999997)
							(0.03125,93.726)
							(0.0625,93.70400000000001)
							(0.125,93.604)
							(0.25,93.35)
						};
						\addlegendentry{$N_1=3, N_2=1$}
						\addplot[blue65, mark=x, thick]  plot coordinates{
							(0.000244140625,95.462)
							(0.00048828125,95.474)
							(0.0009765625,95.48199999999999)
							(0.001953125,95.47600000000001)
							(0.00390625,95.45199999999998)
							(0.0078125,95.29)
							(0.015625,94.892)
							(0.03125,93.708)
							(0.0625,91.568)
							(0.125,87.856)
							(0.25,79.85)
						};
						\addlegendentry{}
						\addplot[orange65, mark=x, thick]  plot coordinates{
							(0.000244140625,95.47)
							(0.00048828125,95.46400000000001)
							(0.0009765625,95.468)
							(0.001953125,95.474)
							(0.00390625,95.44200000000001)
							(0.0078125,95.43199999999999)
							(0.015625,95.362)
							(0.03125,95.272)
							(0.0625,95.162)
							(0.125,95.02399999999999)
							(0.25,94.82)
						};
						\addlegendentry{$N_1=3, N_2=2$}
						\addplot[blue100,  mark=x, thick]  plot coordinates{
							(0.000244140625,96.78)
							(0.00048828125,96.82399999999998)
							(0.0009765625,96.862)
							(0.001953125,96.872)
							(0.00390625,96.77)
							(0.0078125,96.378)
							(0.015625,95.50399999999999)
							(0.03125,93.97)
							(0.0625,91.70400000000001)
							(0.125,87.98599999999999)
							(0.25,79.94199999999998)
						};
						\addlegendentry{}
						\addplot[orange100,  mark=x, thick]  plot coordinates{
							(0.000244140625,96.734)
							(0.00048828125,96.786)
							(0.0009765625,96.84799999999998)
							(0.001953125,96.766)
							(0.00390625,96.69)
							(0.0078125,96.6)
							(0.015625,96.366)
							(0.03125,96.118)
							(0.0625,95.884)
							(0.125,95.46000000000001)
							(0.25,94.96200000000002)
						};
						\addlegendentry{$N_1=4, N_2=3$}
					\end{axis}
				\end{tikzpicture}
			\end{adjustbox}
			\caption{}
		\end{subfigure}
		\caption{Visualization of improved mean CA over 10 runs achieved by the ten-dimensional  classifying functions depending on the regularization parameter $\lambda$ and different bandwidths $\b N$. In (a) we see the results achieved by using the classifying functions $S(\X,\I_{\b N}(U^{0.01}))f^{\cos}(\x)$ with $\bfh=\bfh^{\cos}$. Results using the classifying functions $S(\X,\mathcal{J}_{\b N}(U^{0.01}))f^{\chui}(\x)$ with $\bfh=\bfh^{\chui}$ are shown in (b). Depending on the regularization we used shades of blue to visualize the results obtained by $\ell_1$-norm regularization and shades of of orange in case of $\ell_2$-norm regularization. We used $1000$ training and test data points for the cosine basis approach and $5000$ in case of wavelets.}
		\label{fig:10}
	\end{center}
\end{figure}

In Figure~\ref{fig:10} we have visualized the global sensitivity indices and get the same desired active set $U^{0.01}$. 
In addition, we use the active set $U^{0.01}$ to fit the classifying functions by using only this active set. Note that by using the bandwidths $\b N$ for the cosine approach with $N_1=6$ we get $\bfh^{\cos}\in\R^{26}$ for $N_2=2$, and $\bfh^{\cos}\in\R^{34}$ for $N_2=4$. 
In case of the wavelet approach we get $\bfh^{\chui}\in\R^{80}$ for $N_1=3$ and $N_2=1$, $\bfh^{\chui}\in\R^{92}$ for $N_1=3$, $N_2=2$.  Moreover, we can even choose higher bandwidths in order to obtain better results. As an example, with $N_1=4$ and $N_2=3$ we have $\bfh^{\chui}\in\R^{204}$, which improves the mean CA results. But, for instance, in case of the cosine basis approach we couldn't find any improvement by using higher bandwidths.

As expected, the $\ell_1$-norm regularization provides a sparse solution. 
In terms of accuracy the results for different basis approaches are essentially identical. 
However, we must highlight that the number of basis coefficients when using the wavelet basis approach in high dimensions can be very large even for small bandwidths, which is due to the locality of the basis functions.
Therefore, we will only focus on the cosine basis approach with $\ell_1$-norm regularization in the next section.

\subsection{Data sets \label{sec:datasets}}

Finally, we apply the proposed method to the following four data sets, see Table~\ref{table:1}.
The first two are synthetic benchmark data sets and will be discussed in more detail in Section~\ref{sec:synthetic} and the last two, which are real-world applications, in Section~\ref{sec:real}.

\begin{table}[ht]
	\centering
	\begin{tabular}[t]{lccc}
		\toprule
		dataset&$d$&$n$&$p$\\
		\midrule
		SUSY&$18$&$5.000.000$&$2.287.827$\\
		HIGGS&$28$&$11.000.000$&$5.170.877$\\
		\midrule
		Wisconsin Breast Cancer  (WBC)&$9$&$683$&$239$\\
		Pima Indians Diabetes (PID)&$8$&$768$&$268$\\
		\bottomrule
	\end{tabular}
	\vspace{1.5em}
	\caption{Basic information of synthetic and real-world data sets for numerical experiments, where $d$ is the dimension, $n$ is number of data points and $p$ is the percentage of data points with the positive label.}
	\label{table:1}
\end{table}

\subsubsection{Synthetic data \label{sec:synthetic}}
In this section, we take a look at the synthetically generated benchmark data sets SUSY and HIGGS that were analyzed using deep learning techniques in the original paper~\cite{Baldi2014}.

The first benchmark classification task is to distinguish between a first process where new supersymmetric particles (SUSY) are produced leading to a final state where some particles are detectable and others are invisible to the experimental apparatus, and a second background process with the same detectable particles but fewer invisible particles and different kinematic features.
This benchmark problem is currently of great interest in the field of high-energy physics, and there is a vigorous effort in the literature to construct high-level features that can aid in the classification task.
The second one is to distinguish between a signal process where new theoretical Higgs bosons (HIGGS) are produced, and a background process with the identical decay products but different kinematic features.
These simulated events are generated with the MadGraph generator~\cite{Alwall2011}, where in case of HIGGS the first $21$ and in case of SUSY the first $8$ features are kinematic properties measured by the particle detectors in the accelerator and the last $7$ and $10$ features, respectively, are functions of the first.
Moreover, the data sets are nearly balanced with $53\%$ positive examples in the HIGGS data set and $46\%$ positive examples in the SUSY data set.

We will compare the performance of our classification model in terms of CA with the results published in~\cite{NeStWa2023,WaPeSt2023}, where the authors examine these two data sets based on, among others, kernel ridge regression using ANOVA kernels, and in terms of AUC with the original paper~\cite{Baldi2014} and the current results from~\cite{enouen2023}.

\begin{figure}[!ht]
	\centering
	\begin{tikzpicture}
		\begin{axis}
			[width=5.5cm, height=5.5cm,
			boxplot/draw direction = x, boxplot/every median/.style={black,solid},yticklabel style={anchor=west,xshift=-4.85em}, ytick = {1, 2, 3}, 
			grid = major,grid style={dashed, gray!30}, 
			xtick={78.64, 79.82, 80.43}, xticklabels={$78.64$,$79.82$, $80.43$}, 
			yticklabels = {$\mathcal{I}_{(4,2)}(U_2)$, $\mathcal{I}_{(4,4)}(U_2)$,$\mathcal{I}_{(4,6)}(U_2)$},
			every axis plot/.append style = {fill, fill opacity = 0.1},cycle list={{black}}, 
			scatter/classes={a={mark=*, mark size=1, fill=black, fill opacity = 0.75}}, 
			xlabel = $\mathrm{CA}$]
			\addplot + [mark = *, boxplot, gray]table [row sep =\\, y index = 0] {79.75999999999999\\79.19\\79.75\\78.74\\79.34\\78.64\\79.73\\79.80000000000001\\79.2\\78.93\\};
			\addplot + [mark = *, boxplot, gray]table [row sep =\\, y index = 0] {79.93\\79.60000000000001\\80.0\\79.55\\79.95\\79.58\\80.07\\80.11\\79.82000000000001\\79.62\\};
			\addplot + [mark = *, boxplot, gray]table [row sep =\\, y index = 0] {80.31\\79.91\\79.96\\79.75999999999999\\79.80000000000001\\79.54\\80.01\\80.43\\79.83\\79.45\\};
			\addplot [scatter,only marks,scatter src=explicit symbolic] 
			table [
			y expr={1 + floor(\coordindex/1) + 1/3*mod(\coordindex,1)},
			x=x,
			meta=meta
			]
			{
				x meta		
				79.30800000000002 a
				79.82300000000001 a
				79.9 a
				
			};
		\end{axis}
	\end{tikzpicture}
	\begin{tikzpicture}
		\begin{axis}
			[width=5.5cm, height=5.5cm,
			boxplot/draw direction = x, boxplot/every median/.style={black,solid}, 
			yticklabel style={anchor=west,xshift=-4.85em}, ytick = {1, 2, 3}, 
			grid = major,grid style={dashed, gray!30}, 
			xtick={0.8557, 0.8668, 0.8724}, xticklabels={$0.8557$,$0.8668$, $0.8724$}, 
			yticklabels = {$ $, $ $, $ $},
			every axis plot/.append style = {fill, fill opacity = 0.1},cycle list={{black}}, 
			scatter/classes={a={mark=*, mark size=1, fill=black, fill opacity = 0.75}}, 
			xlabel = $\mathrm{AUC}$]
			]
			\addplot + [mark = *, boxplot, gray]table [row sep =\\, y index = 0] {0.8654184678248453\\0.8613143870480316\\0.8652992152097199\\0.8618030573371269\\0.8599759941672419\\0.8615059985885674\\0.8635180192705249\\0.866934223969175\\0.8625907853614275\\0.855753480716307\\};
			
			\addplot + [mark = *, boxplot, gray]table [row sep =\\, y index = 0] {0.8681863356664059\\0.8656783664874773\\0.8685435717175631\\0.8685132050601984\\0.8647228260307512\\0.8668664583123299\\0.8667527835874125\\0.8705654914297535\\0.867972981147293\\0.8605473066229381\\};
			\addplot + [mark = *, boxplot, gray]table [row sep =\\, y index = 0] {0.8724772558153117\\0.8686889879778417\\0.8707281636129688\\0.8699684943177252\\0.8669288951628958\\0.8675601169472729\\0.8670864916925581\\0.8716854874768262\\0.8693457001713882\\0.8633772123856762\\};
			
			\addplot [scatter,only marks,scatter src=explicit symbolic] 
			table [y expr={1 + floor(\coordindex/1) + 1/3*mod(\coordindex,1)}, x=x, meta=meta]
			{
				x meta		
				0.8624113629492969 a
				0.8668349326062124 a
				0.8687846805560465 a
			};
		\end{axis}
	\end{tikzpicture}
	\caption{Achieved highest (among all considered regularization parameters $\lambda$) CA and AUC for the SUSY data set, averaged over $10$ runs for different index sets.
		We applied the cosine basis approach with $\ell_1$-norm regularization. For each run we choose randomly $10^4$ data points for training the model and the same number of data points for testing.}
	\label{fig:susy}
\end{figure}
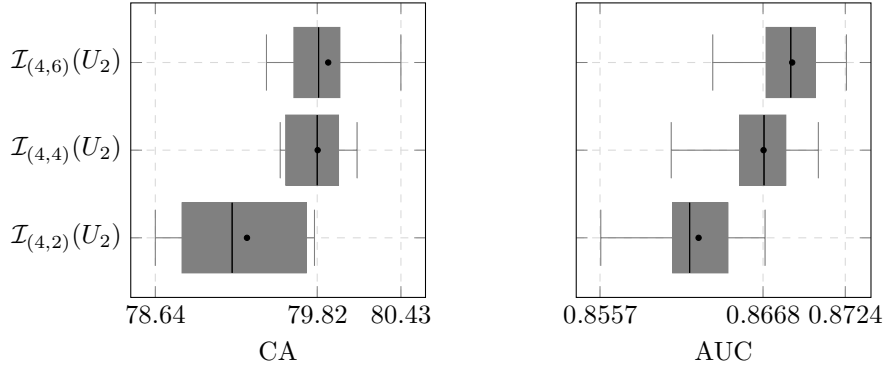

First, we examine the SUSY data set. We select different index sets and compute the highest CA and AUC depending on the regularization parameter $\lambda\in\{2^{-l}: l=1,\dots,5\}\cup\{0.0\}$.
Similar to the previous section, we set the superposition threshold to $d_s=2$ and use the notation
\begin{equation*}
	\b \Phi^{\cos}:=\b \Phi^{\cos}(\X, \I_{\b N}(U_2)),\quad \b N^{\u}=\begin{cases}
		(N_1)&:\quad\abs{\u}=1,\\
		(N_2, N_2)&:\quad\abs{\u}=2.
	\end{cases}
\end{equation*}
Note that in this section we only consider the cosine basis approach with $\ell_1$-norm regularization.
We randomly choose $M=10^4$ training data points $(\x_j, y_j)$, $j=1,2,\dots,M$, and solve the considered optimization problems with different regularization parameters $\lambda$.
Depending on the applied bandwidths $\b N$ with $N_1=4$ and $N_2\in\{2,4,6\}$, we need to compute $\bfh^{\cos}\in\R^{208}$ if $N_2=2$, $\bfh^{\cos}\in\R^{1432}$ if $N_2=4$ and $\bfh^{\cos}\in\R^{3880}$ for $N_2=6$.
Using the optimized basis coefficients, we compute the CA and AUC using the corresponding classifying functions $S(\X,\I_{\b N}(U_2))f^\text{cos}(\x)$ and the selected test data set, containing $M$ test data points.
The box plot is the best way to get a good impression of the achieved results and to see at a glance the median, the mean, the lowest and the highest achieved CA or rather AUC over $10$ runs, where for each run we only take the best result across all regularization parameters $\lambda$.
In Figure~\ref{fig:susy} we visualize the results for the achieved CA and AUC for the SUSY data set.

We can see that our method achieves approximately the same or a slightly higher CA than that in~\cite{NeStWa2023}.
The results for the achieved AUC are also good, compared to~\cite{Baldi2014}. Note that we only used $10^4$ data points for training. 

Since the results for $N_2=4$ and $N_2=6$ are quite similar, we will now only consider only $S(\X,\I_{\b N}(U_2))f^\text{cos}(\x)$ with $N_1=N_2=4$ in order to take a closer look onto the resulting GSIs of the classifying function.
For example, in Figure \ref{fig:susy_gsi} we plot the GSIs of all $171$ ANOVA terms in sorted order, computed with the fixed regularization parameter $\lambda=2^{-4}$, which gives us the highest CA for the $M=10^{4}$ training data points. 
Note that in this case we achieve a CA of $79.7\%$ and an AUC of $0.8554$ on the $M$ test data points.

\begin{figure}
	\centering
	\begin{tikzpicture}
		\begin{axis}
			[width=17.5cm, height=7cm, xmin=1, xmax=171, enlargelimits=0.02, grid = major, grid style={dashed, gray!30}, ymode=log, log origin=infty, ymin=0.0001, xtick={1, 25, 50, 75, 100, 125, 150, 171}, xticklabels={$1$, $25$, $50$, $75$, $100$, $125$, $150$, $171$}]
			\coordinate (pt) at (axis cs:50,0.000495);
			\addplot+[ycomb, orange, mark options={orange}, mark size=0.75pt, mark=*]coordinates {(1,0.43063791927733946)(2,0.22073823787841113)(3,0.053537802839044)(4,0.037363470940172686)(5,0.025766341546569732)(6,0.023812681007342885)(7,0.021409515751783206)(8,0.018673724084414665)(9,0.016706104609652238)(10,0.01572289176088342)(11,0.01403679757652431)(12,0.013675532333056638)(13,0.011256973444178994)};
			\addplot+[ycomb, white, mark options={white}, mark size=0.75pt, mark=*]coordinates {(1.5,0.65)};
			\addplot+[ycomb, gray, mark options={gray}, mark size=0.75pt, mark=*]coordinates {(14,0.008539645918379645)(15,0.008448610666204953)(16,0.007642571501271495)(17,0.006720625394577422)(18,0.0065115253766176865)(19,0.0061597628560080084)(20,0.006022076451596986)(21,0.005613701570741086)(22,0.005565954761564281)(23,0.004833889368798426)(24,0.004458309444599068)(25,0.0038890374036719784)(26,0.0029995754667835736)(27,0.002984998102832247)(28,0.002765536313118568)(29,0.001831969120272857)(30,0.0015047207824379822)(31,0.0014015469077131222)(32,0.0013083097017822791)(33,0.001288864580040484)(34,0.0011852307605175986)(35,0.0011041833322269195)(36,0.0010426343583115505)(37,0.0007148693581809371)(38,0.0006946991736411352)(39,0.00034521072298237735)(40,0.00030987171387226363)(41,0.0002565103752822564)(42,0.00022047524304276428)(43,0.00013028252180804698)(44,7.048766044395016e-5)(45,5.99148554920306e-5)(46,2.0951313940869784e-5)(47,1.5453871871996816e-5)(48,0.0)(49,0.0)(50,0.0)(51,0.0)(52,0.0)(53,0.0)(54,0.0)(55,0.0)(56,0.0)(57,0.0)(58,0.0)(59,0.0)(60,0.0)(61,0.0)(62,0.0)(63,0.0)(64,0.0)(65,0.0)(66,0.0)(67,0.0)(68,0.0)(69,0.0)(70,0.0)(71,0.0)(72,0.0)(73,0.0)(74,0.0)(75,0.0)(76,0.0)(77,0.0)(78,0.0)(79,0.0)(80,0.0)(81,0.0)(82,0.0)(83,0.0)(84,0.0)(85,0.0)(86,0.0)(87,0.0)(88,0.0)(89,0.0)(90,0.0)(91,0.0)(92,0.0)(93,0.0)(94,0.0)(95,0.0)(96,0.0)(97,0.0)(98,0.0)(99,0.0)(100,0.0)(101,0.0)(102,0.0)(103,0.0)(104,0.0)(105,0.0)(106,0.0)(107,0.0)(108,0.0)(109,0.0)(110,0.0)(111,0.0)(112,0.0)(113,0.0)(114,0.0)(115,0.0)(116,0.0)(117,0.0)(118,0.0)(119,0.0)(120,0.0)(121,0.0)(122,0.0)(123,0.0)(124,0.0)(125,0.0)(126,0.0)(127,0.0)(128,0.0)(129,0.0)(130,0.0)(131,0.0)(132,0.0)(133,0.0)(134,0.0)(135,0.0)(136,0.0)(137,0.0)(138,0.0)(139,0.0)(140,0.0)(141,0.0)(142,0.0)(143,0.0)(144,0.0)(145,0.0)(146,0.0)(147,0.0)(148,0.0)(149,0.0)(150,0.0)(151,0.0)(152,0.0)(153,0.0)(154,0.0)(155,0.0)(156,0.0)(157,0.0)(158,0.0)(159,0.0)(160,0.0)(161,0.0)(162,0.0)(163,0.0)(164,0.0)(165,0.0)(166,0.0)(167,0.0)(168,0.0)(169,0.0)(170,0.0)(171,0.0)};
			
			\addplot[darkgray,sharp plot,update limits=false, densely dashed, thin] coordinates {(0.0,0.01) (14.0,0.01)};
			\addplot[darkgray,sharp plot,update limits=false, densely dashed, thin] coordinates {(0.0,0.65) (14.0,0.65)};
			\addplot[darkgray,sharp plot,update limits=false, densely dashed, thin] coordinates {(0.0,0.01) (0.0,0.65)};
			\addplot[darkgray,sharp plot,update limits=false, densely dashed, thin] coordinates {(14.0,0.01)(14.0,0.65)};
			
			\addplot[darkgray,sharp plot,update limits=false,  densely dashed, thin] coordinates {(14.0,0.65)(62.0, 0.172)};
			\addplot[darkgray,sharp plot,update limits=false, densely dashed, thin] coordinates {(14.0,0.01)(62.0,0.00125)};
		\end{axis}
		\node[anchor=south west] at (pt) {
			\begin{tikzpicture}
				\begin{axis}[axis background/.style={fill=white!10}, footnotesize, width=11cm, height=4.5cm, xmin=1,  xmax=13, xtick = {1, 5, 10, 13 }, ytick = {0.1, 0.01}, yticklabels = {$10^{-1}$, $10^{-2}$}, ymode=log,log origin=infty, ymin = 0.0055, xtick = {1, 5, 10, 13},xticklabels={$1$, $5$, $10$, $13$}, enlargelimits=0.05]
					\addplot+[white,mark size=0.1pt] coordinates {	(1.5,3)};
					\addplot+[ycomb, orange, mark options={orange}, mark size=1.3pt, mark=*] plot coordinates {(1,0.43063791927733946)(2,0.22073823787841113)(3,0.053537802839044)(4,0.037363470940172686)(5,0.025766341546569732)(6,0.023812681007342885)(7,0.021409515751783206)(8,0.018673724084414665)(9,0.016706104609652238)(10,0.01572289176088342)(11,0.01403679757652431)(12,0.013675532333056638)(13,0.011256973444178994)};
					\node at (axis cs:1, 4.1) [anchor=north] {\rotatebox{90}{\scriptsize{$\{1,7\}$}}};
					\node at (axis cs:2,4.1) [anchor=north] {\rotatebox{90}{\scriptsize{$\{10,12\}$}}};
					\node at (axis cs:3,0.75) [anchor=north] {\rotatebox{90}{\scriptsize{$\{7,17\}$}}};
					\node at (axis cs:4,0.55) [anchor=north] {\rotatebox{90}{\scriptsize{$\{4,11\}$}}};
					\node at (axis cs:5,0.4) [anchor=north] {\rotatebox{90}{\scriptsize{$\{4,15\}$}}};
					\node at (axis cs:6,0.3) [anchor=north] {\rotatebox{90}{\scriptsize{$\{4,7\}$}}};
					\node at (axis cs:7,0.3) [anchor=north] {\rotatebox{90}{\scriptsize{$\{1,12\}$}}};
					\node at (axis cs:8,0.3) [anchor=north] {\rotatebox{90}{\scriptsize{$\{4,13\}$}}};
					\node at (axis cs:9,0.3) [anchor=north] {\rotatebox{90}{\scriptsize{$\{7,10\}$}}};
					\node at (axis cs:10,0.35) [anchor=north] {\rotatebox{90}{\scriptsize{$\{13,15\}$}}};
					\node at (axis cs:11,0.2) [anchor=north] {\rotatebox{90}{\scriptsize{$\{2,11\}$}}};
					\node at (axis cs:12,0.3) [anchor=north] {\rotatebox{90}{\scriptsize{$\{11,13\}$}}};
					\node at (axis cs:13,0.125) [anchor=north] {\rotatebox{90}{\scriptsize{$\{2,4\}$}}};
				\end{axis}
			\end{tikzpicture}
		};
	\end{tikzpicture}
	\caption{Computed global sensitivity indices $\varrho(\u, S(\X,\I_{\b N}(U_2))f^\text{cos})$ with $\bfh=\bfh^{\cos}$ using $N_1=N_2=4$, $\lambda=2^{-4}$ and $M=10^4$, sorted according to the importance, for the SUSY data set.}
	\label{fig:susy_gsi}
\end{figure}

Furthermore, based on the calculated GSIs, we are able to rank the variable interactions according to their importance. 
As we can see, only $47$ of the ANOVA terms have any effect on the result in this setting. 
We highlight the most important ANOVA terms in orange and show the corresponding variable interactions in this particular case.
Note that this result depends on the chosen training data points as well as on the regularization parameter.
Now our question is whether we can get meaningful results with fewer coefficients than $1432$, which are much faster to compute.
Therefor, using the same training points as before and only the first $13$ ANOVA terms, i.e., using the active set $U^{0.01}=U^\star$ highlighted in Figure~\ref{fig:susy_gsi}, we need to compute only $118$ basis coefficients. By doing this, we are still able to achieve a CA of $76.46\%$ and an AUC of $0.8438$.
Even if we choose a larger threshold parameter and, thus
\begin{equation*}
	U^{0.02}=\{\emptyset,\{1,7\},\{10,12\},\{7,17\},\{4,11\},\{4,15\},\{4,7\},\{1,12\}\}=U^\star,
\end{equation*}
we are still able to achieve a CA of $77.11\%$ and an AUC of $0.8414$ using only $64$ basis coefficients.

In the following, we will examine the HIGGS data set in a similar way. We randomly choose $M= 5\cdot10^4$ training data points $(\x_j, y_j)$, $j=1,2,\dots,M$, and solve the considered optimization problem with different regularization parameters $\lambda\in\{2^{-l}: l=5,\dots,10\}\cup\{0.0\}$ and different bandwidths $\b N$ with $N_1=4$ and $N_2\in\{2,4,6\}$. 
Due to the higher dimension of the HIGGS data set, the number of coefficients to be computed is higher than in the previous example. 
Using $\bfh^{\cos}\in\R^{463}$ if $N_2=2$, $\bfh^{\cos}\in\R^{3487}$ if $N_2=4$ and $\bfh^{\cos}\in\R^{9535}$ for $N_2=6$ we construct the classifying function $S(\X,\I_{\b N}(U_2))f^\text{cos}$. 
Figure\ref{fig:higgs} shows the achieved highest CA or rather AUC over $10$ runs, where for each run we only take the best result across all regularization parameters $\lambda$.

\begin{figure}[!ht]
	\centering
	\begin{tikzpicture}
		\begin{axis}
			[width=5.5cm, height=5.5cm,
			boxplot/draw direction = x, boxplot/every median/.style={black,solid},yticklabel style={anchor=west,xshift=-4.85em}, ytick = {1, 2, 3}, 
			grid = major,grid style={dashed, gray!30}, 
			xtick={64.37, 66.25, 67.74}, xticklabels={$64.37$,$66.25$, $67.74$}, 
			yticklabels = {$\mathcal{I}_{(4,2)}(U_2)$, $\mathcal{I}_{(4,4)}(U_2)$,$\mathcal{I}_{(4,6)}(U_2)$},
			every axis plot/.append style = {fill, fill opacity = 0.1},cycle list={{black}}, 
			scatter/classes={a={mark=*, mark size=1, fill=black, fill opacity = 0.75}}, 
			xlabel = $\mathrm{CA}$]
			\addplot + [mark = *, boxplot, gray]table [row sep =\\, y index = 0] 
			{64.536\\64.634\\63.885999999999996\\64.022\\63.846000000000004\\65.244\\64.546\\64.19399999999999\\64.352\\64.53\\};
			\addplot + [mark = *, boxplot, gray]table [row sep =\\, y index = 0] {66.334\\66.31\\66.122\\66.146\\65.836\\66.658\\66.10000000000001\\66.25999999999999\\66.328\\66.47\\};
			\addplot + [mark = *, boxplot, gray]table [row sep =\\, y index = 0] {68.052\\68.132\\67.416\\67.56\\67.212\\68.21000000000001\\67.754\\67.67800000000001\\67.658\\67.804\\};
			\addplot [scatter,only marks,scatter src=explicit symbolic] 
			table [
			y expr={1 + floor(\coordindex/1) + 1/3*mod(\coordindex,1)},
			x=x,
			meta=meta
			]
			{
				x meta		
				64.37899999999999 a
				66.25640000000001 a
				67.7476 a         
			};
		\end{axis}
	\end{tikzpicture}
	\begin{tikzpicture}
		\begin{axis}
			[width=5.5cm, height=5.5cm,
			boxplot/draw direction = x, boxplot/every median/.style={black,solid}, 
			yticklabel style={anchor=west,xshift=-4.85em}, ytick = {1, 2, 3}, 
			grid = major,grid style={dashed, gray!30}, 
			xtick={0.6943, 0.7187, 0.7367}, xticklabels={$0.6943$,$0.7187$, $0.7367$}, 
			yticklabels = {$ $, $ $, $ $},
			every axis plot/.append style = {fill, fill opacity = 0.1},cycle list={{black}}, 
			scatter/classes={a={mark=*, mark size=1, fill=black, fill opacity = 0.75}}, 
			xlabel = $\mathrm{AUC}$]
			]
			\addplot + [mark = *, boxplot, gray]table [row sep =\\, y index = 0] {0.6980430308913553\\0.6970198747322813\\0.6945672540629287\\0.6928340065854856\\0.6908756109892462\\0.7030758507703613\\0.6926480516847606\\0.6942906105442367\\0.6980550351235637\\0.6971665553337467\\};
			
			\addplot + [mark = *, boxplot, gray]table [row sep =\\, y index = 0] {0.7218767297520187\\0.721353092352506\\0.7174761614258396\\0.718108530025917\\0.7132549081713028\\0.7237314814145576\\0.713979437006818\\0.7193812118871274\\0.7196040021492011\\0.7191896154395635\\};
			\addplot + [mark = *, boxplot, gray]table [row sep =\\, y index = 0] {0.741030237957341\\0.742806724157504\\0.7340247429131308\\0.7343054125695978\\0.7295647441214022\\0.7417776348282799\\0.7342004102589346\\0.7361169749619411\\0.7351190683657138\\0.7384589307756558\\};
			
			\addplot [scatter,only marks,scatter src=explicit symbolic] 
			table [y expr={1 + floor(\coordindex/1) + 1/3*mod(\coordindex,1)}, x=x, meta=meta]
			{
				x meta		
				0.6943576878587493 a
				0.7187955169624851 a
				0.7367404880909503 a
			};
		\end{axis}
	\end{tikzpicture}
	\caption{Achieved highest (among all considered regularization parameters $\lambda$) CA and AUC for the HIGGS data set, averaged over $10$ runs for different index sets.
		We applied the cosine basis approach with $\ell_1$-norm regularization. For each run, we randomly select $5\cdot 10^4$ data points for training the model and the same number of data points for testing.}
	\label{fig:higgs}
\end{figure}

\begin{figure}
	\centering
	\begin{tikzpicture}
		\begin{axis}
			[width=17.5cm, height=7cm, xmin=1, xmax=406, enlargelimits=0.02, grid = major, grid style={dashed, gray!30}, ymode=log, log origin=infty, ymin=0.000001, xtick={1, 50, 100, 150, 200, 250, 300, 350, 400}, xticklabels={$1$, $50$, $100$, $150$, $200$, $250$, $300$, $350$, $406$}]
			\coordinate (pt) at (axis cs:140,0.000525);
			\addplot+[ycomb, orange, mark options={orange}, mark size=0.5pt, mark=*]coordinates {(1,0.05345307855798923)(2,0.032234093136252244)(3,0.03212794264021359)(4,0.03185068243659376)(5,0.030982733026525423)(6,0.029979908469847952)(7,0.027222070835683305)(8,0.025130900179535915)(9,0.024344799154963903)(10,0.022114861936327758)(11,0.021311219244518162)(12,0.020683005648822545)(13,0.02024712982403898)(14,0.019071277670052844)(15,0.018333676739135413)(16,0.01786955598344579)(17,0.01786369569332421)(18,0.016785306294709444)(19,0.015158598430717702)(20,0.0131473157492675)(21,0.012919309213475061)};
			\addplot+[ycomb, white, mark options={white}, mark size=0.5pt, mark=*]coordinates {(1.5,0.75)};
			\addplot+[ycomb, gray, mark options={gray}, mark size=0.5pt, mark=*]coordinates {(22,0.010028967657883437)(23,0.009520176578402168)(24,0.009266338097268286)(25,0.008998595520114172)(26,0.008574903859723464)(27,0.008037441472125739)(28,0.008019202919744714)(29,0.00759729194155722)(30,0.007241297603401202)(31,0.007012088809716523)(32,0.00669338700874535)(33,0.0064944389894319166)(34,0.00612644418530372)(35,0.005979482210828568)(36,0.005933458079236098)(37,0.005904209736630344)(38,0.005750392877371606)(39,0.005737330088628557)(40,0.005720877441516037)(41,0.005682372663417436)(42,0.005607121801643183)(43,0.0052458199829298145)(44,0.00524367770087939)(45,0.005047341579975958)(46,0.004871261165701407)(47,0.004749999478149539)(48,0.004733774488685418)(49,0.0046410449329164155)(50,0.0046241244936442915)(51,0.004606990624798036)(52,0.004572189834167498)(53,0.004563017229146276)(54,0.004531774337605788)(55,0.004494826887371649)(56,0.004271617023236966)(57,0.004100453722781084)(58,0.004087107284850535)(59,0.004042203470156302)(60,0.004022039252651022)(61,0.0039244584410701245)(62,0.003424326619844631)(63,0.0034233680722742877)(64,0.0032750145968554194)(65,0.0031997284757159104)(66,0.003164634855061997)(67,0.003120266020449495)(68,0.0030393094223390704)(69,0.0030091951306769477)(70,0.002982399086956661)(71,0.0029663401968379383)(72,0.0028790109963283744)(73,0.0027829032425548137)(74,0.0027227331199685266)(75,0.0027063999630834506)(76,0.0026885959065632613)(77,0.0026495817018626234)(78,0.002585705357775541)(79,0.00257025547916863)(80,0.0025505838757755874)(81,0.002489231379453636)(82,0.002481003137555506)(83,0.002392681925816701)(84,0.0023381892209699863)(85,0.0022887518299337375)(86,0.002197006283471427)(87,0.0021841160329535863)(88,0.0021625064052836697)(89,0.0021574708002691473)(90,0.00215667264511056)(91,0.002144914764452135)(92,0.0021308574564660424)(93,0.0021293417848136827)(94,0.0020906287824616005)(95,0.0020901534119564156)(96,0.0020847283570967993)(97,0.00207217469638715)(98,0.0020392048450029233)(99,0.002000524247780636)(100,0.0019907770832188697)(101,0.001969764909837881)(102,0.001934949886899666)(103,0.0018276303774386318)(104,0.0018187673167051586)(105,0.0018039324649779835)(106,0.0017897933011398106)(107,0.0017685463581768401)(108,0.0017275790378939286)(109,0.0016952228756380357)(110,0.0016475542895805888)(111,0.0015982127647565403)(112,0.0015963415378787247)(113,0.001577310031163493)(114,0.0015587023291976187)(115,0.0015561385685619827)(116,0.001553957299691328)(117,0.0015398487606383974)(118,0.0015370916183981595)(119,0.001518027672336803)(120,0.0015009070980305209)(121,0.001493276716249383)(122,0.001479880411924853)(123,0.0014642370610952484)(124,0.0014483225217043333)(125,0.00143975389871631)(126,0.0014378344222169882)(127,0.0014287978880476696)(128,0.0013410168527446002)(129,0.0013329171077875186)(130,0.0013292589394569034)(131,0.0013257123465065356)(132,0.0012961180247630228)(133,0.0012803608986963104)(134,0.0012707239636182476)(135,0.001265419944751873)(136,0.0012534716238384344)(137,0.0012533319617487057)(138,0.001235769818990512)(139,0.0012336125655405122)(140,0.0012329809725924756)(141,0.0012139098267045815)(142,0.0012011430291429287)(143,0.0011952797123384376)(144,0.0011950047037442878)(145,0.001179764893234473)(146,0.0011677100798553291)(147,0.0011632956271397743)(148,0.0011572345791101893)(149,0.0011567050336075255)(150,0.0011260851691818606)(151,0.0011174309435364735)(152,0.0011074265537127817)(153,0.0010982701091824245)(154,0.0010936902725817066)(155,0.0010721704357456463)(156,0.0010495162646692003)(157,0.0010434748532264747)(158,0.001031496779096118)(159,0.0010212185924366555)(160,0.001018210231184677)(161,0.0010066084726672755)(162,0.0010013868586201762)(163,0.0009963641561866348)(164,0.0009628722646194842)(165,0.0009627826953678225)(166,0.0009513812309853749)(167,0.0009370739847207346)(168,0.0009356533565236385)(169,0.0009316000523908186)(170,0.0009248405703286597)(171,0.00091875800845646)(172,0.0009043973536689767)(173,0.0009029564638574213)(174,0.0008947750901199467)(175,0.0008946779692446276)(176,0.0008901432176864975)(177,0.000887006949260402)(178,0.0008717500523024386)(179,0.0008686622796008745)(180,0.0008605043599795914)(181,0.0008576462453715998)(182,0.0008360789264539551)(183,0.0008337093549209948)(184,0.0008123264948936645)(185,0.0007935478117663818)(186,0.0007928695324005721)(187,0.0007824451289753754)(188,0.0007820537054649991)(189,0.0007815026605616058)(190,0.0007799104088790858)(191,0.000754580618387631)(192,0.0007442772896234524)(193,0.0007256381124939945)(194,0.0007176757845986147)(195,0.0007076067679387807)(196,0.0006755646819460272)(197,0.00067485831963947)(198,0.0006745357339358251)(199,0.0006334644444217135)(200,0.000623591088661061)(201,0.0005986180290056599)(202,0.0005970760475629386)(203,0.0005941450310632149)(204,0.0005894717943933758)(205,0.0005869926492149441)(206,0.0005859902059513226)(207,0.0005819526780600373)(208,0.0005805951719010935)(209,0.0005772965910921423)(210,0.0005760226508973933)(211,0.0005709543312142901)(212,0.0005708386167870919)(213,0.0005619322378378599)(214,0.0005607423021507001)(215,0.000554018048366775)(216,0.0005485750534646802)(217,0.0005306555482467436)(218,0.0005272007262300532)(219,0.0005213279119509345)(220,0.0005151402206225741)(221,0.0005129024684562638)(222,0.000499949375127003)(223,0.0004944980326856166)(224,0.000489517687934164)(225,0.00048530949455435766)(226,0.00048383311730649956)(227,0.0004813893603832923)(228,0.000479062444653564)(229,0.0004581798684002454)(230,0.00045264799242584927)(231,0.0004434539515273429)(232,0.0004419609706778428)(233,0.00043465574932848163)(234,0.000432224123215804)(235,0.0004313053141309259)(236,0.0004243319985714518)(237,0.0004227095598270856)(238,0.00042125708652318035)(239,0.0004120776015562579)(240,0.00041143932271282564)(241,0.0003961182096458601)(242,0.00037867473283212383)(243,0.0003752008712130218)(244,0.00037416522760968455)(245,0.0003646897156902746)(246,0.00035726514532391827)(247,0.0003518915992361348)(248,0.0003398730582062298)(249,0.00033960802751238937)(250,0.00033769168671710716)(251,0.0003371727750415649)(252,0.0003350949766329436)(253,0.00033497494470923134)(254,0.00033399081773546744)(255,0.00032894706606307174)(256,0.00032818872215667537)(257,0.000327336644570104)(258,0.0003264103607069179)(259,0.00032502564175944897)(260,0.0003244579419478393)(261,0.0003186417764210271)(262,0.00031596175548014266)(263,0.00030409468990000245)(264,0.00030345372021434527)(265,0.0002983589060711151)(266,0.0002958928907366009)(267,0.00029580392975587974)(268,0.00029402442952248934)(269,0.0002903734474316086)(270,0.0002867239856668858)(271,0.000286516522250201)(272,0.0002857606513156233)(273,0.00028454444604194624)(274,0.0002815477732050509)(275,0.00027571617867431006)(276,0.0002743154652563477)(277,0.0002725745719434312)(278,0.0002710845465529579)(279,0.00026745073083542215)(280,0.00025507913202716423)(281,0.0002532154761199378)(282,0.00025058911781121045)(283,0.000248114324985674)(284,0.00023892038877405773)(285,0.00023642930290611972)(286,0.0002361826435145979)(287,0.00023549129053604496)(288,0.00022417578356615436)(289,0.00022168476064047998)(290,0.00022020775774177705)(291,0.00021675714397939295)(292,0.0002146493987532685)(293,0.00021109596928250477)(294,0.00021094443727559574)(295,0.00021038181368158957)(296,0.00020841623983937154)(297,0.00020739116085792108)(298,0.00020360461974568614)(299,0.00019705216764861154)(300,0.00019656084392061316)(301,0.00019505737208143505)(302,0.00019141564455555423)(303,0.00018932565870635686)(304,0.00018925376158289163)(305,0.0001857323043678376)(306,0.00018469295869550018)(307,0.00017850265475087549)(308,0.0001732978100330085)(309,0.0001727818991114642)(310,0.0001682012170434388)(311,0.0001648914421293789)(312,0.00016478004253359145)(313,0.00016372807721468964)(314,0.00016232563952914495)(315,0.00016129678727363036)(316,0.0001578995292938938)(317,0.0001488507324887351)(318,0.00014681404946357467)(319,0.00014535915667445278)(320,0.0001447892915219371)(321,0.00014304522632219488)(322,0.00014035417918267776)(323,0.00013476671696181365)(324,0.00013454617932899855)(325,0.00013254209436761377)(326,0.00013111859147209098)(327,0.0001296883113780171)(328,0.00012782655269261354)(329,0.0001261992253578411)(330,0.0001216365202998811)(331,0.00012058861919977806)(332,0.00012003382296719939)(333,0.0001117164449159934)(334,0.00010492051197819894)(335,0.00010343285328240419)(336,0.00010161550643116505)(337,0.00010039526432779997)(338,9.964604944412462e-5)(339,9.931087805244946e-5)(340,9.929493605275099e-5)(341,9.863984986621937e-5)(342,9.694171872656766e-5)(343,8.499761723589843e-5)(344,8.286088450276429e-5)(345,8.209480713337843e-5)(346,7.993660216315693e-5)(347,7.823508677475337e-5)(348,7.737238347165974e-5)(349,7.677676155782972e-5)(350,7.672402295292438e-5)(351,7.285161908764286e-5)(352,7.106236774017032e-5)(353,6.61041390966984e-5)(354,6.551076646155856e-5)(355,5.9483923372782545e-5)(356,5.703721424280395e-5)(357,5.373425812794411e-5)(358,5.2167486271463475e-5)(359,5.208040485190361e-5)(360,5.176440004113346e-5)(361,4.6382733935456664e-5)(362,4.396241909003537e-5)(363,4.2635955955582124e-5)(364,3.872557080599984e-5)(365,3.423814021304676e-5)(366,2.7754034380932656e-5)(367,2.4496699349596548e-5)(368,2.4407805556469466e-5)(369,2.4146495207093043e-5)(370,2.3289705516215602e-5)(371,2.3232954277835873e-5)(372,1.609889296454502e-5)(373,1.5017538754232956e-5)(374,1.4178649157686324e-5)(375,1.3940069187369928e-5)(376,1.0193433858149713e-5)(377,8.612199251943898e-6)(378,6.8619007468714295e-6)(379,6.746518135282285e-6)(380,5.747449285778801e-6)(381,5.589889637903353e-6)(382,5.081851042558993e-6)(383,4.970845260497799e-6)(384,3.5476319395794227e-6)(385,3.160865702613074e-6)(386,2.9738278895794337e-6)(387,2.3398258837127465e-6)(388,1.5984309055861142e-6)(389,6.50390510965912e-7)(390,1.8333799559485039e-7)(391,1.2445365087631033e-7)(392,2.3136790915138982e-8)(393,2.1084430390867572e-8)(394,1.8537095560257538e-8)(395,0.0)(396,0.0)(397,0.0)(398,0.0)(399,0.0)(400,0.0)(401,0.0)(402,0.0)(403,0.0)(404,0.0)(405,0.0)(406,0.0)};
			
			\addplot[darkgray,sharp plot,update limits=false, densely dashed, thin] coordinates {(0.0,0.01) (22.0,0.01)};
			\addplot[darkgray,sharp plot,update limits=false, densely dashed, thin] coordinates {(0.0,0.09) (22.0,0.09)};
			\addplot[darkgray,sharp plot,update limits=false, densely dashed, thin] coordinates {(0.0,0.01) (0.0,0.09)};
			\addplot[darkgray,sharp plot,update limits=false, densely dashed, thin] coordinates {(22.0,0.01)(22.0,0.09)};
			
			\addplot[darkgray,sharp plot,update limits=false,  densely dashed, thin] coordinates {(22.0,0.09)(169.0,0.615)};
			\addplot[darkgray,sharp plot,update limits=false, densely dashed, thin] coordinates {(22.0,0.01)(169,0.00215)};
		\end{axis}
		\node[anchor=south west] at (pt) {
			\begin{tikzpicture}
				\begin{axis}[axis background/.style={fill=white!10}, footnotesize, width=10.75cm, height=3.75cm, xmin=1,  xmax=21, xtick = {1, 5, 10, 15,20 }, ytick = {0.1, 0.01}, yticklabels = {$10^{-1}$, $10^{-2}$}, ymode=log,log origin=infty, ymin = 0.0055, xtick = {1, 5, 10, 15, 21},xticklabels={$1$, $5$, $10$, $15$, $21$}, enlargelimits=0.05]
					\addplot+[white,mark size=0.1pt] coordinates {	(1.5,0.65)};
					\addplot+[ycomb, orange, mark options={orange}, mark size=1.3pt, mark=*] plot coordinates {(1,0.05345307855798923)(2,0.032234093136252244)(3,0.03212794264021359)(4,0.03185068243659376)(5,0.030982733026525423)(6,0.029979908469847952)(7,0.027222070835683305)(8,0.025130900179535915)(9,0.024344799154963903)(10,0.022114861936327758)(11,0.021311219244518162)(12,0.020683005648822545)(13,0.02024712982403898)(14,0.019071277670052844)(15,0.018333676739135413)(16,0.01786955598344579)(17,0.01786369569332421)(18,0.016785306294709444)(19,0.015158598430717702)(20,0.0131473157492675)(21,0.012919309213475061)};
					\node at (axis cs:1,0.8) [anchor=north] {\rotatebox{90}{\scriptsize{$\{6,27\}$}}};
					\node at (axis cs:2,0.35) [anchor=north] {\rotatebox{90}{\scriptsize{$\{6,9\}$}}};
					\node at (axis cs:3,0.35) [anchor=north] {\rotatebox{90}{\scriptsize{$\{1,4\}$}}};
					\node at (axis cs:4,0.6) [anchor=north] {\rotatebox{90}{\scriptsize{$\{23,28\}$}}};
					\node at (axis cs:5,0.65) [anchor=north] {\rotatebox{90}{\scriptsize{$\{27,28\}$}}};
					\node at (axis cs:6,0.45) [anchor=north] {\rotatebox{90}{\scriptsize{$\{6,10\}$}}};
					\node at (axis cs:7,0.4) [anchor=north] {\rotatebox{90}{\scriptsize{$\{4,26\}$}}};
					\node at (axis cs:8,0.4) [anchor=north] {\rotatebox{90}{\scriptsize{$\{9,26\}$}}};
					\node at (axis cs:9,0.5) [anchor=north] {\rotatebox{90}{\scriptsize{$\{22,23\}$}}};
					\node at (axis cs:10,0.45) [anchor=north] {\rotatebox{90}{\scriptsize{$\{26,27\}$}}};
					\node at (axis cs:11,0.3) [anchor=north] {\rotatebox{90}{\scriptsize{$\{1,26\}$}}};
					\node at (axis cs:12,0.4) [anchor=north] {\rotatebox{90}{\scriptsize{$\{23,25\}$}}};
					\node at (axis cs:13,0.3) [anchor=north] {\rotatebox{90}{\scriptsize{$\{2,15\}$}}};
					\node at (axis cs:14,0.275) [anchor=north] {\rotatebox{90}{\scriptsize{$\{2,19\}$}}};
					\node at (axis cs:15,0.35) [anchor=north] {\rotatebox{90}{\scriptsize{$\{25,26\}$}}};
					\node at (axis cs:16,0.25) [anchor=north] {\rotatebox{90}{\scriptsize{$\{1,27\}$}}};
					\node at (axis cs:17,0.35) [anchor=north] {\rotatebox{90}{\scriptsize{$\{10,27\}$}}};
					\node at (axis cs:18,0.3) [anchor=north] {\rotatebox{90}{\scriptsize{$\{25,28\}$}}};
					\node at (axis cs:19,0.15) [anchor=north] {\rotatebox{90}{\scriptsize{$\{1,6\}$}}};
					\node at (axis cs:20,0.2) [anchor=north] {\rotatebox{90}{\scriptsize{$\{5,12\}$}}};
					\node at (axis cs:21,0.25) [anchor=north] {\rotatebox{90}{\scriptsize{$\{22,28\}$}}};
				\end{axis}
			\end{tikzpicture}
		};
	\end{tikzpicture}
	\caption{Computed global sensitivity indices $\varrho(\u, S(\X,\I_{\b N}(U_2))f^\text{cos})$ with $\bfh=\bfh^{\cos}$ using $N_1=4$, $N_2=6$, $\lambda=2^{-9}$ and $M=5\cdot10^4$, sorted according to the importance, for the HIGGS data set.}
	\label{fig:higgs_gsi}
\end{figure}

Next, we consider $S(\X,\I_{\b N}(U_2))f^\text{cos}(\x)$ with $N_1=4$ and $N_2=6$ in order to take a closer look into the resulting GSIs of the classifying function.
In Figure \ref{fig:higgs_gsi} we visualize the GSIs of all $406$ ANOVA terms in sorted order, computed by using the fixed regularization parameter $\lambda=2^{-9}$.
In this particular case, depending on the chosen $5\cdot10^{4}$ training data points, this leads to a CA of $68.01\%$ and an AUC of $0.7379$. 
The classification results are comparable to~\cite{NeStWa2023,WaPeSt2023}.
In the paper~\cite{Baldi2014} or~\cite{enouen2023} the authors achieved better AUC results.
But it is interesting, that if we use only the active set $U^{0.01}=U^\star$, see the highlighted GSIs in orange in Figure \ref{fig:higgs_gsi}, we need to compute a much smaller number of coefficients $\bfh^{\cos}\in\R^{551}$ and still achieve a CA of $68.05\%$ and AUC of $0.7374$.

\subsubsection{Real-world data~\label{sec:real}}
In this section we consider real-world data sets, for which we summarize the included features and their corresponding value ranges in Table~\ref{table:attributes}.
The first one is the well-known Wisconsin Breast Cancer (WBC) data set,~\cite{Mangasarian1995}, collected by Wolberg at the University of Wisconsin.
This database contains some noise and residual variation in its $683$ data points. 
There are nine integer inputs, each with a value between $1$ and $10$.
The two output classes, benign and malignant, are represented by $65.5\%$ and $34.5\%$, respectively, so they are not well balanced. 
The second data set, the Pima Indians Diabetes (PID) data set,  was first mentioned in~\cite{Smith1988} and is originally from the National Institute of Diabetes and Digestive and Kidney Diseases.
Pima Indians are a Native American group that lives in Mexico and Arizona, USA. This group was considered to have a high incidence rate of diabetes.
Research on them was therefore important and representative of global health.
The PID data set, consisting of data from Pima Indian women aged 21 years and older, is a popular benchmark data set in the literature. The two non-balanced output classes account for $65.1\%$ and $34.9\%$, respectively, and represent the proportions of the $768$ test persons who were tested negative or rather positive for diabetes. 

\begin{table}[!ht]
	\centering
	\begin{tabular}[t]{clc}
		\toprule
		$i$&Attribute&Domain\\
		\midrule
		$1$&Clump Thickness&$1-10$\\
		$2$&Uniformity of Cell Size&$1-10$\\
		$3$&Uniformity of Cell Shape&$1-10$\\
		$4$&Marginal Adhesion&$1-10$\\
		$5$&Single Epithelial Cell Size&$1-10$\\
		$6$&Bare Nuclei&$1-10$\\
		$7$&Bland Chromatin&$1-10$\\
		$8$&Normal Nucleoli&$1-10$\\
		$9$&Mitoses&$1-10$\\
		\bottomrule
	\end{tabular}
	\hspace{20pt}
	\begin{tabular}[t]{cll}
		\toprule
		$i$&Attribute&Domain\\
		\midrule
		$1$&Number of times pregnant&$0-17$\\
		$2$&Plasma glucose concentration&$0-199$\\
		$3$&Diastolic blood pressure $(\mathrm{mmHg})$&$0-122$\\
		$4$&Triceps skin-fold thickness $(\mathrm{mm})$&$0-99$\\
		$5$&$2-\mathrm{h}$ serum insulin $(\mathrm{\mu U}/\mathrm{mL})$&$0-846$\\
		$6$&Body mass index $(\mathrm{kg}/\mathrm{m}^2)$&$0-67.1$\\
		$7$&Diabetes pedigree function&$0.1-2.4$\\
		$8$&Age&$21-81$\\
		$ $& &$ $\\
		\bottomrule
	\end{tabular}
	\vspace{1.5em}
	\caption{The nine attributes of the WBC data set and the eight attributes of PID data set.}
	\label{table:attributes}
\end{table}

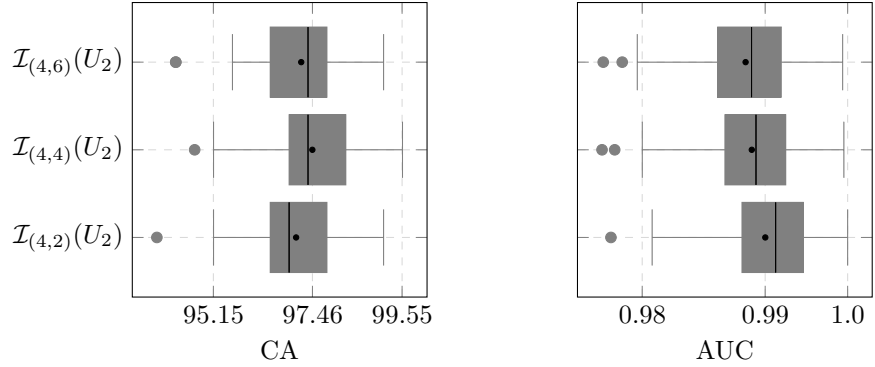
\begin{figure}[!ht]
	\centering
	\begin{tikzpicture}
		\begin{axis}
			[width=5.5cm, height=5.5cm,
			boxplot/draw direction = x, boxplot/every median/.style={black,solid},yticklabel style={anchor=west,xshift=-4.85em}, ytick = {1, 2, 3}, 
			grid = major,grid style={dashed, gray!30}, 
			xtick={95.15, 97.46, 99.55}, xticklabels={$95.15$,$97.46$, $99.55$}, 
			yticklabels = {$\mathcal{I}_{(4,2)}(U_2)$, $\mathcal{I}_{(4,4)}(U_2)$,$\mathcal{I}_{(4,6)}(U_2)$},
			every axis plot/.append style = {fill, fill opacity = 0.1},cycle list={{black}}, 
			scatter/classes={a={mark=*, mark size=1, fill=black, fill opacity = 0.75}}, 
			xlabel = $\mathrm{CA}$]
			\addplot + [mark = *, boxplot, gray]table [row sep =\\, y index = 0] {97.3568281938326\\96.47577092511013\\96.0352422907489\\98.23788546255507\\96.91629955947137\\96.91629955947137\\98.23788546255507\\98.6784140969163\\96.91629955947137\\96.91629955947137\\96.47577092511013\\96.47577092511013\\98.6784140969163\\96.91629955947137\\96.91629955947137\\98.6784140969163\\98.23788546255507\\97.79735682819384\\97.3568281938326\\96.91629955947137\\98.23788546255507\\98.23788546255507\\97.79735682819384\\98.6784140969163\\97.3568281938326\\96.0352422907489\\97.79735682819384\\96.47577092511013\\96.47577092511013\\96.47577092511013\\97.3568281938326\\95.59471365638767\\97.79735682819384\\97.3568281938326\\98.23788546255507\\95.15418502202643\\99.11894273127754\\96.47577092511013\\97.3568281938326\\97.79735682819384\\96.91629955947137\\97.79735682819384\\95.59471365638767\\97.79735682819384\\96.0352422907489\\95.15418502202643\\96.91629955947137\\97.79735682819384\\96.47577092511013\\98.6784140969163\\93.83259911894272\\96.0352422907489\\96.47577092511013\\98.23788546255507\\96.0352422907489\\96.47577092511013\\96.0352422907489\\96.47577092511013\\97.3568281938326\\96.91629955947137\\98.23788546255507\\97.3568281938326\\96.0352422907489\\97.3568281938326\\98.23788546255507\\96.0352422907489\\98.6784140969163\\95.59471365638767\\95.59471365638767\\96.91629955947137\\96.91629955947137\\98.6784140969163\\95.59471365638767\\96.0352422907489\\96.91629955947137\\95.59471365638767\\96.0352422907489\\96.47577092511013\\98.6784140969163\\97.3568281938326\\97.79735682819384\\98.23788546255507\\97.3568281938326\\96.91629955947137\\98.6784140969163\\96.47577092511013\\97.79735682819384\\96.91629955947137\\96.0352422907489\\95.15418502202643\\97.3568281938326\\96.91629955947137\\97.3568281938326\\97.79735682819384\\96.47577092511013\\95.59471365638767\\98.23788546255507\\97.79735682819384\\97.3568281938326\\97.3568281938326\\};
			\addplot + [mark = *, boxplot, gray]table [row sep =\\, y index = 0] {96.91629955947137\\96.91629955947137\\96.47577092511013\\97.79735682819384\\97.79735682819384\\97.79735682819384\\98.23788546255507\\98.23788546255507\\98.23788546255507\\96.91629955947137\\96.47577092511013\\96.47577092511013\\99.11894273127754\\97.3568281938326\\96.91629955947137\\98.6784140969163\\98.23788546255507\\97.79735682819384\\97.3568281938326\\97.3568281938326\\98.6784140969163\\98.6784140969163\\98.23788546255507\\98.6784140969163\\97.3568281938326\\96.47577092511013\\98.6784140969163\\97.3568281938326\\96.91629955947137\\97.3568281938326\\98.23788546255507\\96.91629955947137\\98.6784140969163\\96.91629955947137\\97.3568281938326\\95.59471365638767\\98.23788546255507\\96.91629955947137\\97.79735682819384\\97.3568281938326\\96.91629955947137\\98.23788546255507\\96.0352422907489\\98.6784140969163\\96.47577092511013\\95.15418502202643\\97.79735682819384\\97.3568281938326\\96.0352422907489\\99.11894273127754\\95.59471365638767\\96.91629955947137\\96.91629955947137\\97.79735682819384\\96.91629955947137\\96.91629955947137\\97.3568281938326\\96.91629955947137\\97.79735682819384\\97.3568281938326\\98.6784140969163\\97.3568281938326\\96.91629955947137\\98.23788546255507\\98.23788546255507\\97.3568281938326\\98.6784140969163\\97.3568281938326\\96.47577092511013\\97.3568281938326\\97.79735682819384\\98.6784140969163\\96.91629955947137\\96.91629955947137\\96.91629955947137\\96.47577092511013\\96.91629955947137\\97.3568281938326\\98.6784140969163\\98.23788546255507\\97.3568281938326\\99.55947136563876\\98.23788546255507\\97.3568281938326\\97.3568281938326\\96.91629955947137\\98.6784140969163\\97.3568281938326\\97.79735682819384\\94.7136563876652\\96.91629955947137\\97.3568281938326\\97.79735682819384\\97.3568281938326\\96.47577092511013\\97.3568281938326\\97.79735682819384\\98.6784140969163\\97.3568281938326\\96.0352422907489\\};
			\addplot + [mark = *, boxplot, gray]table [row sep =\\, y index = 0] {96.47577092511013\\97.3568281938326\\96.91629955947137\\98.23788546255507\\97.79735682819384\\97.79735682819384\\97.79735682819384\\97.3568281938326\\97.79735682819384\\96.91629955947137\\96.47577092511013\\96.47577092511013\\99.11894273127754\\97.3568281938326\\96.91629955947137\\98.6784140969163\\97.79735682819384\\96.0352422907489\\97.3568281938326\\97.3568281938326\\98.23788546255507\\97.79735682819384\\98.23788546255507\\98.6784140969163\\96.91629955947137\\96.0352422907489\\98.23788546255507\\96.91629955947137\\96.91629955947137\\97.3568281938326\\98.23788546255507\\96.0352422907489\\98.23788546255507\\96.47577092511013\\96.0352422907489\\95.59471365638767\\98.23788546255507\\96.47577092511013\\97.79735682819384\\97.79735682819384\\96.91629955947137\\98.6784140969163\\95.59471365638767\\98.6784140969163\\96.47577092511013\\94.27312775330397\\97.3568281938326\\97.79735682819384\\95.59471365638767\\99.11894273127754\\96.0352422907489\\96.91629955947137\\97.3568281938326\\96.47577092511013\\96.91629955947137\\96.47577092511013\\96.91629955947137\\96.47577092511013\\97.3568281938326\\98.23788546255507\\98.6784140969163\\97.3568281938326\\96.47577092511013\\96.91629955947137\\98.6784140969163\\96.47577092511013\\98.23788546255507\\97.3568281938326\\96.0352422907489\\96.91629955947137\\98.23788546255507\\98.6784140969163\\95.59471365638767\\96.91629955947137\\95.59471365638767\\96.0352422907489\\96.47577092511013\\96.91629955947137\\97.3568281938326\\97.79735682819384\\97.3568281938326\\99.11894273127754\\97.3568281938326\\97.3568281938326\\96.47577092511013\\96.91629955947137\\98.23788546255507\\96.91629955947137\\97.79735682819384\\94.27312775330397\\96.91629955947137\\97.3568281938326\\97.79735682819384\\97.3568281938326\\96.0352422907489\\96.91629955947137\\97.79735682819384\\97.79735682819384\\96.91629955947137\\96.47577092511013\\};
			\addplot [scatter,only marks,scatter src=explicit symbolic] 
			table [
			y expr={1 + floor(\coordindex/1) + 1/3*mod(\coordindex,1)},
			x=x,
			meta=meta
			]
			{
				x meta		
				97.07929515418503 a
				97.4581497797357 a
				97.19823788546255 a
				
			};
		\end{axis}
	\end{tikzpicture}
	\begin{tikzpicture}
		\begin{axis}
			[width=5.5cm, height=5.5cm,
			boxplot/draw direction = x, boxplot/every median/.style={black,solid}, 
			yticklabel style={anchor=west,xshift=-4.85em}, ytick = {1, 2, 3}, 
			grid = major,grid style={dashed, gray!30}, 
			xtick={0.98605, 0.9944, 1.0}, xticklabels={$0.98$,$0.99$, $1.0$}, 
			yticklabels = {$ $, $ $, $ $},
			every axis plot/.append style = {fill, fill opacity = 0.1},cycle list={{black}}, 
			scatter/classes={a={mark=*, mark size=1, fill=black, fill opacity = 0.75}}, 
			xlabel = $\mathrm{AUC}$]
			]
			\addplot + [mark = *, boxplot, gray]table [row sep =\\, y index = 0] {0.9934091729866378\\0.9957302195887069\\0.9939444911690496\\0.9985087719298246\\0.9953400915631131\\0.9952631578947368\\0.9892006802721088\\0.9964696223316913\\0.9971613949716139\\0.9946633825944171\\0.9917218543046358\\0.9949829931972789\\0.9996701302985321\\0.9930721861101608\\0.994700582935877\\1.0\\0.9941176470588236\\0.997224558452481\\0.9899328859060402\\0.9953492305090479\\0.9975806451612903\\0.9976291693917593\\0.9956265769554248\\0.998145204027557\\0.9928453494771602\\0.9920831271647699\\0.9963619402985074\\0.9928070175438597\\0.9896991497710922\\0.9976689976689976\\0.9973326449836517\\0.9935575635876841\\0.9995837495837496\\0.9907706093189964\\0.9930846223839854\\0.9993859649122807\\0.9908602150537634\\0.9935897435897436\\0.9982338099243061\\0.9893859649122807\\0.9969209716045159\\0.9885204081632653\\0.9978014544224589\\0.9936379928315412\\0.9907592407592407\\0.9945261717413616\\0.9951248717071501\\0.9867227445159162\\0.9936843977939869\\0.9868399935804847\\0.9963054187192119\\0.9896815513253869\\0.9950955521731777\\0.9923238526179703\\0.9939759036144579\\0.997080291970803\\0.99580910023948\\0.9961750578188935\\0.9957116938809171\\0.9970745138530374\\0.9941291585127201\\0.9920175438596491\\0.9964022757697456\\0.9953145917001339\\0.9966442953020134\\0.9971931862175765\\0.9963222716387273\\0.9940563426152093\\0.9925988225399496\\0.9971249788601386\\0.9961734693877551\\0.9938271604938271\\0.9934548467274233\\0.9883449883449883\\0.9871794871794872\\0.9972630858706808\\0.983927091963546\\0.9994868286007527\\0.9973359973359973\\0.9953201970443349\\0.9986863711001642\\0.9927675148135239\\0.9933757286698464\\0.9951815522285321\\0.9942269076305221\\0.9961753731343284\\0.9971938775510204\\0.9954954954954955\\0.9888444888444888\\0.9934994582881906\\0.9960526315789474\\0.9911374978489073\\0.9942833471416735\\0.9966233766233766\\0.9938528138528139\\0.9975806451612903\\0.9970568999345978\\0.9952631578947368\\0.9924319727891157\\};
			
			\addplot + [mark = *, boxplot, gray]table [row sep =\\, y index = 0] {0.9905200433369448\\0.9950331125827815\\0.9883936080740118\\0.9979824561403509\\0.9950130804447351\\0.9935964912280701\\0.9912414965986395\\0.9958128078817734\\0.9960259529602595\\0.9940065681444992\\0.9921575461833392\\0.9916666666666667\\0.9995051954477981\\0.9864009579199453\\0.9938173467585232\\0.9997463216641299\\0.9945318972659486\\0.9939444911690496\\0.9874376183101016\\0.9923896499238964\\0.9982078853046595\\0.9982831916285154\\0.9957947855340622\\0.9967320261437909\\0.9923867180333884\\0.9905987135081643\\0.996268656716418\\0.9935087719298246\\0.9889633747547416\\0.9975857475857476\\0.996816382722423\\0.9918005354752343\\0.9995004995004995\\0.9875448028673836\\0.9903548680618744\\0.9833224329627207\\0.9985087719298246\\0.9884408602150537\\0.9937703141928494\\0.9973927670311186\\0.9885087719298246\\0.9966643859048923\\0.9860544217687075\\0.9983933705394893\\0.992741935483871\\0.9915084915084915\\0.9938419432090319\\0.9950393431406089\\0.9867227445159162\\0.9953744885251734\\0.9871609693468143\\0.9942528735632183\\0.9896815513253869\\0.9930661254862168\\0.9930510665804784\\0.9907128514056225\\0.9965125709651257\\0.9927300718439959\\0.9951965842376801\\0.9940623453735774\\0.9977628635346756\\0.9932396370752535\\0.9903508771929824\\0.9957329317269076\\0.992218875502008\\0.9949234210979178\\0.9939024390243902\\0.9955525145398563\\0.9927536231884058\\0.9937762825904121\\0.9975477760865888\\0.9948979591836735\\0.9903602232369355\\0.9929577464788732\\0.9883449883449883\\0.9904491481672689\\0.996578857338351\\0.987903893951947\\0.9995723571672939\\0.9968364968364969\\0.9956486042692939\\0.9977011494252873\\0.9936388985709307\\0.9942589648472001\\0.9935467217346412\\0.9946452476572959\\0.994589552238806\\0.9953231292517006\\0.9943472884649355\\0.9841824841824842\\0.9918743228602384\\0.9935964912280701\\0.9920839786611598\\0.991880695940348\\0.9956709956709957\\0.992900432900433\\0.9972222222222222\\0.9980379332897319\\0.9936842105263158\\0.9922619047619048\\};
			\addplot + [mark = *, boxplot, gray]table [row sep =\\, y index = 0] {0.9899783315276273\\0.9948588358313001\\0.9846930193439866\\0.9982456140350877\\0.993378024852845\\0.9937719298245614\\0.9900510204081633\\0.994088669950739\\0.9961881589618816\\0.9954844006568144\\0.9905019170442663\\0.990391156462585\\0.9995876628731651\\0.9869996578857338\\0.9928457869634341\\0.9996617622188398\\0.9932062966031483\\0.9928511354079058\\0.9878678368611254\\0.993573482157957\\0.9975806451612903\\0.9967298888162197\\0.9952060555088309\\0.9967320261437909\\0.992294991744634\\0.9911759854857332\\0.9959888059701493\\0.9930701754385964\\0.9881458469587966\\0.9981684981684982\\0.9967303390122182\\0.9917168674698795\\0.9985847485847485\\0.985931899641577\\0.9881710646041856\\0.9834041857423153\\0.9973684210526316\\0.988978494623656\\0.9949440231130372\\0.9982338099243061\\0.9878070175438597\\0.9962367430721861\\0.9857142857142858\\0.9982242516489092\\0.9915770609318997\\0.9915917415917416\\0.9937564146424905\\0.9954669859733151\\0.9863104073890813\\0.9940402063689735\\0.9859573102230782\\0.9945812807881773\\0.989948407756627\\0.9931506849315068\\0.9931318681318682\\0.9928045515394913\\0.9964314679643147\\0.9936708860759493\\0.9950186799501868\\0.9940623453735774\\0.9969024264326277\\0.9911937377690803\\0.9912280701754386\\0.9960676037483266\\0.9934738955823293\\0.9946652899673034\\0.9936120789779327\\0.9952959288402327\\0.992835043152581\\0.9912531539108494\\0.9968713005242685\\0.9954931972789116\\0.9917131743615761\\0.9933719966859983\\0.9860972360972361\\0.9877817931509206\\0.9964933287718097\\0.9865782932891466\\0.9996578857338351\\0.997086247086247\\0.9935960591133005\\0.9958128078817734\\0.9916347159288951\\0.9947889065536124\\0.9932885906040269\\0.9903781793842035\\0.9943097014925373\\0.9944727891156463\\0.9940823176117294\\0.9857642357642358\\0.9914228963524738\\0.9931578947368421\\0.9911374978489073\\0.9930405965202983\\0.9953246753246753\\0.9925541125541125\\0.996505376344086\\0.9970568999345978\\0.9939473684210526\\0.9902210884353742\\};
			
			\addplot [scatter,only marks,scatter src=explicit symbolic] 
			table [y expr={1 + floor(\coordindex/1) + 1/3*mod(\coordindex,1)}, x=x, meta=meta]
			{
				x meta		
				0.9944082282503204 a
				0.9934904046242309 a
				0.9930703246210681 a
			};
		\end{axis}
	\end{tikzpicture}
	\caption{Achieved highest (among all considered regularization parameters $\lambda$) CA and AUC for the WBC data set, averaged over $100$ runs for different index sets.
		We applied the cosine basis approach with $\ell_1$-norm regularization. For each run, we randomly select $456$ data points for training the model and the remaining $227$ data points for testing.}
	\label{fig:wbcd}
\end{figure}

\begin{figure}[!ht]
	\centering
	\begin{tikzpicture}
		\begin{axis}
			[width=5.5cm, height=5.5cm,
			boxplot/draw direction = x, boxplot/every median/.style={black,solid},yticklabel style={anchor=west,xshift=-4.85em}, ytick = {1, 2, 3}, 
			grid = major,grid style={dashed, gray!30}, 
			xtick={73.04, 78.28, 83.91}, xticklabels={$73.04$,$78.28$, $83.91$}, 
			yticklabels = {$\mathcal{I}_{(4,2)}(U_2)$, $\mathcal{I}_{(4,4)}(U_2)$,$\mathcal{I}_{(4,6)}(U_2)$},
			every axis plot/.append style = {fill, fill opacity = 0.1},cycle list={{black}}, 
			scatter/classes={a={mark=*, mark size=1, fill=black, fill opacity = 0.75}}, 
			xlabel = $\mathrm{CA}$]
			\addplot + [mark = *, boxplot, gray]table [row sep =\\, y index = 0] {76.52173913043478\\79.13043478260869\\77.82608695652173\\78.26086956521739\\80.43478260869566\\74.78260869565217\\79.13043478260869\\79.13043478260869\\77.82608695652173\\73.91304347826086\\80.0\\76.95652173913044\\77.39130434782608\\77.82608695652173\\80.43478260869566\\79.13043478260869\\79.13043478260869\\80.43478260869566\\79.56521739130434\\80.0\\80.43478260869566\\80.8695652173913\\83.47826086956522\\76.95652173913044\\76.95652173913044\\75.65217391304347\\80.0\\73.04347826086956\\82.17391304347827\\76.95652173913044\\76.52173913043478\\76.52173913043478\\83.91304347826087\\77.82608695652173\\75.65217391304347\\74.78260869565217\\76.95652173913044\\79.13043478260869\\80.43478260869566\\75.21739130434783\\78.26086956521739\\75.65217391304347\\73.91304347826086\\76.95652173913044\\77.82608695652173\\77.39130434782608\\74.78260869565217\\80.0\\80.43478260869566\\80.0\\76.52173913043478\\76.52173913043478\\80.8695652173913\\76.52173913043478\\81.73913043478261\\78.69565217391305\\73.91304347826086\\79.13043478260869\\78.26086956521739\\79.56521739130434\\74.34782608695653\\77.39130434782608\\76.95652173913044\\76.52173913043478\\79.56521739130434\\77.82608695652173\\76.08695652173914\\73.91304347826086\\78.69565217391305\\79.13043478260869\\82.17391304347827\\76.95652173913044\\80.0\\79.56521739130434\\76.52173913043478\\80.43478260869566\\80.8695652173913\\80.0\\76.52173913043478\\81.30434782608695\\77.39130434782608\\78.26086956521739\\81.30434782608695\\79.13043478260869\\76.52173913043478\\78.69565217391305\\78.69565217391305\\74.34782608695653\\83.47826086956522\\76.95652173913044\\80.43478260869566\\80.8695652173913\\79.13043478260869\\75.65217391304347\\82.17391304347827\\78.26086956521739\\80.0\\78.26086956521739\\78.26086956521739\\77.39130434782608\\};
			\addplot + [mark = *, boxplot, gray]table [row sep =\\, y index = 0] {75.21739130434783\\78.69565217391305\\77.39130434782608\\77.39130434782608\\77.39130434782608\\73.47826086956522\\79.13043478260869\\79.56521739130434\\78.26086956521739\\74.78260869565217\\78.26086956521739\\76.95652173913044\\77.82608695652173\\78.69565217391305\\79.13043478260869\\77.82608695652173\\77.82608695652173\\80.8695652173913\\78.26086956521739\\78.26086956521739\\80.8695652173913\\80.43478260869566\\82.6086956521739\\79.13043478260869\\76.52173913043478\\75.21739130434783\\79.13043478260869\\72.60869565217392\\80.8695652173913\\79.13043478260869\\74.78260869565217\\76.08695652173914\\83.04347826086956\\77.82608695652173\\76.08695652173914\\73.04347826086956\\74.34782608695653\\77.39130434782608\\80.0\\74.78260869565217\\76.95652173913044\\76.08695652173914\\73.91304347826086\\76.95652173913044\\77.39130434782608\\76.95652173913044\\76.95652173913044\\79.56521739130434\\80.43478260869566\\77.82608695652173\\77.39130434782608\\76.95652173913044\\79.56521739130434\\78.26086956521739\\80.0\\77.39130434782608\\74.78260869565217\\78.26086956521739\\78.26086956521739\\79.13043478260869\\75.65217391304347\\73.47826086956522\\76.95652173913044\\76.95652173913044\\77.82608695652173\\76.52173913043478\\74.34782608695653\\74.34782608695653\\76.52173913043478\\77.39130434782608\\83.04347826086956\\79.13043478260869\\76.52173913043478\\80.0\\76.95652173913044\\80.43478260869566\\81.30434782608695\\78.69565217391305\\75.65217391304347\\81.30434782608695\\77.82608695652173\\75.65217391304347\\80.0\\78.69565217391305\\76.95652173913044\\77.39130434782608\\78.69565217391305\\74.34782608695653\\82.17391304347827\\77.39130434782608\\76.95652173913044\\80.8695652173913\\78.26086956521739\\75.65217391304347\\82.17391304347827\\77.82608695652173\\80.0\\80.43478260869566\\76.95652173913044\\77.82608695652173\\};
			\addplot + [mark = *, boxplot, gray]table [row sep =\\, y index = 0] {75.21739130434783\\78.69565217391305\\76.52173913043478\\76.95652173913044\\76.52173913043478\\73.47826086956522\\79.56521739130434\\77.82608695652173\\77.82608695652173\\73.04347826086956\\77.82608695652173\\76.08695652173914\\78.26086956521739\\76.52173913043478\\79.13043478260869\\77.82608695652173\\78.26086956521739\\79.13043478260869\\79.56521739130434\\77.82608695652173\\80.0\\79.13043478260869\\82.6086956521739\\76.52173913043478\\75.21739130434783\\74.78260869565217\\78.69565217391305\\73.04347826086956\\77.82608695652173\\76.95652173913044\\76.08695652173914\\76.52173913043478\\81.73913043478261\\78.26086956521739\\74.34782608695653\\72.17391304347827\\74.34782608695653\\76.52173913043478\\80.0\\74.78260869565217\\75.21739130434783\\74.78260869565217\\73.91304347826086\\75.21739130434783\\76.52173913043478\\76.95652173913044\\76.52173913043478\\79.56521739130434\\77.39130434782608\\77.82608695652173\\75.65217391304347\\76.95652173913044\\80.0\\77.82608695652173\\78.26086956521739\\74.78260869565217\\73.47826086956522\\76.95652173913044\\77.82608695652173\\80.43478260869566\\76.52173913043478\\72.60869565217392\\76.95652173913044\\77.39130434782608\\76.08695652173914\\76.52173913043478\\75.21739130434783\\74.34782608695653\\76.52173913043478\\77.39130434782608\\80.8695652173913\\78.26086956521739\\76.52173913043478\\77.39130434782608\\76.95652173913044\\79.56521739130434\\80.43478260869566\\78.26086956521739\\75.65217391304347\\81.30434782608695\\78.26086956521739\\74.34782608695653\\80.0\\79.56521739130434\\77.82608695652173\\76.95652173913044\\77.39130434782608\\74.34782608695653\\82.17391304347827\\76.52173913043478\\77.39130434782608\\80.43478260869566\\77.82608695652173\\74.78260869565217\\82.17391304347827\\76.95652173913044\\79.13043478260869\\80.43478260869566\\74.78260869565217\\76.08695652173914\\};
			\addplot [scatter,only marks,scatter src=explicit symbolic] 
			table [
			y expr={1 + floor(\coordindex/1) + 1/3*mod(\coordindex,1)},
			x=x,
			meta=meta
			]
			{
				x meta		
				78.28260869565217 a
				77.81304347826087 a
				77.23913043478261 a
				
			};
		\end{axis}
	\end{tikzpicture}
	\begin{tikzpicture}
		\begin{axis}
			[width=5.5cm, height=5.5cm,
			boxplot/draw direction = x, boxplot/every median/.style={black,solid}, 
			yticklabel style={anchor=west,xshift=-4.85em}, ytick = {1, 2, 3}, 
			grid = major,grid style={dashed, gray!30}, 
			xtick={0.7863, 0.84, 0.8886}, xticklabels={$0.79$,$0.84$, $0.89$}, 
			yticklabels = {$ $, $ $, $ $},
			every axis plot/.append style = {fill, fill opacity = 0.1},cycle list={{black}}, 
			scatter/classes={a={mark=*, mark size=1, fill=black, fill opacity = 0.75}}, 
			xlabel = $\mathrm{AUC}$]
			]
			\addplot + [mark = *, boxplot, gray]table [row sep =\\, y index = 0] {0.8317243544462758\\0.8602395411605938\\0.827688651218063\\0.8580465587044535\\0.84375\\0.8291748206131768\\0.8426022176022177\\0.8135714285714286\\0.8253166799539374\\0.8230045816252712\\0.8470608339029392\\0.8179784946236559\\0.8544722944085841\\0.8233035714285715\\0.847284967667641\\0.8329988657185237\\0.8246045406114253\\0.8293485793485793\\0.8416942534589593\\0.8411607142857143\\0.8711281738068232\\0.8543353576248313\\0.874543756896698\\0.8252380952380952\\0.8388266315095584\\0.8001012145748988\\0.8462795698924731\\0.7935902636916836\\0.8869166666666667\\0.85093669250646\\0.7863924050632911\\0.8278701891715591\\0.8886526707414297\\0.8493903896724838\\0.8513825802852817\\0.815529298287919\\0.8052862706106554\\0.8467871997283762\\0.8328461334041988\\0.8130606086008886\\0.84175\\0.8115516807779362\\0.802547770700637\\0.8204464285714286\\0.8271531100478469\\0.8378571428571429\\0.8221624266144814\\0.8519574147036634\\0.8668333333333333\\0.866978116547824\\0.8345418589321029\\0.855713828425096\\0.8529633665856317\\0.8324055330634278\\0.8636177362511269\\0.8365877080665813\\0.8115516807779362\\0.8436583116774248\\0.8344155844155844\\0.8524225444957152\\0.7926195426195426\\0.8206704368494385\\0.8194408602150538\\0.8434077079107505\\0.8419918800231999\\0.8187803187803188\\0.824853228962818\\0.8075148319050758\\0.8637816572412097\\0.8668917289606944\\0.8726611226611226\\0.8398564905414221\\0.8366890380313199\\0.8764603529704201\\0.8280627159181504\\0.8593498005262711\\0.8546812664459723\\0.8513597307113119\\0.8095114345114345\\0.8389976958525346\\0.8389638695615727\\0.8071378340365682\\0.8814271588661833\\0.8229464285714285\\0.8169462365591398\\0.8675518400131137\\0.8436436611115958\\0.812\\0.856497475418549\\0.8404365904365905\\0.8463431786216596\\0.8430339985218034\\0.8585953520164047\\0.8258982396913431\\0.8549519586104952\\0.8553087586641462\\0.8545133027891648\\0.8571428571428571\\0.8349358974358975\\0.8390688259109311\\};
			
			\addplot + [mark = *, boxplot, gray]table [row sep =\\, y index = 0] {0.8138293697511969\\0.8447199730094467\\0.8258212375859435\\0.8388157894736842\\0.8359166666666666\\0.797863666014351\\0.8411295911295912\\0.8146428571428571\\0.8234564620426964\\0.8219596495458564\\0.8434723171565277\\0.8092903225806451\\0.8478497778522928\\0.819375\\0.8474621312782354\\0.816595410522642\\0.8278009999180395\\0.8239778239778239\\0.8437314319667261\\0.8166071428571429\\0.8794171538260187\\0.8435391363022942\\0.8696205755029285\\0.8376984126984127\\0.8303394858272907\\0.7789304993252362\\0.8394838709677419\\0.7922920892494929\\0.8785833333333334\\0.8510174418604651\\0.7697784810126582\\0.8272994129158513\\0.8792629993799274\\0.848354450553829\\0.8353653677583871\\0.8044369423679768\\0.7949291573452647\\0.8360071301247772\\0.819115953583134\\0.7941151814904853\\0.83775\\0.8116355101014335\\0.8104877410348137\\0.8036607142857143\\0.8157894736842105\\0.8195535714285714\\0.8226516634050881\\0.8465923379998324\\0.84575\\0.8467338742726006\\0.8393210283454186\\0.8467509603072984\\0.8439097996479168\\0.8275134952766532\\0.8473075977378903\\0.8225832266325224\\0.8028334311342108\\0.8350238913571968\\0.8394565960355435\\0.837425840474621\\0.7914067914067914\\0.7944430784361938\\0.8130752688172043\\0.8340770791075051\\0.8469632943905875\\0.8014553014553014\\0.8109915198956295\\0.7995220830586685\\0.8359970494221786\\0.8590949280604453\\0.8776853776853777\\0.8485812133072407\\0.8169690943740161\\0.8619603943988732\\0.8318717335459297\\0.8551056786350903\\0.843986079280197\\0.8359464965896005\\0.8102044352044352\\0.831989247311828\\0.8397183334730489\\0.7923699015471167\\0.8705504284772577\\0.8173214285714285\\0.8113548387096774\\0.8517334644701254\\0.8342203996160893\\0.7985833333333333\\0.852777039596067\\0.8248440748440748\\0.8340365682137834\\0.8425720620842572\\0.8442412850307587\\0.8156900570693674\\0.8458056171470806\\0.850346565847511\\0.8484044690941243\\0.8569642857142857\\0.816970310391363\\0.8404183535762483\\};
			\addplot + [mark = *, boxplot, gray]table [row sep =\\, y index = 0] {0.8128090416764775\\0.8480937921727395\\0.8238689415160003\\0.8334176788124157\\0.838\\0.7907697325505545\\0.8297817047817048\\0.8050892857142857\\0.812649481796439\\0.828791897757415\\0.8438140806561859\\0.8116989247311828\\0.8418140665604829\\0.8127678571428572\\0.8410842412968377\\0.825058895384347\\0.8290304073436604\\0.8199064449064449\\0.8513708513708513\\0.8198214285714286\\0.8764505715033593\\0.8501180836707153\\0.8663101604278075\\0.8215873015873015\\0.8098220171390903\\0.7904858299595142\\0.8193548387096774\\0.7919675456389452\\0.8649166666666667\\0.8429425064599483\\0.7703059071729957\\0.8289302022178735\\0.8688103463548588\\0.8397481871065423\\0.8337716152681488\\0.7932642070573105\\0.794432015908526\\0.8302351243527714\\0.8223048985738329\\0.7853131025232626\\0.8315833333333333\\0.8066895800150893\\0.8015007416455807\\0.7982142857142858\\0.7952836637047164\\0.8083035714285715\\0.8209393346379648\\0.8422332131779696\\0.8351666666666666\\0.8358331284320958\\0.8305866842452209\\0.8500320102432779\\0.8225333221560902\\0.8253205128205128\\0.8418982050651586\\0.8187419974391805\\0.7893369100511359\\0.8368681364741386\\0.8286056049213943\\0.8342946605141727\\0.7746015246015246\\0.7850995820014753\\0.8151397849462365\\0.8329411764705882\\0.8318004805700555\\0.7685377685377686\\0.807648401826484\\0.7987804878048781\\0.8359150889271372\\0.8449481552929828\\0.8494455994455995\\0.8582844096542727\\0.7999005717126523\\0.8412461678680918\\0.8274426432810701\\0.8568033273915627\\0.8447500212206095\\0.8359464965896005\\0.792966042966043\\0.8320852534562212\\0.8389638695615727\\0.7841068917018285\\0.8651944627554383\\0.8088392857142858\\0.813505376344086\\0.8546840422916154\\0.8288107494982986\\0.79175\\0.8544600938967136\\0.821985446985447\\0.8262130801687764\\0.8333333333333334\\0.8393711551606289\\0.8126356402218471\\0.8452512934220251\\0.8495589161940769\\0.8429386705248775\\0.8642857142857143\\0.8044028340080972\\0.8160425101214575\\};
			
			\addplot [scatter,only marks,scatter src=explicit symbolic] 
			table [y expr={1 + floor(\coordindex/1) + 1/3*mod(\coordindex,1)}, x=x, meta=meta]
			{
				x meta		
				0.838661609997867 a
				0.8305688441024951 a
				0.82505291796483 a
			};
		\end{axis}
	\end{tikzpicture}
	\caption{Achieved highest (among all considered regularization parameters $\lambda$) CA and AUC for the PID data set, averaged over $100$ runs for different index sets.
		We applied the cosine basis approach with $\ell_1$-norm regularization. For each run, we randomly select $538$ data points for training the model and the $230$ data points for testing.}
	\label{fig:pima}
\end{figure}
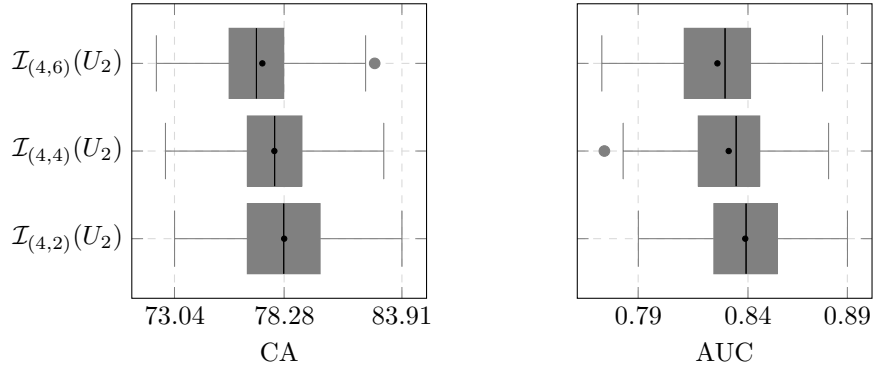

We use the same notation as in the previous section for the feature map, depending on different bandwidth parameters $\bf N$ and superposition dimension $d_s=2$. 
To make a general statement about our prediction results, we perform our computations over $100$ runs and visualize the results using box plots, in the same way as in the previous section.
In each run, we randomly select our training and test data sets. 
The model is trained using $456$ data points for the WBC data set, as it was done in~\cite{Sidey2019}. 
The remaining data points are used for testing.
We split the PID data set into $538$ data points for training and $230$ for testing, respectively $70/30$ split, as it was done in~\cite{Chang2023}. 
Depending on the regularization parameter $\lambda\in\{2^{-l}: l=1,\dots,10\}\cup\{0.0\}$ we compute basis coefficients $\bfh^{\cos}\in\R^{|\I_{\b N}(U_{d_s})|}$ by solving \eqref{b}. 
Using different bandwidth parameters $\bf N$ we have to calculate $\bfh^{\cos}\in\R^{64}$ for $N_2=2$, $\bfh^{\cos}\in\R^{352}$ for $N_2=4$ and $\bfh^{\cos}\in\R^{928}$ for $N_2=6$ in case if we consider WBC data. 
For PID data we have $\bfh^{\cos}\in\R^{53}$ for $N_2=2$, $\bfh^{\cos}\in\R^{277}$ for $N_2=4$ and $\bfh^{\cos}\in\R^{725}$ for $N_2=6$.
Using these coefficients we compute the corresponding CA and AUC for the classifying function $S(\X,\I_{\b N}(U_2))f^\text{cos})$ for both data sets. 
This procedure ensures that only the highest values of CA and AUC are selected in each run and are then considered in the box plots. 
The results for the WBC data set are visualized in Figure \ref{fig:wbcd} and for PID in \ref{fig:pima}. 
Note that the performance strongly depends on the applied regularization parameter $\lambda$.

We compare our results for the WBC data set with those from~\cite{Sidey2019}, where the authors used different machine learning methods to analyse this data set. 
As we see in Figure~\ref{fig:wbcd}, using different index sets gives us similar values for CA and AUC. Overall, we can say that we have achieved very good results. They are in fact as good as those from the literature, where the authors compared different machine learning algorithms, for example singe-layer artificial neural networks (ANNs), classical support vector machines with a radial basis function (RBF) kernel, or regularized general linear model regression (GLMs).

Figure~\ref{fig:pima} shows the the results for the PID data set. The results do not show a high degree of accuracy. 
However, they are comparable to~\cite{Chang2023}, where three different algorithms were used. 

Finally, we want to see whether our method can also provide meaningful information about interpretability. Therefore we compute the global sensitivity indices for fixed parameters, see Figure~\ref{fig:real_gsi}. 
Note that we now use more data points for training to get more accurate interpretability results. 
In case of WBC data set, we randomly select $614$ training data points and use the bandwidth parameters $N_1=N_2=4$.
We compute the basis coefficients and test the corresponding classification function on the remaining $69$ data points to figure out the best regularization parameter for the highest accuracy. 
For the PID data set, we will proceed in the same way, and use $691$ data points for training and $77$ points for testing.
We assume that this setting will give us the best information.
As we can see, we can now actually say which attributes are most important. 
We hope that these interactions between the attributes are also medically correct and that this relationship can be explained by the experts.

\begin{figure}[!ht]
	\begin{subfigure}[t]{0.47\linewidth}
		\centering
		
		\begin{tikzpicture}
			\begin{axis}[width=7.5cm, height=7.5cm, xmin=0, xmax=46,  enlargelimits=0.05, grid = major, grid style={dashed, gray!30},  xtick = {0,10,20,30,40}, ytick = {0, 0.05, 0.1}, enlargelimits=0.02, yticklabels = {$0$, $0.05$, $0.1$},xtick={1, 15,30,45}, xticklabels={$1$, $15$,$30$,$45$}]
				\coordinate (pt) at (axis cs:15,0.045);
				\addplot+[ycomb, orange, mark options={orange}, mark size=1pt, mark =*] plot coordinates {(1,0.23875501189912143)(2,0.21809528460581848)(3,0.06418695608519177)(4,0.06166155596762656)(5,0.05564084338528993)(6,0.05198344198185455)(7,0.05130469869733022)(8,0.0497680789523129)(9,0.04860311084780771)
				};
				\addplot+[ycomb, gray, mark options={gray}, mark size=1pt, mark =*] plot coordinates {(10,0.031100931965599293)(11,0.028180022984937155)(12,0.025309753639043506)(13,0.024985057480003858)(14,0.01587500380624399)(15,0.01317540897993811)(16,0.008269354807502599)(17,0.0051435330152614705)(18,0.003959570938876137)(19,0.0012170146170795018)(20,0.0009267000508445236)(21,0.0008057581176412089)(22,0.0005926212131807971)(23,0.0004602859614944056)(24,0.0)(25,0.0)(26,0.0)(27,0.0)(28,0.0)(29,0.0)(30,0.0)(31,0.0)(32,0.0)(33,0.0)(34,0.0)(35,0.0)(36,0.0)(37,0.0)(38,0.0)(39,0.0)(40,0.0)(41,0.0)(42,0.0)(43,0.0)(44,0.0)(45,0.0)
				};
				\addplot+[ycomb, white, mark options={white}, mark size=1pt, mark =*] plot coordinates {(1.5, 0.25)
				};
				\addplot[darkgray,sharp plot,update limits=false, dashed, thick] coordinates {(0.0,0.039) (10.0,0.039)};
				\addplot[darkgray,sharp plot,update limits=false, dashed, thick] coordinates {(0.0,0.2425) (10.0,0.2425)};
				\addplot[darkgray,sharp plot,update limits=false, dashed, thick] coordinates {(0.0,0.039) (0.0,0.243)};
				\addplot[darkgray,sharp plot,update limits=false, dashed, thick] coordinates {(10,0.039)(10,0.243)};
				
				\addplot[darkgray,sharp plot,update limits=false, dashed, thick] coordinates {(10,0.2425)(24,0.1915)};
				\addplot[darkgray,sharp plot,update limits=false, dashed, thick] coordinates {(10,0.039)(23,0.0715)};
			\end{axis}
			\node[anchor=south west] at (pt) {
				\begin{tikzpicture}
					\begin{axis}[footnotesize, width=4.5cm, height=4.5cm, xmin=0, xmax=10, xtick = {0,5,9},  ytick = {0.15, 0.05}, yticklabels = {0.15, 0.05}, enlargelimits=0.02,axis background/.style={fill=white}, xtick={1, 5,9}, xticklabels={$1$, $5$,$9$}]
						\addplot+[ycomb, orange, mark options={orange}, mark size=1.5pt, mark=*] plot coordinates {(1,0.23875501189912143)(2,0.21809528460581848)(3,0.06418695608519177)(4,0.06166155596762656)(5,0.05564084338528993)(6,0.05198344198185455)(7,0.05130469869733022)(8,0.0497680789523129)(9,0.04860311084780771)};
						(1,[1])(2,[6])(3,[1, 4])(4,[3, 8])(5,[6, 8])(6,[2])(7,[2, 6])(8,[3, 4])(9,[6, 9])
						\addplot+[ mark options={white}, mark size=0.5pt, mark=*] plot coordinates {(1.5,0.3)};
						\node at (axis cs:1,0.3) [anchor=north] {\rotatebox{90}{\scriptsize{$\{1\}$}}};
						\node at (axis cs:2,0.28) [anchor=north] {\rotatebox{90}{\scriptsize{$\{6\}$}}};
						\node at (axis cs:3,0.15) [anchor=north] {\rotatebox{90}{\scriptsize{$\{1,4\}$}}};
						\node at (axis cs:4,0.145) [anchor=north] {\rotatebox{90}{\scriptsize{$\{3,8\}$}}};
						\node at (axis cs:5,0.14) [anchor=north] {\rotatebox{90}{\scriptsize{$\{6,8\}$}}};
						\node at (axis cs:6,0.10999) [anchor=north] {\rotatebox{90}{\scriptsize{$\{2\}$}}};
						\node at (axis cs:7, 0.135) [anchor=north] {\rotatebox{90}{\scriptsize{$\{2,6\}$}}};
						\node at (axis cs:8,0.1325) [anchor=north] {\rotatebox{90}{\scriptsize{$\{3,4\}$}}};
						\node at (axis cs:9,0.1325) [anchor=north] {\rotatebox{90}{\scriptsize{$\{6,9\}$}}};
					\end{axis}
				\end{tikzpicture}
			};
		\end{tikzpicture}
		\caption{}
	\end{subfigure}
	\hspace{10pt}
	\begin{subfigure}[t]{0.47\linewidth}
		\centering
		\begin{tikzpicture}
			\begin{axis}[width=7.5cm, height=7.5cm, xmin=0, xmax=37, enlargelimits=0.02, grid = major, grid style={dashed, gray!30},  ytick = {0, 0.05, 0.1, 0.15},  yticklabels = {$0$, $0.05$, $0.1$, $0.15$}, xtick={1, 15, 30,36}, xticklabels={$1$, $15$,$30$,$36$}]
				\coordinate (pt) at (axis cs:10,0.095);
				\addplot+[ycomb, white, mark options={white}, mark size=1pt, mark=*] plot coordinates {(1,0.55)};
				\addplot+[ycomb, orange, mark options={orange}, mark size=1pt, mark=*] plot coordinates {(1,0.4621205078531591)(2,0.12673869207362606)(3,0.08445460401976507)(4,0.08060853415515004)(5,0.07143969383275091)(6,0.06588399093362805)};
				\addplot+[ycomb, gray, mark options={gray}, mark size=1pt, mark=*] plot coordinates {(7,0.040342183746659224)(8,0.025411814863812235)(9,0.023182807792134504)(10,0.011551253348626824)(11,0.008096956379980553)(12,7.530046158595495e-5)(13,6.66836324018321e-5)(14,2.6976906719421263e-5)(15,0.0)(16,0.0)(17,0.0)(18,0.0)(19,0.0)(20,0.0)(21,0.0)(22,0.0)(23,0.0)(24,0.0)(25,0.0)(26,0.0)(27,0.0)(28,0.0)(29,0.0)(30,0.0)(31,0.0)(32,0.0)(33,0.0)(34,0.0)(35,0.0)(36,0.0)};
				\addplot[darkgray,sharp plot,update limits=false, dashed, thick] coordinates {(0,0.05) (7,0.05)};
				\addplot[darkgray,sharp plot,update limits=false, dashed, thick] coordinates {(0.0,0.475) (7,0.475)};
				\addplot[darkgray,sharp plot,update limits=false, dashed, thick] coordinates {(0.0,0.05) (0.0,0.475)};
				\addplot[darkgray,sharp plot,update limits=false, dashed, thick] coordinates {(7,0.05)(7,0.475)};
				\addplot[darkgray,sharp plot,update limits=false, dashed, thick] coordinates {(7,0.05)(18,0.172)};
				\addplot[darkgray,sharp plot,update limits=false, dashed, thick] coordinates {(7,0.475)(18,0.42)};
			\end{axis}
			\node[anchor=south west] at (pt) {
				\begin{tikzpicture}
					\begin{axis}[footnotesize, width=4.5cm, height=4.5cm, xmin=0, xmax=7, xtick = {0,4,7},  ytick = {0.15, 0.05}, yticklabels = {0.15, 0.05}, enlargelimits=0.02,axis background/.style={fill=white}, xtick={1, 3,6}, xticklabels={$1$, $3$,$6$}]
						\addplot+[ycomb, orange, mark options={orange}, mark size=1.5pt, mark=*] plot coordinates {(1,0.4621205078531591)(2,0.12673869207362606)(3,0.08445460401976507)(4,0.08060853415515004)(5,0.07143969383275091)(6,0.06588399093362805)};
						\addplot+[ mark options={white}, mark size=0.5pt] plot coordinates {(1.5,0.65)};
						\node at (axis cs:1,0.6) [anchor=north] {\rotatebox{90}{\scriptsize{$\{2\}$}}};
						\node at (axis cs:2,0.325) [anchor=north] {\rotatebox{90}{\scriptsize{$\{5,6\}$}}};
						\node at (axis cs:3,0.225) [anchor=north] {\rotatebox{90}{\scriptsize{$\{7\}$}}};
						\node at (axis cs:4,0.275) [anchor=north] {\rotatebox{90}{\scriptsize{$\{2,5\}$}}};
						\node at (axis cs:5,0.205) [anchor=north] {\rotatebox{90}{\scriptsize{$\{8\}$}}};
						\node at (axis cs:6,0.265) [anchor=north] {\rotatebox{90}{\scriptsize{$\{2, 7\}$}}};
					\end{axis}
				\end{tikzpicture}
			};
		\end{tikzpicture}
		\caption{}
	\end{subfigure}
	\caption{Computed global sensitivity indices $\varrho(\u, S(\X,\I_{\b N}(U_2))f^\text{cos})$ with $\bfh=\bfh^{\cos}$. The results for the WBC data set, using $N_1=N_2=4$ and $\lambda=2^{-4}$, are shown in (a). For the PID data set we use $N_1=4$, $N_2=2$  and $\lambda=2^{-4}$ and visualize the results in (b). The GSIs depend on the randomly chosen $614$ (WBC) or rather $691$ (PID) training points and are sorted according to the importance. The first variable interactions with the highest importance are highlighted in orange.}
	\label{fig:real_gsi}
\end{figure}

\section{Summary and outlook}\label{sec:summary}

The truncated ANOVA decomposition in the form of Fourier partial sums, and the like, have previously been used primarily for regression tasks.
The goal of this work was to use such approaches in the form of a feature map within Support Vector Machines.
The advantage here is that we can calculate the necessary matrix-vector products efficiently with grouped transformations, since we only restrict to couplings of small groups of variables.
Furthermore, the models can be interpreted well in terms of the global sensitivity indices.
Our numerical examples, in which we restricted our considerations to very simple test functions,  show that we can achieve good classification results and that these can be interpreted as well. 
We have also demonstrated the applicability of our method on freely available data sets and achieved good results as obtained previously in the literature.
The next step would be more theoretical considerations, as well as the extension of our methods to multiclass problems.

\section*{Acknowledgments}

The authors gratefully acknowledge their support by the BMBF grant 01|S20053A (project SA$\ell$E) and thank Felix Bartel, Michael Schmischke, Pascal Schröter and Laura Weidensager for fruitful discussions.
Furthermore, we thank the anonymous reviewers for their very helpful suggestions.





\newpage
\appendix

\section{The Friedman Function}
In this section we discuss the Frieman 1 function~\eqref{eq:friedman1} in more detail.
First, we compute the active ANOVA terms in Section~\ref{sec:gsi_friedman}.
Based on a numerical test we verify how significant the single ANOVA terms contribute to the signum of the zero-mean function $F^1-M(F^1)$, see Section~\ref{sec:importance_friedman}.

\subsection{Computing the ANOVA-Decomposition of the Friedman Function\label{sec:gsi_friedman}}

For the Friedman 1 function as defined in \eqref{eq:friedman1} it is easy to see that the active set of subsets $\u$ will be included in $U^\star=\{\emptyset,\{1\},\{2\},\{3\},\{4\},\{5\},\{1,2\}\}$.
In the following we state the ANOVA terms $F^1_\u(\x_\u)$ for all $\u\in U^\star$.

The constant part of the function is given by
$$
F^1_\emptyset=P_\emptyset F^1 = M(F^1) \approx 14.4133,
$$
as already stated above, cf.~\eqref{eq:MF1}. Next, we compute for the subsets $\u\in U^\star$ with $|\u|=1$
\begin{align*}
	F^1_{\{1\}}(x_1)=P_{\{1\}}F^1(x_1)-M(F^1) &= \frac{10-10\cos(\pi x_1)}{\pi x_1}+\frac{55}{6} - M(F^1), \\
	F^1_{\{2\}}(x_2)=P_{\{2\}}F^1(x_2)-M(F^1) &= \frac{10-10\cos(\pi x_2)}{\pi x_2}+\frac{55}{6} - M(F^1), \\
	F^1_{\{3\}}(x_3)=P_{\{3\}}F^1(x_3)-M(F^1) &= 10\int\frac{1-\cos(\pi t)}{\pi t}\mathrm dt+20\left(x_3-\tfrac 12\right)^2+\frac{15}{2} -M(F^1), \\
	F^1_{\{4\}}(x_4)=P_{\{4\}}F^1(x_4)-M(F^1) &= 10\int\frac{1-\cos(\pi t)}{\pi t}\mathrm dt +10x_4+ \frac{25}{6} -M(F^1),\\
	F^1_{\{5\}}(x_5)=P_{\{5\}}F^1(x_5)-M(F^1) &= 10\int\frac{1-\cos(\pi t)}{\pi t}\mathrm dt +5x_5+ \frac{20}{3} -M(F^1).
\end{align*}
The value of the remaining integral
$$
10\int\frac{1-\cos(\pi t)}{\pi t}\mathrm dt \approx 5.2466
$$
has to be computed numerically.
	Finally, the ANOVA term of the highest order is given by
	\begin{align*}
		F^1_{\{1,2\}}(x_1,x_2)
		&=P_{\{1,2\}}F^1(x_1,x_2)-F^1_{\{1\}}(x_1)-F^1_{\{2\}}(x_2)-M(F^1)\\
		&=10\sin(\pi x_1x_2)+\frac{55}{6}-F^1_{\{1\}}(x_1)-F^1_{\{2\}}(x_2)-M(F^1)\\
		&=10\sin(\pi x_1x_2)+ \frac{10\cos(\pi x_1)-10}{\pi x_1} + \frac{10\cos(\pi x_2)-10}{\pi x_2}-\frac{55}{6}+M(F^1).
	\end{align*}
	It can easily be checked that the sum of all ANOVA terms gives in fact the function $F^1$.
	The computed ANOVA terms are all non-zero and, thus, the active set is indeed given by $U^\star$ as stated above.
	
	
	\subsection{Importance of ANOVA Terms to the Signum \label{sec:importance_friedman}}
	
	Based on the following numerical test we check the importance of the single ANOVA terms to the signum of the zero-mean function $F^{1}(\x)-M(F^1)$ in $d=5$ dimensions, i.e., we restrict the Friedman-1 function to the important variables only.
	
	We generated $N=10^6$ randomly and uniformly distributed points $\boldsymbol z_j\in[0,1]^5$, $j=1,\dots,5$.
	By $y_j=\sign(F^1(\boldsymbol z_j)-M(F^1))$ we obtain a well-balanced data set, see Table~\ref{tab1:friedman}. 
	Then, we checked whether leaving out one of the ANOVA terms will change the signum of the function values.
	It can be seen that all ANOVA terms contribute significantly to $\sign(F^1-M(F^1))$, see Table~\ref{tab2:friedman}.
	As an example, when omitting the ANOVA term $F^1_{\{4\}}$, the sign of the function value will change for approximately $20\%$ of the data points.
	
	\begin{table}[ht]
		\centering
		\begin{tabular}{r|l}
			Total number of points $\boldsymbol z_j$ & $10^6$ \\
			\hline
			Points with $\sign(F^{1}(\boldsymbol z_j)-M(F^1))=+1$ & $\approx 499600$ \\
			Points with $\sign(F^{1}(\boldsymbol z_j)-M(F^1))=-1$ & $\approx 500400$
		\end{tabular}
		\vspace{1.5em}
		\caption{By subtracting the mean value of the function we obtain two classes of approximately the same size.
			\label{tab1:friedman}}
	\end{table}
	
	\begin{table}[ht]
		\centering
		\begin{tabular}{c|c}
			subset $\u$ & $\sign(F^{1}(\boldsymbol z_j)-M(F^1)) \neq$\\
			& $\sign(F^{1}(\boldsymbol z_j)-M(F^1)-F^{1}_\u(\boldsymbol z_{j,\u}))$\\
			\hline
			$\emptyset$ & $\approx 50.05\%$ \\
			$\{1\}$ & $\approx 15.60\%$ \\
			$\{2\}$ & $\approx 15.58\%$ \\
			$\{3\}$ & $\approx 10.15\%$ \\
			$\{4\}$ & $\approx 20.10\%$ \\
			$\{5\}$ & $\approx \phantom{0}9.89\%$ \\
			$\{1,2\}$ & $\approx \phantom{0}8.05\%$ \\
		\end{tabular}
		\vspace{1.5em}
		\caption{For each subset $\u$ or rather ANOVA term $F^1_\u$ we state the percentage of the randomly chosen points $\boldsymbol z_j$, $j=1,\dots,10^6$, for which the sign changes if the ANOVA term is omitted.
			\label{tab2:friedman}}
	\end{table}

	\end{document}